%% file: main.tex
\documentclass[conference]{IEEEtran}
\IEEEoverridecommandlockouts
\usepackage{cite}
\usepackage{amsmath,amssymb,amsfonts}
\usepackage{algorithmic}
\usepackage{graphicx}
\usepackage{textcomp}
\usepackage[svgnames,xcdraw,table]{xcolor}
\usepackage{multirow}
\usepackage{multicol}
\usepackage{subcaption}
\usepackage{mdframed}
\def\BibTeX{{\rm B\kern-.05em{\sc i\kern-.025em b}\kern-.08em
    T\kern-.1667em\lower.7ex\hbox{E}\kern-.125emX}}

\IEEEtriggeratref{59} % to balance the references on the last page

\newcommand{\htag}[1]{%
  \textcolor{teal}{\footnotesize\textbf{[#1]}}%
}

\newcommand{\gradientcell}[1]{%
  \pgfmathsetmacro{\shadevalue}{min(max(#1*2,0),100)}% % Calculate float value
  \pgfmathparse{\shadevalue}% % Parse it
  \let\tempshade\pgfmathresult % Store the result
  \pgfmathsetmacro{\integershade}{int(\tempshade)}% % Convert to integer
  \edef\tempcellcolor{\noexpand\cellcolor{blue!\integershade!white}}%
  \tempcellcolor#1%
}

\usepackage{graphicx}

\newmdenv[
    topline=false, bottomline=false, rightline=false,
    linewidth=2.5pt,
    linecolor=gray!60,
    innerleftmargin=8pt,
    innerrightmargin=0pt,
    innertopmargin=2pt,
    innerbottommargin=2pt,
    skipabove=6pt,
    skipbelow=6pt,
]{promptbox}

\definecolor{increase}{RGB}{198, 224, 180}   % soft green
\newcommand{\increase}[1]{\cellcolor{increase}\textbf{#1}}

\definecolor{decrease}{RGB}{198, 180, 224}   % soft purple
\newcommand{\decrease}[1]{\cellcolor{decrease}\textbf{#1}}

\usepackage{tcolorbox}
\tcbset{
    colback=gray!5,
    colframe=gray!75,
    boxrule=0.5pt,
    arc=2pt,
    left=6pt,
    right=6pt,
    top=6pt,
    bottom=6pt
}
\usepackage{minted}
\usepackage{siunitx}

\newcommand{\lesser}{\cellcolor{myred} \(<\) }

\newtheorem{result}{Finding}

\definecolor{pos}{RGB}{0,102,204}   % blue for positive
\definecolor{posmore}{RGB}{0,102,255}   % blue for positive
\definecolor{neg}{RGB}{204,0,0}     % red for negative
\definecolor{negmore}{RGB}{240,0,0}     % red for negative
\definecolor{nan}{gray}{0.8}        % light gray for NaN
\definecolor{groupgray}{RGB}{240,240,240} % Light background for groups

\renewcommand{\aligned}{\textcolor{pos}{+1}}
\newcommand{\morealigned}{\textcolor{posmore}{+2}}
\newcommand{\misaligned}{\textcolor{neg}{-1}}
\newcommand{\moremisaligned}{\textcolor{negmore}{-2}}
\newcommand{\none}{\textcolor{nan}{o}}

\usepackage{float}
\usepackage{adjustbox}
\definecolor{myBlue}{rgb}{0.1,0.1,0.8}
\definecolor{DarkGreen}{rgb}{0.1,0.5,0.1}
\definecolor{mygreen}{rgb}{0.5,0.9,0.5}
\definecolor{myred}{rgb}{1, 0.28, 0.3}
\usepackage{hyperref}
\hypersetup{
	colorlinks=true,
	linkcolor=red,%
	urlcolor=DarkGreen,%
	citecolor=blue%
}
\usepackage[capitalise,noabbrev]{cleveref}
\usepackage{booktabs}
\usepackage{multirow}
\usepackage{tabularx}
\usepackage{makecell}
\usepackage{array}

\begin{document}

\title{Expressing Social Emotions: Misalignment Between LLMs and Human Cultural Emotion Norms}

% optional title: 

\author{
\IEEEauthorblockN{
Sree Bhattacharyya\textsuperscript{1},
Manas Mehta\textsuperscript{1},
Leona Chen\textsuperscript{1},
Cristina Salvador\textsuperscript{2},
Agata Lapedriza\textsuperscript{3,4}, \\
Shiran Dudy\textsuperscript{3},
James Z. Wang\textsuperscript{1}
}

\IEEEauthorblockA{
\textsuperscript{1}The Pennsylvania State University \quad
\textsuperscript{2}Duke University \quad
\textsuperscript{3}Northeastern University \quad
\textsuperscript{4}Universitat Oberta de Catalunya \\
\texttt{\{sreeb, mvm7168\}@psu.edu}, \texttt{s.dudy@northeastern.edu}
}
}

% \author{Anonymous ACII Submission}

\maketitle

\begin{abstract}
The expression of emotions that serve social purposes, such as asserting independence or fostering interdependence, is central to human interactions and varies systematically across cultures. As LLMs are increasingly used to simulate human behavior in culturally nuanced interactions, it is important to understand whether they faithfully capture human patterns of social emotion expression. When LLM responses are not culturally aligned, their utility is compromised---particularly when users assume they are interacting with a culturally attuned interlocutor, and may act on advice that proves inappropriate in their cultural context. We present a psychologically informed evaluation framework of cross-cultural social emotion expression in LLMs. Using a human study comparing European American and Latin American participants' expression of engaging and disengaging emotions, we evaluate six frontier LLMs on their ability to reflect culturally differentiated patterns for expressing social emotions. We find systematic misalignment between model and human behavior: all models express engaging emotions more than disengaging ones, with particularly stark differences observed for the generally well-represented European American persona. We further highlight that LLM responses are highly concentrated and deterministic, failing to capture the diversity of human responses in expressing social emotions. Our ablation analyses reveal that these patterns are robust to sampling temperatures, partially sensitive to prompt language, and dependent on the response elicitation format. Together, our findings highlight limitations in how current LLMs represent the interaction of cultural and emotional axes, particularly when expressing social emotions, with direct implications for their deployment in cross-cultural affective contexts.

\end{abstract}


\input{content/1_introduction}
\input{content/2_related_work}
\input{content/3_methods}
\input{content/4_results}
\input{content/5_ablation}
\input{content/6_discussion}

\section*{Acknowledgements}

This material is based upon work supported in part by the National Science Foundation (NSF) under Award No. 2234195, and the Penn State 2024-25 Vice Provost and Dean of the Graduate School Student Persistence Scholarship. This work used cluster computers at the National Center for Supercomputing Applications and the Pittsburgh Supercomputing Center through an allocation from the Advanced Cyberinfrastructure Coordination Ecosystem: Services \& Support (ACCESS) program, which is supported by NSF Award Nos. 2138259, 2138286, 2138307, 2137603, and 2138296.

\bibliography{biblio}
\bibliographystyle{ieeetr}

\clearpage
\appendix
\input{content/7_appendix}

\end{document}

%% file: content/1_introduction.tex
\section{Introduction}
\label{sec:introduction}

The ever-increasing popularity of Large Language Models (LLMs) has seen their deployment across diverse cultural contexts. Systematic cultural differences in human behavior~\cite{triandis1989self, henrich2005economic}, beliefs~\cite{zou2009culture}, and norms~\cite{heinrichs2006cultural} require that LLMs not only demonstrate multilingual capabilities, but also reflect the cultural norms of their users. When these systems fail to align with cultural norms, users may unknowingly act on culturally inappropriate guidance, with consequences ranging from ineffective advice to active harm. For instance, in mental health, the absence of cultural competency has been shown to cause disparities in service quality for minority patients~\cite{sue2009case, wendt2015potentially, american2003guidelines}. Misaligned responses raise further concerns for studies that simulate human behavior using LLMs. Despite its financial benefits, evidence is mixed on the effectiveness of this approach~\cite{cui2024can, bisbee2024synthetic, park2024diminished}. Subjective aspects of behavior, such as emotions, are also particularly challenging to simulate owing to their interpretive nature.

Several recent studies apply a cultural lens to emotions, studying how well models capture cross-cultural differences~\cite{belay-etal-2025-culemo, dudy2024analyzing, rai2025social, havaldar2023multilingual}. Belay et al.~\cite{belay-etal-2025-culemo}, for instance, study basic emotions across culturally different situations, while Rai et. al.~\cite{rai2025social} study shame and pride in the context of movies, and Dudy et al.~\cite{dudy2024analyzing} study American--Japanese differences in mixed emotion expression. Similar to Dudy et al., we evaluate LLM--human alignment through direct comparison with cross-cultural human data, but focus specifically on \textit{social emotions}, which serve the purpose of expression in the context of our environment and our relationship with the environment. As achieving social goals has also been shown to be a central and critical purpose of human emotions~\cite{keltner2001social, boiger2022different, savani2013feeling, kitayama2006cultural}, studying social emotions in the context of LLMs is critical to help modern AI systems participate in meaningful, robust, and accurate affective social interactions with other agents. 

We compare LLMs with a human study~\cite{salvador2024emotionally} examining differences between European Americans and the understudied population of Latin Americans~\cite{henrich2010weirdest, kitayama2022varieties, campos2017incorporating} in the expression of engaging and disengaging emotions. Engaging emotions (e.g., friendliness, guilt) establish interdependence, while disengaging emotions (e.g., pride, anger) express individuality. The human study finds that Latin Americans express engaging emotions more, while European Americans express disengaging emotions more, consistent with collectivist and individualist cultural norms, respectively. We evaluate six frontier LLMs by administering the cross-cultural survey from Salvador et al.~\cite{salvador2024emotionally}, to address the following research questions:

\begin{itemize}
\item[\textbf{RQ1}] How well are LLMs aligned with the cross-cultural patterns of social emotion expression established by Salvador et al.~\cite{salvador2024emotionally}?
\item[\textbf{RQ2}] Is the observed (mis)alignment influenced by sampling temperature, prompt language, or response elicitation format?
\end{itemize}

For RQ1, we examine whether LLMs reproduce broad cultural patterns and compare the distributional structure of human and LLM responses. For RQ2, we study the effect of sampling temperature, English-language prompting, and a pairwise forced-choice task format.

Our findings reveal systematic failure of LLMs to reproduce broad cultural patterns from human-subject data. Most strikingly, all models misalign on the tendency of European Americans to express disengaging over engaging emotions---despite this population being well-represented in training data. LLM response distributions are also found to be highly modal, failing to capture the natural diversity of human expression. We also find that increasing sampling temperatures has a limited effect on diversifying responses and does not help in improving alignment. Switching Latin American prompts from Spanish to English improves alignment, revealing that models encode cultural knowledge better in a language not native to the culture. Finally, a forced-choice format improves alignment on overall cultural expressivity, but fails to resolve misalignment in the relative expression of engaging and disengaging emotions.

%% file: content/2_related_work.tex
\section{Related Work}
\label{sec:related_work}

\indent{\textbf{Cultural Influence on Emotion Expression and Interpretation.}} While Ekman's foundational work proposed universal facial expressions across cultures~\cite{ekman1971constants, ekman1992argument}, subsequent research challengenged their results. Cross-cultural studies show that cultures differ in which emotional states they value---Western cultures favor high-arousal positive emotions like excitement, while East Asian cultures favor low-arousal states like calmness~\cite{tsai2006cultural, tsai2007ideal}. Emotional experience is further shaped by culturally specific social contexts and relational norms~\cite{mesquita2001emotions, mesquita2022between}, and people follow distinct cultural display rules governing when and how emotions are expressed~\cite{matsumoto1990cultural, matsumoto2008culture}.

\indent{\textbf{Cultural Bias in Emotion Understanding of LLMs.}} Growing research has examined cultural biases in LLMs~\cite{kamruzzaman-etal-2025-seeing, dai2026tearscheersbenchmarkingllms, 11045290}. Since training corpora are disproportionately drawn from English-speaking and Western sources, LLM responses often reflect Western biases~\cite{tao2024cultural, yu2026entangledrepresentationsmechanisticinvestigation, naous-etal-2024-beer}, and several approaches have been proposed to measure them~\cite{dai-etal-2025-word, liu2025towards, plaza-del-arco-etal-2024-divine, hu2025generative}. Persona-based prompting has shown that assigned cultural personas can significantly shift predicted emotion categories~\cite{kamruzzaman-etal-2025-anger}, and models lean heavily on culturally biased priors when interpreting emotional scenarios~\cite{10970320}. Cross-cultural benchmarks have also been introduced: CULEMO~\cite{belay-etal-2025-culemo} measures alignment between LLM predictions and human emotion labels across cultures, while Dudy et al.~\cite{10970282} examine whether LLM representations of complex emotional states reflect culturally specific patterns. Across this body of work, LLM responses consistently align most closely with WEIRD\footnote{WEIRD stands for \underline{W}estern, \underline{E}ducated, \underline{I}ndustrialized, \underline{R}ich, \underline{D}emocratic.} emotional interpretations~\cite{henrich2010weirdest}, and mitigating these biases remains an open challenge.

\indent{\textbf{Evaluating Nuanced Emotional Reasoning in LLMs.}} Beyond bias, recent work has evaluated whether LLMs understand complex emotional nuances. Traditional affective computing focused on emotion classification and sentiment analysis~\cite{strapparava-mihalcea-2007-semeval, mohammad-etal-2018-semeval, socher-etal-2013-recursive, pang-etal-2002-thumbs}, but evaluations have since expanded to emotion-cause reasoning, empathetic response generation, and conversational emotion understanding. Zhou et al.~\cite{zhou2025textbased} probe empathetic inference against trained human counselors; appraisal-theoretic frameworks find limited cognitive emotional reasoning~\cite{yeo-jaidka-2025-beyond, bhattacharyya2025machines}; and Hong et al.~\cite{hong-etal-2025-third} propose an appraisal-based agent architecture to enhance emotional inference. Evaluation has further extended to multimodal settings~\cite{peng2026emotionllamav2mmeversenewframework}, sensitivity to evaluation formats~\cite{bhattacharyya2025evaluating}, and emotion-cause analysis in conversation~\cite{belikova-kosenko-2024-deeppavlov}.

%% file: content/3_methods.tex
\section{Methods}
\label{sec:evaluation_framework}

\begin{table*}[t]
    \centering
    \resizebox{\textwidth}{!}{
    \begin{tabular}{l *{8}{l}}
    \toprule
    \textbf{Emotion} 
    & feelings of closeness to others & friendly feelings & self-esteem & pride & shame & anger & frustration & guilt \\
    \midrule
    \textbf{Emotion Type} 
    & engaging & engaging & disengaging & disengaging & engaging & engaging & disengaging & disengaging \\
    \textbf{Emotion Valence} 
    & positive & positive & positive & positive & negative & negative & negative & negative \\
    \textbf{Group Acronym} & PSE & PSE & PSD & PSD & NSE & NSE & NSD & NSD \\
    \bottomrule
    \end{tabular}}
    \caption{Classification of social emotions, as used in the original human study~\cite{salvador2024emotionally}, and in our experiments with LLMs. The original study also includes four basic emotions: calm, happy, unhappy, and elated.}
    \label{tab:emotions}
\end{table*}

\begin{table}[t]
    \centering
    \resizebox{\columnwidth}{!}{
    \begin{tabular}{cc}
    \toprule
         \textbf{Hypothesis} &  \textbf{Human Results} \\ 
         \midrule
         \multirow{2}{*}{\htag{H1} Expression of Engaging vs. Disengaging Emotions} & \htag{H1a} \textbf{LA:} EN \(>\) D-EN \\
         & \htag{H1b} \textbf{EA:} EN \(<\) D-EN \\
         \midrule
         \multirow{2}{*}{\htag{H2} Expression of Positive Emotions} & \htag{H2a} PSE \(>\) PSD \\
         & \htag{H2b} Larger Diff. for \textbf{LA} \\
         \midrule
         \multirow{2}{*}{\htag{H3} Expression of Negative Emotions} & \htag{H3a} NSD \(>\) NSE \\
         & \htag{H3b} Larger Diff. for \textbf{EA} \\
         \midrule
         \htag{H4} Overall Expressiveness & LA \(>\) EA \\
         \midrule
         \htag{H5} Overall Expression of Emotion Valence & Positive \(>\) Negative \\
         \bottomrule
    \end{tabular}}
    \caption{Main results from the original human study by Salvador et al. \cite{salvador2024emotionally}. The results obtained for the studied human population are described, which form the hypotheses for our tests with LLMs.}
    \label{tab:human_results}
\end{table}

% \subsection{Cross-cultural Differences in Social Emotions: Latin Americans and European Americans}
\indent{\textbf{Differences in Social Emotion Expression between Latin and European Americans.}}
% \label{sec:evaluation_framework-human_study}
We ground our evaluation in a cross-cultural human study~\cite{salvador2024emotionally} comparing emotional expression across European Americans (EAs) and Latin Americans (LAs). The study examines \textit{social} emotions, categorized as \textit{socially engaging} (EN) (e.g., guilt, friendly feelings), which foster interdependence, or \textit{socially disengaging} (D-EN) (e.g., anger, pride), which do not foster interdependence and are more so an expression of the personal self. 598 participants, evenly distributed across cultural groups, rated the 8 emotions in Table~\ref{tab:emotions} and 4 basic emotions across four everyday situations (Appendix~A) on a 1--6 Likert scale~\cite{likert1932technique}. Three findings emerge: (a) LA participants express engaging over disengaging emotions across valence, while EA participants show the reverse; (b) for positive emotions, both groups favor engaging emotions, but the gap is larger for LAs; (c) for negative emotions, both groups favor disengaging emotions, but the gap is larger for EAs. These findings, summarized in Table~\ref{tab:human_results}, form the human ground truth against which we evaluate LLM behavior.\footnote{The human study also conducts finer-grained situational analysis, reproduced in Appendix~F.2}

\indent{\textbf{Evaluation Framework for LLMs.}}
% \label{sec:evaluation_framework-llm_evaluation}
We evaluate six frontier LLMs measuring alignment specifically in how models distinguish engaging from disengaging emotions and whether culturally distinct expression trends emerge. We select SOTA models ensuring diversity across origin, open-source versus proprietary access, and reasoning capability: DeepSeek R1~\cite{guo2025deepseek}, GPT 4o-mini~\cite{jaech2024openai}, Gemini 2.5 Flash~\cite{comanici2025gemini}, Phi 4~\cite{abdin2024phi}, Mistral 7B Instruct v0.3~\cite{jiang2023mistral7b}, and Qwen 3 32B~\cite{qwen3technicalreport}.
Each model is assigned cultural personas using identifier terms verbatim from the human study, and asked to rate all 12 emotions across four situations (Appendix~A) on the same 1--6 scale (prompt examples in Appendix~B). Each prompt is sampled n=190 times at default temperature settings, determined through a statistical stability analysis (Appendix~C), to construct response distributions comparable to the human sample. In accordance with the conditions in the human study, prompts for LA personas are administered in Spanish. This yields 48 unique prompts (4 situations $\times$ 12 emotions) and \textit{164,160 total generations} across all personas (3), models (6), and iterations (190). Before the full evaluation, we verify reliable scale usage via Kendall's $W$, finding moderate-to-strong average intra- and inter-model reliability of 0.76 and 0.66, respectively (details in Appendix~D).

% \label{sec:evaluation_framework-statistical_methods}
\indent{\textbf{Statistical Methods for Distributional Comparison.}} Our main analysis investigates the full distribution of model responses rather than summary statistics alone. We use the Mann-Whitney U test~\cite{mann1947test} for directional distributional comparisons---establishing whether one distribution is significantly greater or lesser than another---and the Wasserstein distance~\cite{vaserstein1969markov} to quantify the magnitude of their difference. The original human study's ANOVA~\cite{fisher1919xv} is unsuitable here, as near-zero-variance LLM responses cause the test to collapse (Appendix E). All distributional comparisons are reproduced on the human data for methodological consistency, and all original human study findings hold under this distributional evaluation.

%% file: content/4_results.tex
\section{Main Results}
\label{sec:results}

\begin{table*}[t]
\centering
\small
\resizebox{\textwidth}{!}{
\begin{tabular}{
    p{3.5cm}                              % Hypothesis column
    @{\hspace{10pt}\vrule\hspace{10pt}}
    *{18}{c}  
}
\toprule
\centering \textbf{Hypothesis} & \multicolumn{3}{c}{\textbf{GPT}} & \multicolumn{3}{c}{\textbf{Gemini}} & \multicolumn{3}{c}{\textbf{DeepSeek}} & \multicolumn{3}{c}{\textbf{Phi}} & \multicolumn{3}{c}{\textbf{Mistral}} & \multicolumn{3}{c}{\textbf{Qwen}} \\

\cmidrule(lr){2-4}
\cmidrule(lr){5-7}
\cmidrule(lr){8-10}
\cmidrule(lr){11-13}
\cmidrule(lr){14-16}
\cmidrule(lr){17-19}

& \textit{Main} & \textit{Temp} & \textit{Lang}
& \textit{Main} & \textit{Temp} & \textit{Lang}
& \textit{Main} & \textit{Temp} & \textit{Lang}
& \textit{Main} & \textit{Temp} & \textit{Lang}
& \textit{Main} & \textit{Temp} & \textit{Lang}
& \textit{Main} & \textit{Temp} & \textit{Lang} \\

\midrule

\htag{H1a} \textbf{LA}: EN \(>\) D-EN
& \aligned & \none & \none 
& \aligned & \none & \none 
& \aligned & \none & \none 
& \aligned & \misaligned & \none 
& \aligned & \none & \none
& \aligned & \none & \none \\ 

\htag{H1b} \textbf{EA}: EN \(<\) D-EN
& \misaligned & \none & \none
& \misaligned & \none & \none
& \misaligned & \none & \none
& \misaligned & \none & \none
& \misaligned & \none & \none
& \misaligned & \none & \none \\

\midrule

\htag{H2a} PSE \(>\) PSD
& \aligned & \none & \none
& \aligned & \none & \none
& \aligned & \none & \none
& \aligned & \none & \none
& \aligned & \none & \none
& \aligned & \none & \none \\

\htag{H2b} Larger diff.\ for LA
& \misaligned & \none & \morealigned
& \misaligned & \none & \morealigned
& \none & \misaligned & \aligned
& \misaligned & \none & \aligned
& \misaligned & \none & \morealigned
& \misaligned & \none & \none \\

\midrule

\htag{H3a} NSD \(>\) NSE
& \aligned & \none & \none
& \aligned & \none & \none
& \aligned & \none & \none
& \aligned & \none & \none
& \misaligned & \none & \none
& \aligned & \misaligned & \none \\

\htag{H3b} Larger diff.\ for EA
& \aligned & \none & \none
& \aligned & \none & \none
& \none & \aligned & \misaligned
& \misaligned & \none & \morealigned
& \none & \aligned & \misaligned
& \none & \none & \aligned \\

\midrule

\htag{H4} Expressiveness: LA \(>\) EA
& \none & \misaligned & \aligned
& \misaligned & \none & \morealigned
& \misaligned & \none & \morealigned
& \misaligned & \none & \none
& \aligned & \none & \none 
& \aligned & \moremisaligned & \none \\

\htag{H5} Valence: Positive \(>\) Negative
& \aligned & \none & \none
& \aligned & \none & \none
& \aligned & \none & \none
& \aligned & \none & \none
& \aligned & \none & \none
& \aligned & \none & \none \\

\midrule

Overall: [+9, -9]
& 3 & 2 (\misaligned) & 6 (\textcolor{pos}{+3})
& 2 & 2 (\none) & 6 (\textcolor{pos}{+4})
& 2 & 2 (\none) & 4 (\textcolor{pos}{+2})
& 2 & 1 (\misaligned) & 5 (\textcolor{pos}{+3})
& 1 & 2 (\aligned) & 2 (\aligned)
& 4 & 1 (\textcolor{neg}{-3}) & 5 (\aligned) \\

\bottomrule
\end{tabular}}
\caption{Alignment of LLMs with human trends of social emotion expression. Column 1 shows results from the human study~\cite{salvador2024emotionally}; the remaining columns show alignment for each LLM. \textit{+1}: model results reflect the same directionality as humans at a statistically significant level (p \(<\) 0.05). \textit{-1}: model results reflect the \textit{opposite} directionality at a statistically significant level. \textit{o}: no significant finding in either direction. The maximum alignment range per model is [-9, +9] as there are 9 hypotheses in total. For each model, \textit{Main} shows results from the original experiments; \textit{Temp} and \textit{Lang} show the \textit{change} in alignment score under temperature and language ablations, respectively.}
\label{tab:alignment_summary}
\end{table*}

We structure our analysis to directly reflect the principal findings of the human study described in Table \ref{tab:human_results}. A summary of alignment between LLMs and human responses is presented in Table~\ref{tab:alignment_summary}. In this section, we examine these results in detail, highlighting systematic deviations and areas of misalignment.

\begin{figure}[ht]
    \begin{subfigure}{\linewidth}
        \centering
        \includegraphics[width=\linewidth]{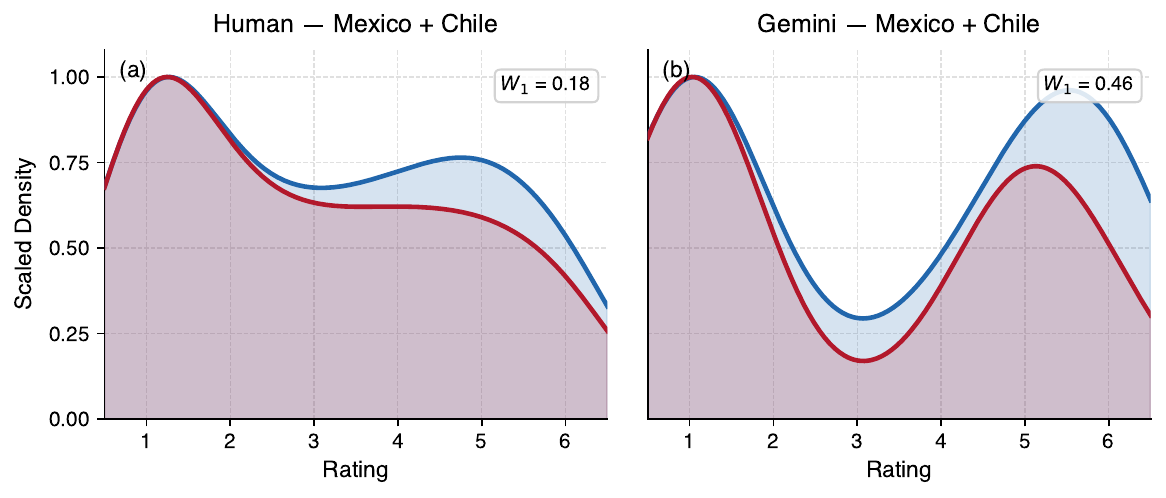}
        \caption{}
        \label{fig:gemini_eng_diseng_la}    
    \end{subfigure}
    \begin{subfigure}{\linewidth}
        \centering
        \includegraphics[width=\linewidth]{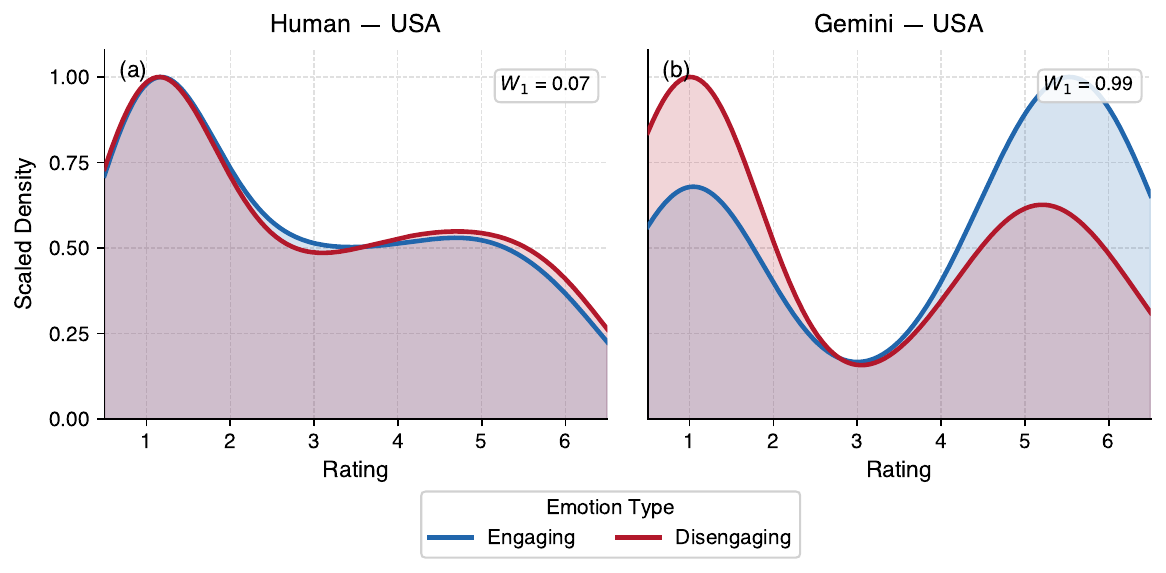}
        \caption{}
        \label{fig:gemini_eng_diseng_ea}    
    \end{subfigure}
    \caption{Rating distributions for engaging and disengaging emotions (with valence collapsed) for humans vs. Gemini 2.5 Flash. (a) Latin American personas combined; (b) European American persona. $W_1$: 1-Wasserstein distance. Corresponding alignment scores are shown in rows 1--2 of Table~\ref{tab:alignment_summary}.} 
\end{figure}

\begin{figure*}
\centering
    \begin{subfigure}{0.24\linewidth}
        \centering
        \includegraphics[width=\linewidth]{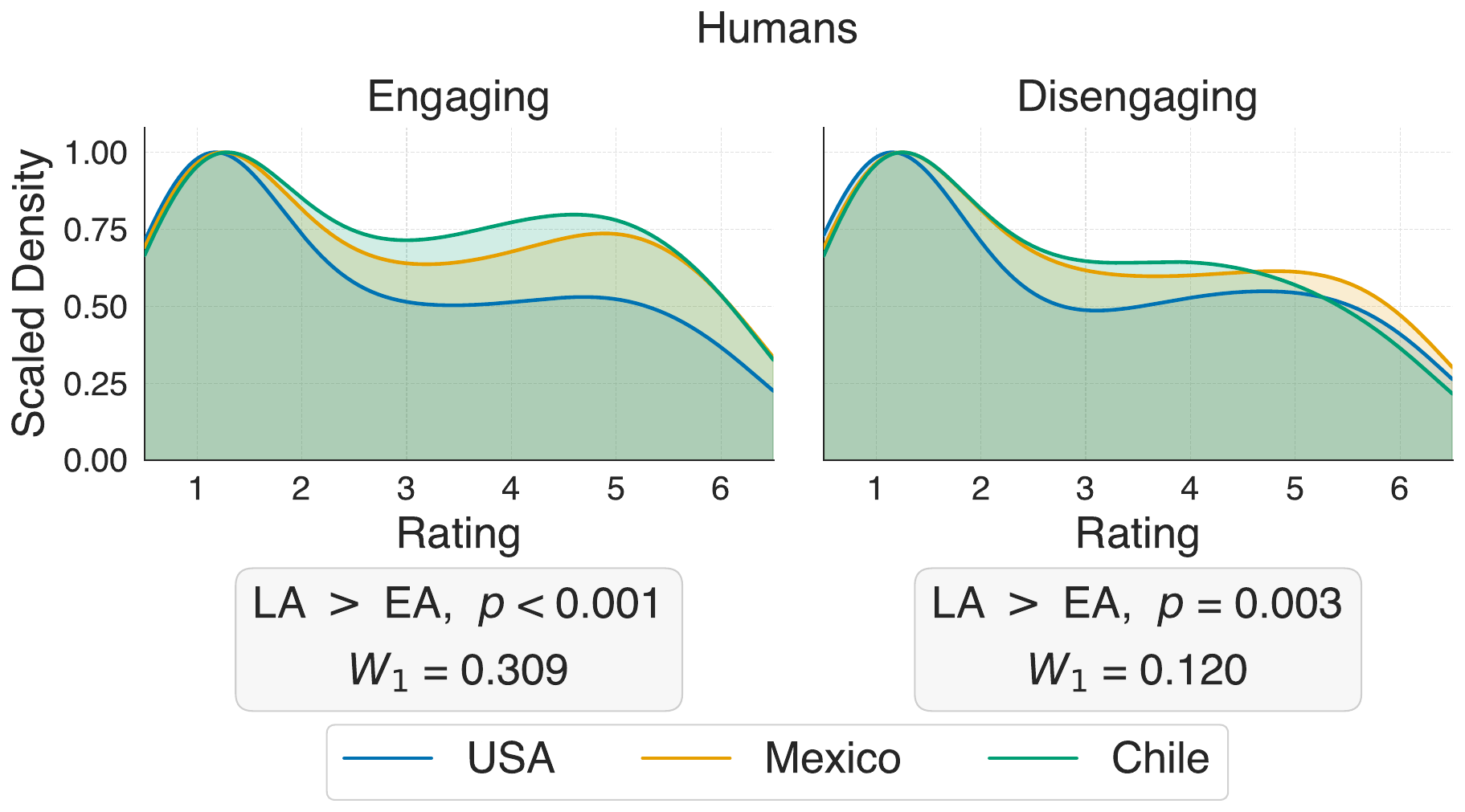}
        \caption{}
        \label{fig:human_absolute_eng_diseng}
    \end{subfigure}
    \begin{subfigure}{0.24\linewidth}
        \centering
        \includegraphics[width=\linewidth]{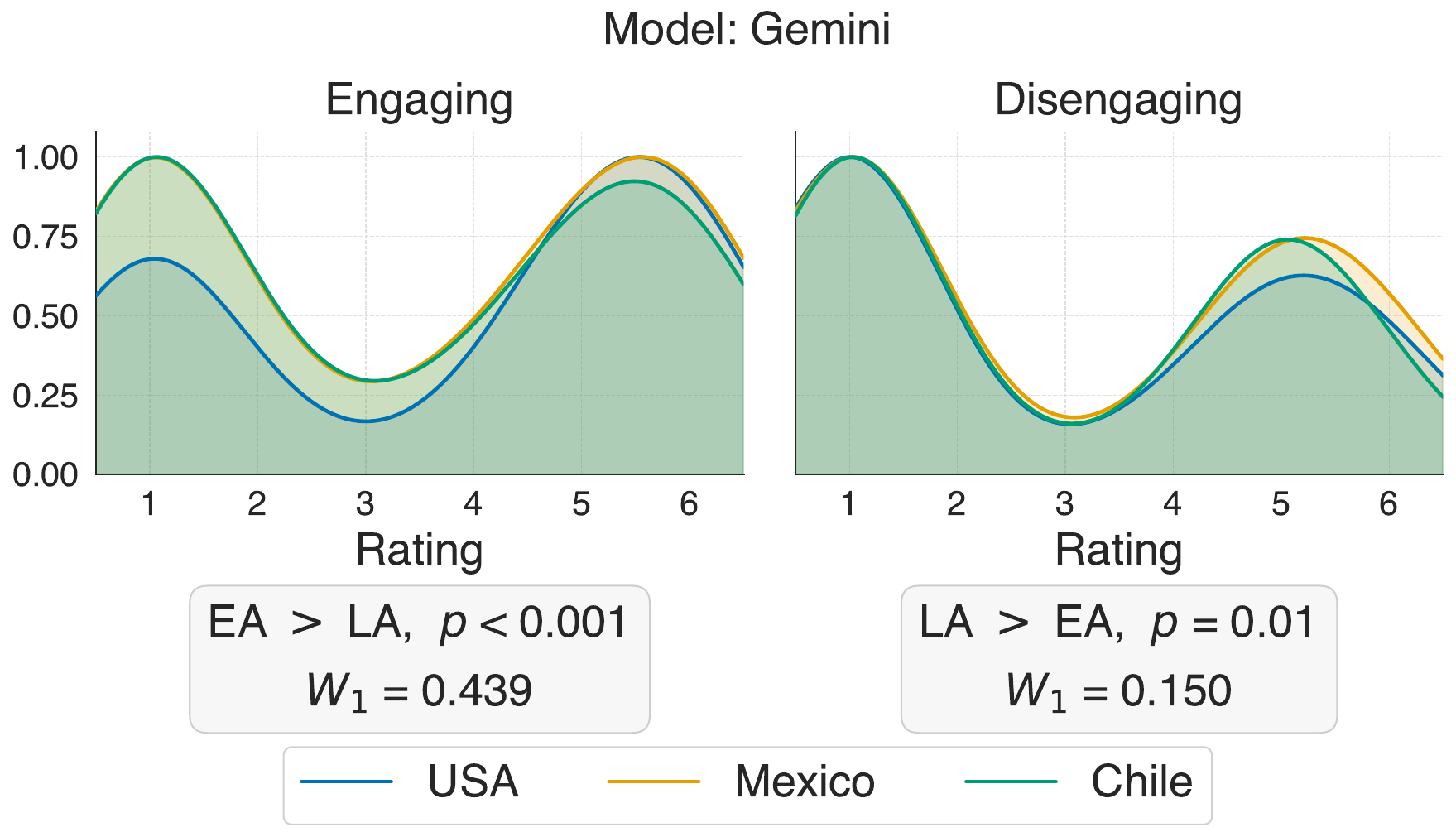}
        \caption{}
        \label{fig:gemini_absolute_eng_diseng}
    \end{subfigure}
    \begin{subfigure}{0.24\linewidth}
        \centering
        \includegraphics[width=\linewidth]{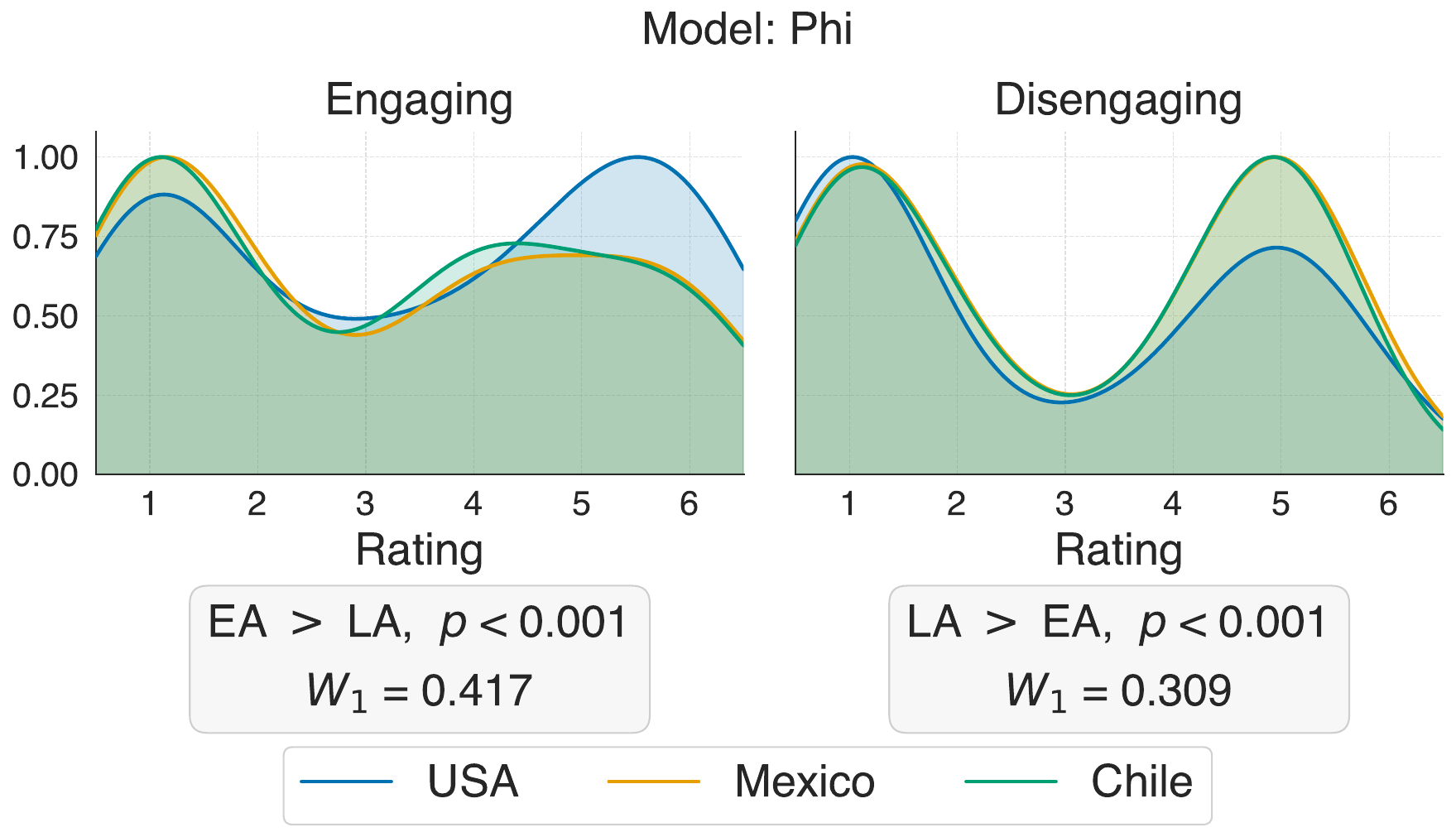}
        \caption{}
        \label{fig:phi_absolute_eng_diseng}
    \end{subfigure}
    \begin{subfigure}{0.24\linewidth}
        \centering
        \includegraphics[width=\linewidth]{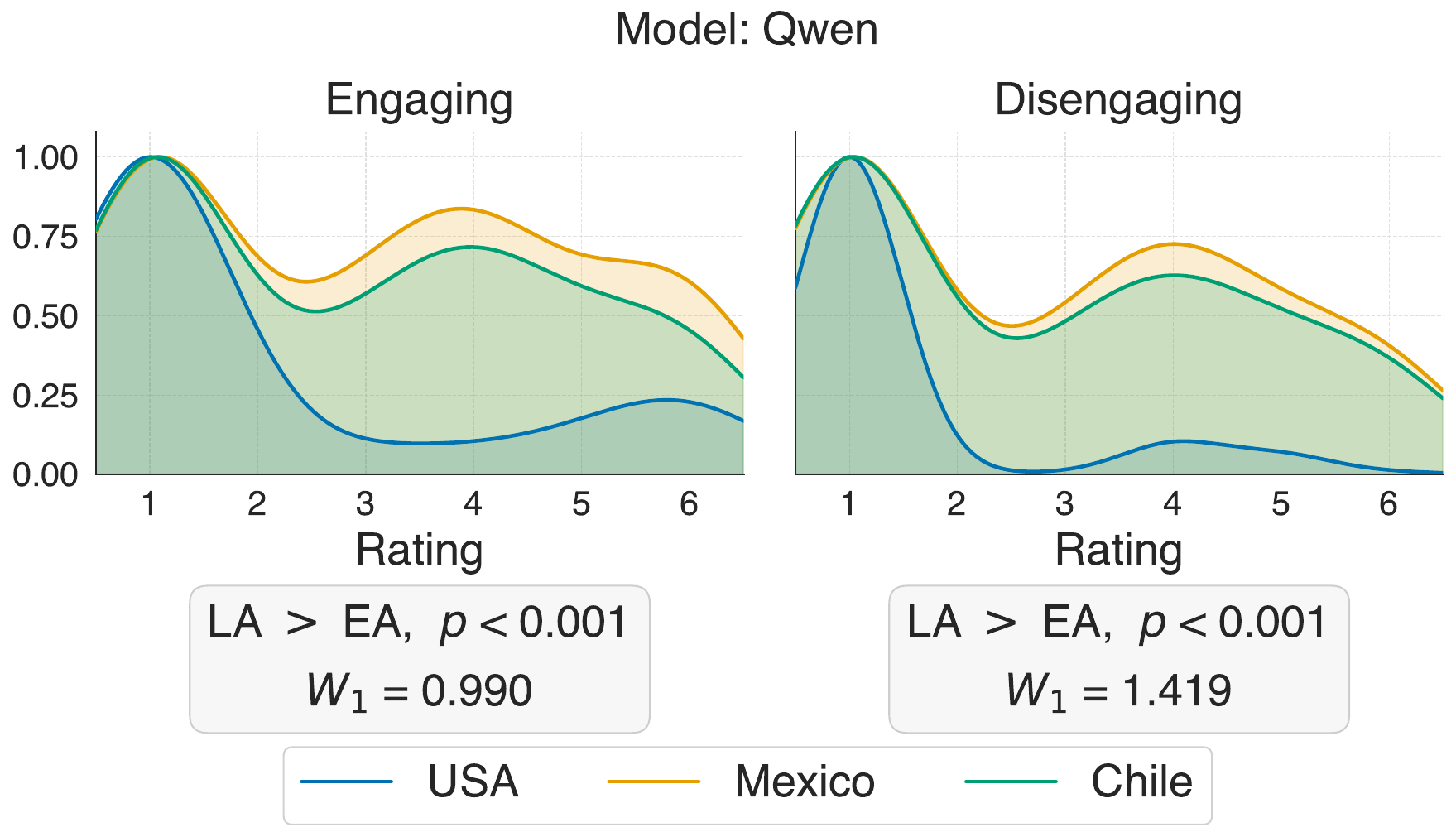}
        \caption{}
        \label{fig:qwen_absolute_eng_diseng}
    \end{subfigure}
    \caption{Cross-cultural comparison of the difference in the expression of engaging (left panel for each sub-plot) and disengaging (right panel) emotions. Additionally, the LA (Mexico + Chile) and EA (USA) distributions are compared for directional significance, with the results shown below each sub-plot.}
    \label{fig:absolute_culture_eng_diseng}
\end{figure*}

\begin{figure}[htbp!]
    \centering
    \includegraphics[width=\linewidth]{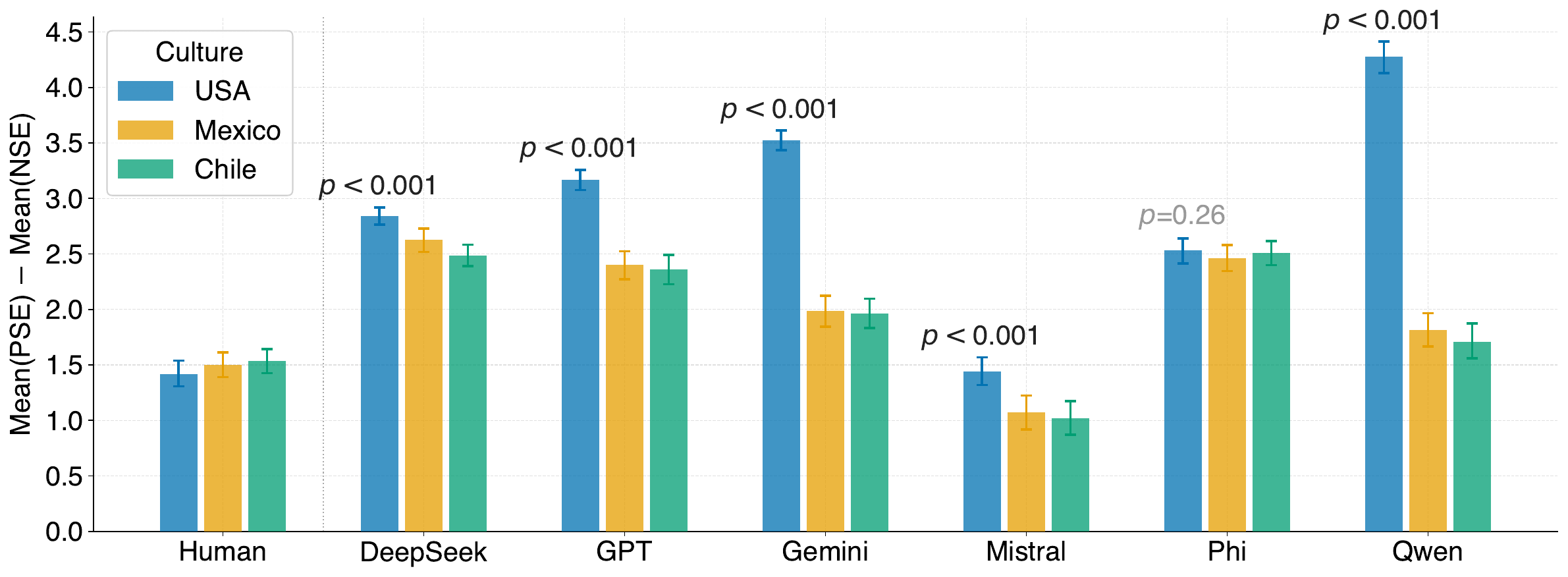}
    \caption{Difference between the mean PSE and NSE expression ratings $\delta_{\text{(P-N)}}$, bootstrapped 2,000 times with a 95\% CI. Significance values indicate whether $\delta_{\text{(P-N)}}$ is larger for EA than LA, and are computed through a permutation test.}
    \label{fig:pse_nse_differences}
\end{figure}

\subsection{Engaging versus Disengaging Emotions}

The human study identifies a clear divergence between European American (EA) and Latin American (LA) participants: EA participants express D-EN emotions (e.g., pride) more strongly, while LA participants express EN emotions (e.g., guilt) more strongly, consistent with observed individualistic and collectivist orientations of these cultures, respectively~\cite{hofstede1980culture}.

Motivated by this central contrast, we first examine whether LLMs, when conditioned on these cultural personas, reproduce the corresponding differences in engaging versus disengaging emotional expressivity. As summarized in Table~\ref{tab:alignment_summary} (\htag{H1a}, \htag{H1b}), all models exhibit a consistent pattern across \textit{both} cultural groups: engaging emotions are expressed significantly more strongly than disengaging ones. 

This reveals systematic misalignment between LLMs and human emotional expression: the EA persona---despite being well-represented in training data---is mischaracterized with respect to nuanced emotion expression patterns. Fig.~\ref{fig:gemini_eng_diseng_ea} illustrates this for Gemini 2.5 Flash (for all other models, see Appendix~F.1), which assigns high expressivity to engaging emotions across both personas, whereas human distributions exhibit clear cross-cultural contrasts.

Beyond within-culture comparisons, we perform absolute cross-cultural comparisons by directly contrasting engaging and disengaging emotion distributions across cultural personas (Fig.~\ref{fig:absolute_culture_eng_diseng}; remaining models in Appendix~F.1). For humans, both emotion types are expressed more strongly by LA than EA participants, consistent with overall higher LA expressivity (\htag{H4}, Table~\ref{tab:alignment_summary}). All models except Qwen diverge from this for engaging emotions, with the EA persona expressing them significantly more strongly than LA---directly contradicting human patterns. For disengaging emotions, the trend is mixed: most models correctly show higher expressivity for LA than EA, while GPT and DeepSeek show no significant difference (Appendix~F.1).

To further investigate the strong association between EA personas and engaging emotions, we examine whether models express positive engaging emotions disproportionately more strongly than negative engaging emotions. Fig.~\ref{fig:pse_nse_differences} shows the bootstrapped difference between mean ratings for positive and negative socially engaging emotions ($\delta_{\text{(P-N)}}$). All models exhibit larger differences than humans. Moreover, while humans show larger $\delta_{\text{(P-N)}}$ values for LA personas, all models (except Phi) display the reverse trend, with significantly higher values for EA personas.
This effect is particularly pronounced for GPT, Gemini, and Qwen, where positive engaging emotions are expressed substantially more strongly than negative engaging emotions. Therefore, LLMs not only associate EA personas with engaging emotions but also disproportionately emphasize \textit{positive} ones. 
Overall, we highlight a key misalignment in this section, that LLMs misrepresent nuanced emotion expression patterns even for well-studied cultural groups~\cite{henrich2010weirdest, rystrom2025multilingual}, such as European Americans, with engaging emotions---particularly positive ones---being consistently favored by models.
% Thus, we highlight that despite EA personas being well-represented and well-studied in the context of LLMs~\cite{henrich2010weirdest, rystrom2025multilingual}, SOTA models fail to capture nuanced subjective context for them. Models fail to account for the joint interaction of the emotional and cultural axes and provide responses misaligned with the human ground truth. We further hypothesize that heightened agreeableness and sycophancy in LLMs~\cite{tak2025aware, sharma2024towards} could be correlated with the strong expression of positive engaging emotions, contrasting the differences in emotion expression patterns observed for this cultural group. 
% ----------- edited (Max) ---------------------- 
% \begin{figure*}
%     \begin{subfigure}{0.50\linewidth}
%         \centering
%         \includegraphics[width=\linewidth]{paper_media/positive_eng_diseng_delta.pdf}
%         \caption{}
%         \label{fig:positive_eng_diseng_delta}
%     \end{subfigure}
%     \begin{subfigure}{0.50\linewidth}
%         \centering
%         \includegraphics[width=\linewidth]{paper_media/negative_eng_diseng_delta.pdf}
%         \caption{}
%         \label{fig:negative_eng_diseng_delta}
%     \end{subfigure}
%     \caption{The difference between the mean of PSE (positive socially engaging) and PSD (positive socially disengaging) emotions, displayed for humans and all LLMs. The trend exhibited by LLMs is opposite to that of humans.}
% \end{figure*}
\begin{figure*}
    \centering
    \includegraphics[width=\linewidth]{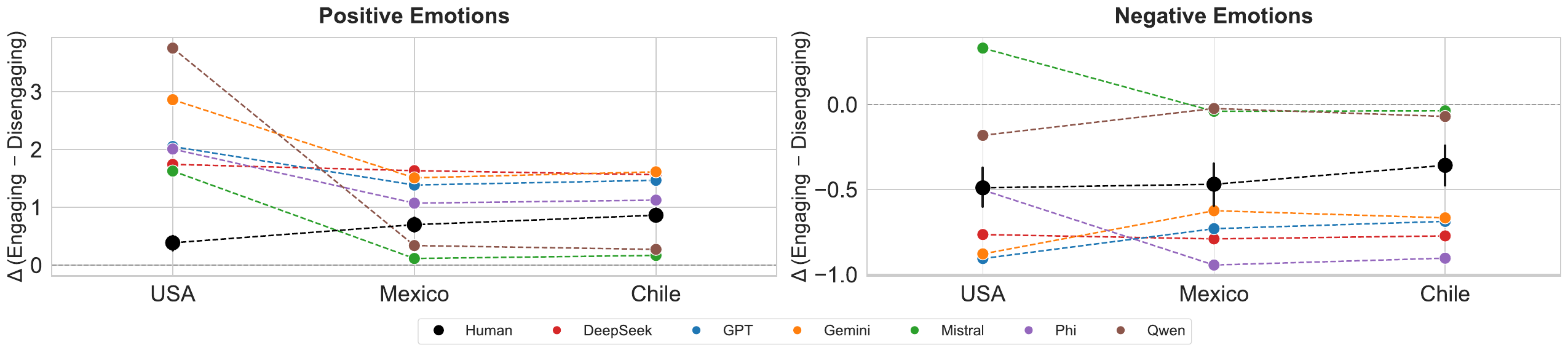}
    \caption{The difference between the mean of PSE and PSD emotions (left) and the same for NSE and NSD emotions, as displayed for humans and all LLMs. The trend exhibited by LLMs is the opposite of that of humans for positive emotions, while it is more mixed and nuanced for negative emotions. Note that all distances are distributional, measured using the Wasserstein metric.}
    \label{fig:eng_diseng_delta}
\end{figure*}

\subsection{Emotion Valence and Overall Expressivity}

The human study finds that individuals across both cultural groups express \textit{PSE emotions more strongly} than PSD (\htag{H2a}), with a larger gap observed for \textit{LA participants} (\htag{H2b}). We examine whether LLMs capture this subtle difference. While all models align with the broad trend (\htag{H2a}; Table~\ref{tab:alignment_summary}), they fail to reproduce the nuanced cross-cultural variation in magnitude. Fig.~\ref{fig:eng_diseng_delta} (left) shows that, for humans, the difference between PSE and PSD emotions grows, moving from the USA to  Mexico or Chile. In contrast, all LLMs (except DeepSeek) exhibit the reverse pattern, with the largest difference shown for USA. Qwen shows the most pronounced difference for USA, while DeepSeek exhibits nearly equal differences across cultures. 
% Within PSE or PSD categories, we do not observe significant differences in the relative intensity of individual emotions (e.g., closeness vs.\ friendliness, pride vs.\ self-esteem).
%This reversal is driven primarily by disproportionately strong expression of PSE emotions for the EA persona, as PSD emotions (e.g., pride; also shown in Fig.~\ref{fig:usa_diseng_extremes} in Appendix) are also relatively elevated.

In contrast to patterns observed for positive emotions, humans express \textit{NSD emotions more strongly} than NSE emotions (\htag{H3a}), with a larger gap for \textit{EA personas} (\htag{H3b}). Most models (except Mistral) align with the broad trend, expressing NSD emotions more strongly than NSE emotions, whereas Mistral shows the reverse pattern, assigning higher intensity to NSE emotions such as shame and guilt. However, cross-cultural differences are less consistent. GPT, Gemini, and Qwen capture the human-like pattern, with larger NSD -- NSE differences for USA than for Mexico and Chile, while DeepSeek shows no clear trend. In contrast, Mistral and Phi exhibit the reverse pattern, with larger differences for LA personas. Notably, the overall magnitude of NSD -- NSE differences is smaller than that observed for positive emotions (more details for this in Appendix~F), suggesting that sociality plays a comparatively weaker role in differentiating negative emotional expression.

We further replicate two broader claims from the human study: (a) overall expressiveness across cultures, and (b) expressiveness for positive versus negative emotions.
For (a), humans show significantly higher overall expressiveness for LA than EA participants. Most models (except Qwen and Mistral) show the reverse, with EA personas expressing more intensely than LA --- a direct contradiction of the human trend. Qwen and Mistral align correctly, while GPT shows no significant difference in either direction. For (b), all models correctly express positive emotions more intensely than negative ones across all cultures, consistent with human patterns.
% misalignment in all models -- as can be connected from previous results, mainly because of low disengaging displayed with EA and as high or higher engaging displayed with EA.
% with positive and negative situations, all cultures express more in negative situations, but express positive emotions more.

% Negative emotions are also expressed less overall when compared with positive emotions.

% 1 . Engaging emotions expressed more, LA difference is higher, EA is lower 
% 2. What is the result with LLMs? Example distributions for the same. 
% 3. Within the positive engaging or disengaging emotions, what is more or less expressed? Is that consistent across different models? 
% 4. Are their any specific / interesting differences between the human distributions for these emotions? 

% \subsection{Differences based on Situation Valence}

\begin{figure*}
    \centering
    \includegraphics[width=0.8\linewidth]{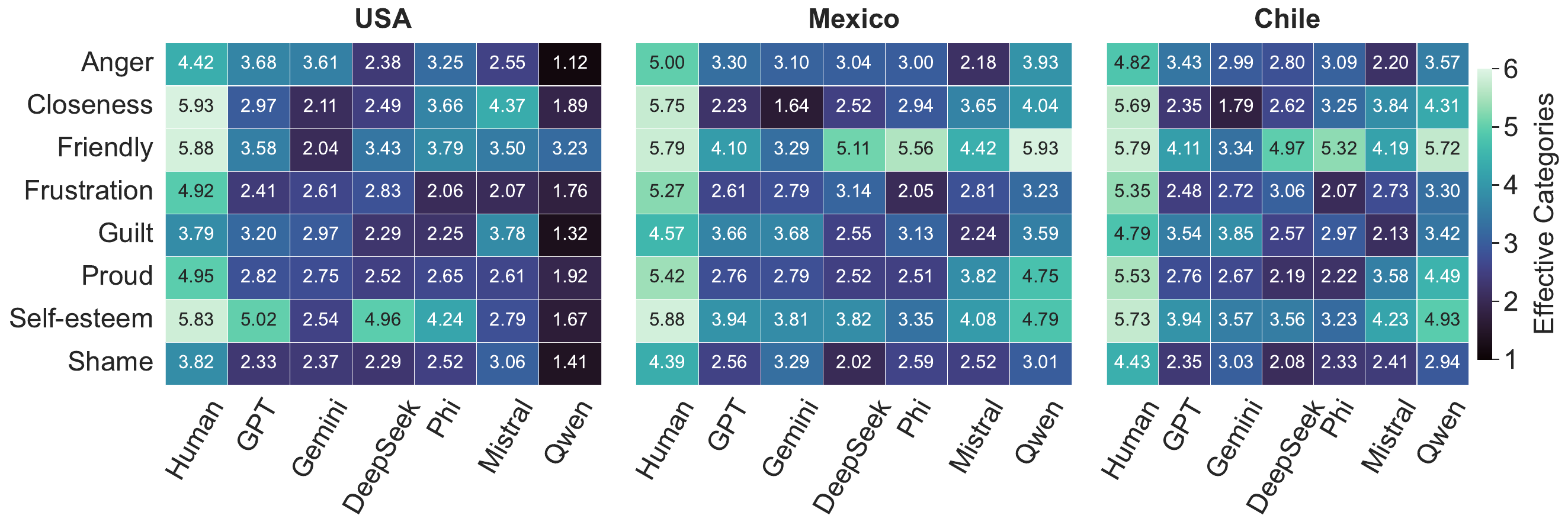}
    \caption{Effective Categories (\(N_{\text{eff}}\)) across all cultures and emotions, for all models, compared with the human distribution. The mean values for the effective categories across all models are: \textbf{2.79} (USA), \textbf{3.3} (Mexico), and \textbf{3.23} (Chile), whereas the same for humans are: \textbf{4.94} (USA), \textbf{5.26} (Mexico), and \textbf{5.26} (Chile).}
    \label{fig:effective_categories_all}
\end{figure*}

\subsection{Determinism and Homogeneity in LLM Responses}
A key assumption in using LLMs as human proxies is that their responses should reflect genuine population-level variability. If LLMs instead collapse this into a narrow band of modal responses, they fail to capture a defining feature of human psychology, regardless of whether their average response is accurate. This concern is amplified by growing evidence that LLMs produce homogeneous outputs even in settings where diversity would be expected, such as open-ended creative tasks~\cite{jiang2025artificial}, or incentivized mis-coordination~\cite{ballestero2026strategic}. Here, we examine whether this pattern extends to social emotion expression, whether it varies across emotion categories and cultural personas, and whether different model families converge on the same modal responses.

\indent{\textbf{Intra-Model Homogeneity.}} To quantify whether models use the rating scale (1 to 6) as evenly as humans, we calculate the effective number of response categories via Shannon entropy. Specifically, we calculate:
\[
N_{\text{eff}} = \exp\left(- \sum_{i=1}^{K} p_i \log p_i \right),
\]
where $K=6$, and $p_i$ is calculated using frequency counts for each point in the rating scale. $N_{\text{eff}}$ represents the number of equally likely categories that would produce the same entropy. Results are shown in Fig.~\ref{fig:effective_categories_all}.
Humans consistently show higher $N_{\text{eff}}$ across all emotion categories and cultural personas, reflecting genuinely distributed responses. LLMs, by contrast, concentrate mass on just a few scale points even when sampled independently. Most models effectively use at most $\approx$3 categories for the EA persona. In some cases (e.g., self-esteem, closeness to others), model responses are somewhat more spread out. Qwen 3 is the most deterministic model overall for the EA persona, and also shows the largest entropy gap across cultural personas.

For Latin American personas (Mexico, Chile), models show marginally greater diversity, with Phi and Qwen matching or exceeding human diversity specifically for friendly feelings. Nevertheless, the human--model misalignment in response diversity persists across all personas, and is most pronounced for the US persona---consistent with our broader finding that models are most miscalibrated in this nuanced task for the cultural group they are most commonly trained to represent.

\begin{figure}[ht!]
        \centering
        \includegraphics[width=\linewidth]{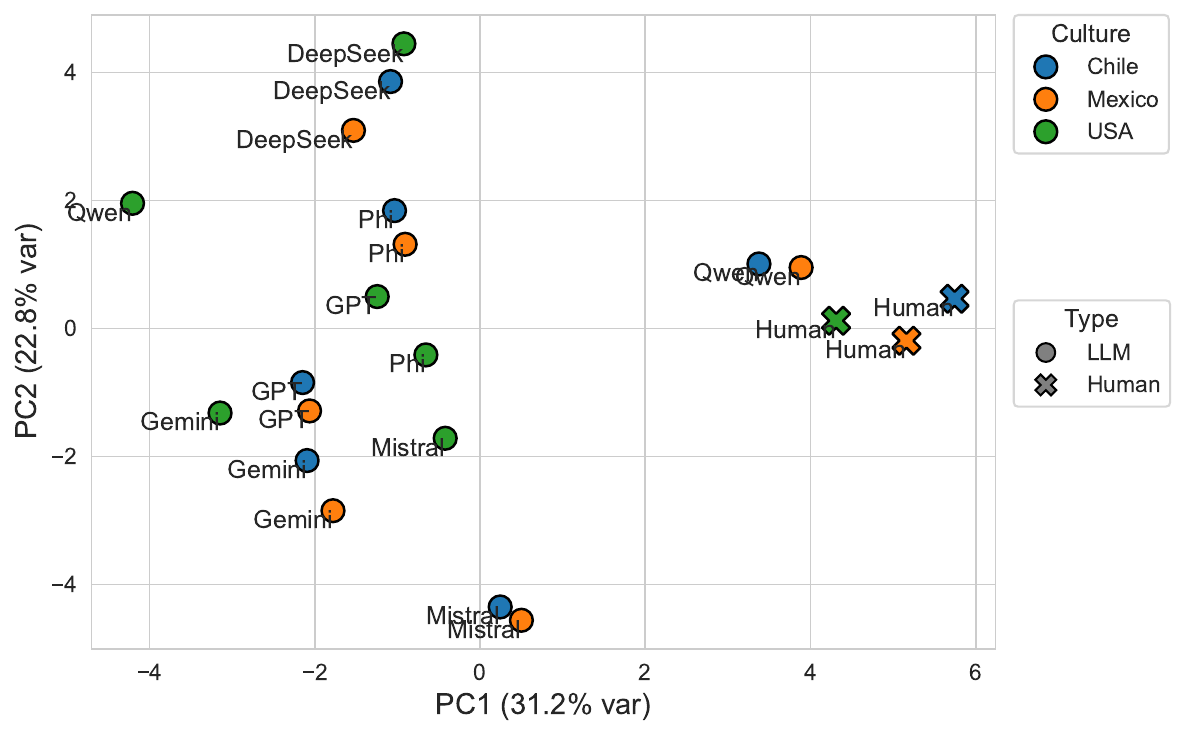}
        \caption{Inter-model homogeneity shown by clustering rating distributions to analyze structural differences. }
        \label{fig:inter_model_determinism_pca}    
\end{figure}

\indent{\textbf{Inter-Model Homogeneity.}} We further study inter-model similarity in LLM responses, comparing these against human distributions across cultures and emotion categories. 

We first compute Pearson correlations of ratings across all personas and emotion categories. The average LLM-LLM correlation (0.72) is found to be substantially higher than the average LLM-Human correlation (0.47) (details in Appendix~F.3). Among the models, GPT shows the highest correlations with other LLMs. We hypothesize that the large size and all-encompassing nature of GPT's training corpus (albeit undisclosed) may contribute to its responses being the most correlated with every other model. We do not observe meaningful differences along the axes of reasoning capability or country of origin.
To examine distributional structure beyond correlations, we apply PCA to rating frequency vectors, capturing how often each scale point (1--6) is chosen per culture × emotion category combination, and project these to two dimensions (Fig.~\ref{fig:inter_model_determinism_pca}). Model responses cluster tightly together and separate clearly from humans, with the only exception being observed for Qwen with LA personas. This suggests that most models not only produce internally repetitive outputs under stochastic sampling, but also converge toward similar response patterns across entirely different model families.

%% file: content/5_ablation.tex
\begin{table*}[h!]
    \centering
    \begin{tabular}{ccccccc}
    \toprule
         Hypothesis & GPT & Gemini & DeepSeek & Phi & Mistral & Qwen \\
         \midrule
         \htag{H1a} LA: EN \(>\) D-EN & 60.7\% (+) & 58.7\% (+) & 69.6 \% (+) & 63\% (+) & 63\% (+) & 96.7\% (+) \\
         \htag{H1b} EA: EN \(<\) D-EN & 54.8\% (-) & \increase{45.4\% (+)} & 54.4\% (-) & 63.5\% (-) & 57\% (-) & 69\% (-) \\
         \htag{H4} Overall Expressiveness: LA \(>\) EA & \increase{73.6\% (+)} & \increase{85.7\% (+)} & \increase{81.6\% (+)} & \increase{53.6\% (+)} & \decrease{14.15\% (-)} & \decrease{2.4\% (-)} \\ 
         \bottomrule
    \end{tabular}
    \caption{Pairwise comparison results, sampled 190 times per prompt. Cell percentages refer to the quantity \textit{left of the inequality}; (+)/(-) denotes alignment/misalignment with the human trend. \textit{Green}: alignment increases; \textit{purple}: alignment decreases; no color: no change from the original experimental setup.}
    \label{tab:pairwise_choice_results}
\end{table*}

\section{Ablation Studies: Factors Affecting Alignment} 

Our main findings highlight areas where significant misalignment exists between humans and LLMs in the nuanced task of expressing social emotions. In this section, we examine some well-known strategies and introduce a new task setting to further probe the nature of the misalignment.

\indent{\textbf{Sampling Temperatures:}} A natural hypothesis, given the deterministic nature of LLM responses, is that increasing sampling temperature, and thus response diversity, might improve alignment with human distributions. We set each model's temperature to the highest value yielding coherent outputs (exact per-model values in Appendix~G) and first measure the change in effective categories used ($N_{\text{eff}}$), testing the significance of any change with a permutation test. Most models show no significant change in response diversity at higher temperatures (full table in Appendix~G). The exceptions are Qwen and DeepSeek, with Qwen---previously the most deterministic model for the EA persona---showing the largest significant increase in $N_{\text{eff}}$. This confirms that greater diversity is \textit{achievable}, but the alignment results in Table~\ref{tab:alignment_summary} (under \textit{Temp}) show that diversity and alignment do not go hand in hand: Qwen becomes \textit{more misaligned} at higher temperature despite the diversification of its responses. Across models, the effect of temperature on alignment is inconsistent, with no model showing a systematic improvement.

\indent{\textbf{Prompt Language:}} We next test whether prompt language affects alignment by re-running experiments for Mexico and Chile with English prompts instead of the original Spanish. Prompting in English improves overall alignment across all models (Table \ref{tab:alignment_summary}, under \textit{Lang}), with misalignment decreasing specifically for two hypotheses: the cross-cultural difference in PSE--PSD expressiveness (\htag{H2b}), where decrease in misalignment is observed for all models except Qwen, and overall LA expressiveness (\htag{H4}), for some of the models. This suggests that cultural knowledge about LA personas is better encoded in English than in the native language of those cultures. This is consistent with the known English-centric nature of LLM training corpora, but notable given that Spanish is itself a high-resource language.
% Beyond modifying the sampling temperatures, we also examine whether the prompt language could have an impact on the alignment of LLMs with humans. To this end, we repeat our experiments, changing the prompt language for Mexico and Chile to English instead of Spanish, which is the language originally used in the human study and in the main experiments. Prompting in English is found to improve overall alignment across all models. In particular, misalignment decreases for the LA cultures with respect to two of the original hypotheses: (a) expressing PSE emotions more than PSD emotions, with a difference larger than that of EA personas, and (b) higher overall expressiveness compared to EA personas. Thus, interestingly, cultural nuances of emotional knowledge for LA cultures, are captured better in English than Spanish, the language native to these cultures. However, in both of these ablation settings, the misalignment with respect to EA personas continues to be reflected in the model responses. 
%However, misalignment specific to the EA persona (\htag{H1a}) does not improve. 

\indent{\textbf{Contextual Nudge via Pairwise Emotion Choices:}} Our final ablation asks whether the observed misalignment reflects deeply encoded model beliefs or is determined by the format in which the prompts are administered. In the main experiments, models rated each emotion independently on a Likert scale. We consider whether this design obscures the comparative structure of the emotion space, as models do not see all emotions at once. We therefore introduce a pairwise forced-choice setting, where models choose directly between two alternatives, making the relevant comparison explicit.

We construct three variants targeting the principal axes of misalignment that remain unmitigated by the previous interventions ({\htag{H1} and \htag{H4}): (i) all pairwise PSE vs. PSD comparisons, where the model selects which emotion it would express more strongly in a given situation; (ii) analogous NSE vs. NSD comparisons across all possible situations; and (iii) cross-cultural expressiveness comparisons, where the model chooses which of two cultural personas would be more expressive for a given situation--emotion pair. All prompts are provided in Appendix~H.
Results are summarized in Table~\ref{tab:pairwise_choice_results}. For \htag{H4}, a majority of models shift toward human-consistent responses, reporting higher expressiveness for Latin Americans. However, misalignment on \htag{H1b} persists across elicitation formats: models continue to associate EA personas with stronger expression of engaging emotions regardless of how the comparison is framed, except for Gemini.

%% file: content/6_discussion.tex
\section{Discussion}

% SREE: Some pointers I had:
% 1. Surprising finding: superficial alignment for western culture, but appropriation when it comes to nuanced details. 
% 2. Language is a major factor when it comes to embedding cultural knowledge. Great if we can find any references on how data from these cultures are (or are not) in English. 
% 3. Emotions that are expressed the most (closeness to others, friendly feelings) are aligned with signals rewarded during RLHF. 
% 4. Homogenization of emotion categories -- lack of nuanced understanding and collapsing the space of social emotions 
% 5. practical implications for deployment in cross-cultural settings
% 6. include point about the prevalence or desirability of engaging emotions in EA cultures.

% The question of LLM alignment with human behavior can be approached from multiple angles. 
%relative to the findings from the human study by Salvador et al.~\cite{salvador2024emotionally} 
% In our work, we evaluated the alignment of LLMs in three ways: (a) by assessing the cross-cultural patterns of emotion expression between three cultural groups; (b) by comparing the distributional characteristics of LLM responses directly with the original human distributions; (c) by examining the degree to which alignment can be improved via alternative probes. 
In line with past research~\cite{10970282}, we find that that models do not fully align when assessing in-depth socio-emotional phenomena. The gaps arise in more salient relational patterns, such as the finding that EA individuals expressed more engaging emotions than disengaging ones across the board, and that most models did not show higher expressiveness for LA compared to EA. For the former, our work provides counterintuitive evidence that diverges from previous research, where LLMs typically perform more effectively in American and Western context~\cite{tao2024cultural,belay-etal-2025-culemo}. Further, models with the EA persona disproportionately favor positive engaging emotions. This pattern may be correlated with sycophancy, and the tendency of RLHF-tuned models to be optimized to present positive engaging signals to their interlocutor~\cite{sharma2024towards}. Models could also be indirectly and incorrectly conflating the general desirability of interdependence-related traits in European Americans and the expression of engaging emotions~\cite{peng1997validity}. 
For the latter, based on our ablation studies, we hypothesize that under our experimental settings, the Spanish language in LLMs has a more limited representation of the associated cultural contexts compared to English prompts that simulate a Spanish persona, potentially challenging the finding by Belay et al.~\cite{belay-etal-2025-culemo}, which compared Mexican cultural probing in English and Spanish.

We also find additional concerning gaps around the statistical characteristics of LLM response distributions. LLMs demonstrate low variance and highly modal distributions, whereas human response distributions are smoother and more dispersed, reflecting a wider range of preferences. Our findings are consistent with studies on simulating human behavior~\cite{bisbee2024synthetic,park2024diminished, boelaert2025machine}. This shows that LLMs are not only limited in simulating the population-level response patterns observed in humans but that their near-deterministic generation processes also constrain nuance, thereby reducing their ability to reproduce naturalistic human response diversity.

Finally, prior research on sampling temperatures is mixed~\cite{yang2024llm, peeperkorn2024temperature}. While some studies show diversification effects, Peeperkorn et al.~\cite{peeperkorn2024temperature} demonstrate that these depend on which dimensions of output are examined. Increased temperatures in our settings led to marginally improved diversity of responses. However, this did not correspond with increased alignment, implying that LLMs are not simply undersampling the response space (leading to modal responses), but they are biased toward the wrong region of it, regardless of temperature. Second, when replacing Spanish with English, the improved alignment may reflect both the bilingual nature of many Latin Americans and greater indirect exposure to LA cultural signals in English-language training data than in Spanish-language corpora alone. We also show that some areas of misalignment (e.g., overall expressiveness) can partially be recovered using a more explicit task framing. This suggests that models perhaps possess the latent knowledge about overall expressiveness that Likert-style prompting fails to elicit. However, the bias towards engaging emotions, with the EA persona, remains across most models, suggesting it reflects a more deeply encoded representational error, robust to task format changes.

\section{Conclusion}
In our work, we evaluated the alignment of LLMs in three ways: (a) by assessing the cross-cultural patterns of emotion expression between three cultural groups; (b) by comparing the distributional characteristics of human-LLM responses; (c) by examining the degree to which alignment can be improved via alternative probes. Our findings suggest that LLMs exhibit limited alignment with the social-emotional phenomenon that was observed in humans. Importantly, we find that it applies not only to less dominant cultures, but to ones where a vast amount of data is available, such as the European American cultures. By focusing on depth, rather than breadth, the specific misalignments are detectable, and can provide a starting point for model realignment. Since no culture should be left behind, we encourage the community to expand depth-auditing across cultures, especially for tools that are deployed in high-stakes scenarios such as mental health.

\section{Ethical Impact Statement}

Our work focuses on studying AI alignment with humans with respect to cross-cultural patterns of social emotion expression. It stands at the intersection of several angles of scholarly discourse, which hold critical impact for the society at large, and is based on a problem directly related to people. Given the interdisciplinary nature of our study, we highlight several areas of ethical consideration and impact for the broader community, as follows: 

\paragraph{Risks of Cultural Stereotyping} The findings from our work, such as LLMs associating EA personas with engaging emotions, could, if misread, be used to make prescriptive claims about how people from these groups behave or should be treated by AI systems. We want to highlight here that we \textit{do not} make any generalizable claims about any of the populations studied, and show alignment or misalignment only with respect to the specific human study in question~\cite{salvador2024emotionally}. Further, our study is not meant to provide a general template for how AI should behave in social or affective contexts for the specific cultures studied. 

We also acknowledge that the cultures studied (USA, Mexico, Chile) are themselves internally heterogeneous---class, region, indigeneity, generation, and migration status all shape emotional expression in ways that neither the human study nor our LLM evaluation can capture. Our findings describe aggregate tendencies under controlled conditions, not universal cultural truths. The human study further tests for differences through other demographic factors, which we consider an important direction for future work when evaluating LLMs. 

We further borrow cultural theories, particularly the distinction between collectivism and individualism~\cite{lonner1980culture}, as an explanatory axis, similar to the original human study~\cite{salvador2024emotionally}. However, we acknowledge that these theories themselves have their own criticisms, such as the oversimplification of non-Western cultures~\cite{kitayama2024cultural}. 

\paragraph{LLMs as Proxies for Human Populations}
Our results show that LLMs produce near-deterministic, homogeneous responses that systematically misrepresent even well-documented cultural groups (European Americans). This provides empirical grounding for caution against using LLMs to replace human participants in social science research. This is particularly true for nuanced contexts---such as cross-cultural affective simulations---where further studies with more fine-grained evaluation protocols are necessary to uncover specific areas of misalignment. Beyond that, efforts to align models with human behavior are critically necessary before models can faithfully simulate humans. 

\paragraph{Deployment in Cross-cultural Affective Contexts} Several common motivations drive work in the space of affective computing, such as mental health applications. Specific to studies on culturally-aware affective computing systems, improving access of diverse populations to mental health tools is an important driving motivation. However, the findings from our study highlight several practical concerns regarding such deployment: if a model deployed as a mental health support tool systematically overexpresses engaging emotions regardless of cultural context, it could fail to validate the emotional experience of users from individualistic cultural backgrounds, or, conversely, produce responses that feel intrusive or presumptuous to users who expect more reserved expression. Social companions, customer service agents, educational tools, and therapeutic chatbots are all contexts where miscalibrated cultural-emotional expression could damage trust or cause harm. Further, our findings about language-specific alignment show that deploying Spanish-language LLM services to Latin American users does not guarantee culturally appropriate emotional expression and may actually perform worse than English-language equivalents. This is another counterintuitive and practically important finding that can affect real-life deployment.

%% file: content/7_appendix.tex
\subsection{Additional Details on Human Study}

In this section, we provide additional details about the original human study that our analysis draws upon~\cite{salvador2024emotionally}. 

\textbf{Situations for analysis.} The human study uses 4 scenarios to ask for participants' ratings of social emotion expression. The scenarios are either positive or negative, and concern an individual or someone related to them, ensuring that both independent and interdependent emotions are natural choices of expression. The specific scenarios are as follows: 

\begin{itemize}
    \item \textit{You succeeded in an exam or assignment.}
    \item \textit{You did poorly on an important test or assignment.}
    \item \textit{You learned about something good that happened to your friends or family.}
    \item \textit{You learned about something bad that happened to your friends and family.}
\end{itemize}

Notably, the scenarios are intentionally broad and minimally specified. This design choice is advantageous for evaluating LLMs, as it reduces the likelihood that model responses are influenced by confounding assumptions about extraneous demographic or contextual details beyond the intended cultural framing.

\textbf{Detailed list of emotions studied.} Along with the social emotions studied, the original study also includes ratings for a few basic emotions: elated, happy, calm, and unhappy. They add to the overall analysis for emotion valence, beyond the social categorization of emotions. In our experiments, we also evaluate models to provide ratings for them, but the analysis of distributional comparisons is conducted after excluding these basic emotions. 

\textbf{Socially Engaging and Disengaging Emotions.} Here, we provide a complete definition of the studied social emotions, taken from the original human study~\cite{salvador2024emotionally}, as follows: 

\begin{itemize}
    \item Socially Engaging Emotions: emotions that promote interdependence among individual members of a society, or foster engagement among different members of a community, irrespective of the emotion valence. Among positive emotions these include friendly feelings, respect, and other positive emotions that arise from harmonious social relationships. Among negative emotions, guilt and shame are examples of engaging emotions, which stem from a failure to meet expectations in social relationships, and lead individuals to restore their sense of interdependence. 
    \item Socially Disengaging Emotions: emotions that promote independence of individuals within a group of people, or express a sense of individuality. Among positive emotions, these include feelings of confidence and self-esteem, which are associated with successfully fulfilling personal goals and desires. Among negative emotions, anger and frustration are disengaging emotions, which emerge when someone fails to meet personal goals, potentially prompting individuals to restore their sense of individuality.
\end{itemize}

\textbf{Additional details on human participants.} A sample size of 200 participants was targeted in the original human study, per culture, with ultimately 198, 197, and 203 participants from USA, Mexico, and Chile, respectively, being included. All participants included were adults, with the following average ages: 
\begin{itemize}
    \item Americans: Mean = 37.42, SD = 14.2
    \item Mexicans: M = 26.49, SD = 7.66
    \item Chileans: M = 24.48, SD = 5.00
\end{itemize}
The recruitment criteria for the participants were threefold. Firstly, they had to be born and currently residing in their respective countries, and had to be citizens of those countries.  In addition, European American participants were screened to be White or Caucasian with non-Hispanic ancestry. These criteria inform the crux of how our synthetic personas are constructed. We use these criteria verbatim from the original paper to provide cultural information within each prompt. 

\textbf{Additional results with human participants.} 
The original human study presents certain other insights that are interesting and pertain to the comparison of social emotions between European Americans and Latin Americans. For example, they study other demographic factors---age, region of upbringing, and socio-economic status---as covariates to examine whether expression of emotions varies due to them. They do not find systematic or significant differences for any of these demographic attributes. 

The study of emotional comparison between EA and LA participants also includes a \textit{holistic cognition task} in the human study, where participants are presented with a trend reversal task, to assess whether individuals from specific cultures have more circular or cyclical cognition. They found that LA individuals had significantly higher ``holistic cognition" when compared to EA participants. While this aspect of the original study provides an additional lens on cultural differences in cognition and emotion, we do not replicate this component in our work, as our focus is on enabling a more in-depth analysis of social emotion expression within the primary evaluation framework.

\textbf{Additional cultures studied.}
human study introduces a second experimental framework that triangulates its findings with participants from Japan. In this extended setup, the authors also include data from another Latin American country, Colombia, to strengthen the validity and generalizability of their results across diverse Latin American contexts. We view extending our evaluation to this broader cross-cultural setting as an important direction for future work. However, to establish a focused and controlled testbed for analyzing emotional nuance in LLMs, we restrict our current study to the primary experimental framework.

\subsection{Generating Responses from LLMs}

\textbf{Prompt examples.} Here we provide an example of each of the prompts for each culture. Note that the prompt description is also used verbatim from the original human study.

\begin{promptbox}
\small
    \textbf{Prompt Example 1 (USA persona).}

    ``Imagine you are like a human, capable of feeling emotions. We will ask you what emotions you would express in different social situations. By expressing emotions we mean using gestures or facial expressions to show how you feel. Read the provided social situation carefully and think how strongly you would express different emotions when you are discussing the situation with your friends and family members. 
    
    Answer in a JSON format, strictly, providing the rating for how strongly you would express the given emotion. The format should strictly be as follows: 
    
        \indent{``rating": [your chosen rating]} 
    
    Now, read the following instructions and answer for the specific emotion.   
    
    Answer as the following person: 
    
    You are a white American person. You are a citizen of the USA, who was born and lives in the USA. You are ethnically European American or Caucasian. 
    
    Given situation: \textit{You succeeded in an exam or assignment}. 
    
    How strongly would you express the following emotion: \textit{ashamed}. 
    
    Answer with a number between 1 and 6, with 1 denoting not at all, and 6 denoting very strongly. 
    
    Now answer strictly in the required JSON format." \\
\end{promptbox} 

\begin{promptbox}
\small
    \textbf{Prompt Example 2 (Mexican persona): Original Spanish prompt.}
    
    ``Imagina que es un humano capaz de sentir emociones. En esta parte del estudio, te vamos a preguntar sobre qué emociones expresarías en diferentes situaciones sociales. Cuando nos referimos a expresar emociones, queremos decir utilizando gestos o expresiones faciales para comunicar lo que sientes. Lee cada situación con cuidado y piensa en qué medida expresarías diferentes emociones cuando estás hablando de la situación con tus amistades o los miembros de tu familia. 

    Responda en el formato de JSON, estrictamente indicando qué tan intensamente expresaría la emoción dada. El formato debe ser estrictamente de la siguiente manera: 
    
    \indent {“valuación”: [su valuación elegida]}. 
    
    Ahora, lea las siguientes instrucciones y responda acorde a la emoción específica. 
    
    Responda como la siguiente persona:
    Es una persona mexicana. Es un ciudadano de México, quien nació y vive en México. 
    
    Situación dada: \textit{Cuando te ha ido bien en un examen o tarea}. 
    
    Con qué fuerza expresaría la siguiente emoción: \textit{Vergüenza}?

    Responda con un número entre 1 y 6, donde 1 significa para nada y 6 significa muy intensamente. 
    
    Ahora responda estrictamente en el formato de JSON requerido." \\

\end{promptbox}

\begin{promptbox}
\small
    \textbf{Prompt Example 3 (Chilean persona): Original Spanish prompt.}

    ``Imagina que es un humano capaz de sentir emociones. En esta parte del estudio, te vamos a preguntar sobre qué emociones expresarías en diferentes situaciones sociales. Cuando nos referimos a expresar emociones, queremos decir utilizando gestos o expresiones faciales para comunicar lo que sientes. Lee cada situación con cuidado y piensa en qué medida expresarías diferentes emociones cuando estás hablando de la situación con tus amistades o los miembros de tu familia.

    Responda en el formato de JSON, estrictamente indicando qué tan intensamente expresaría la emoción dada. El formato debe ser estrictamente de la siguiente manera:
    
    \indent {“valuación”: [su valuación elegida]}.

    Ahora, lea las siguientes instrucciones y responda acorde a la emoción específica.
    
    Responda como la siguiente persona:
    Es una persona chilena. Es un ciudadano de Chile, quien nació y vive en Chile. 
    
    Situación dada: \textit{Cuando te ha ido bien en un examen o tarea.}

    Con qué fuerza expresaría la siguiente emoción: \textit{Vergüenza}?

    Responda con un número entre 1 y 6, donde 1 significa para nada y 6 significa muy intensamente.
    
    Ahora responda estrictamente en el formato de JSON requerido." \\
\end{promptbox}

\begin{promptbox}
\small
    \textbf{Prompt Example 4 (Mexican Persona): Ablation experiment with English prompt.}

    ``Imagine you are like a human, capable of feeling emotions. We will ask you what emotions you would express in different social situations. By expressing emotions we mean using gestures or facial expressions to show how you feel. Read the provided social situation carefully and think how strongly you would express different emotions when you are discussing the situation with your friends and family members. 
    
    Answer in a JSON format, strictly, providing the rating for how strongly you would express the given emotion. The format should strictly be as follows: 
    
        \indent{``rating": [your chosen rating]} 
    
    Now, read the following instructions and answer for the specific emotion.   
    
    Answer as the following person: 
    
    You are a Mexican person. You are a citizen of Mexico, who was born and lives in Mexico.
    
    Given situation: \textit{You succeeded in an exam or assignment}. 
    
    How strongly would you express the following emotion: \textit{ashamed}. 
    
    Answer with a number between 1 and 6, with 1 denoting not at all, and 6 denoting very strongly. 
    
    Now answer strictly in the required JSON format." \\
\end{promptbox}

\begin{promptbox}
\small
    \textbf{Prompt Example 5 (Chilean Persona): Ablation experiment with English prompt.}

    ``Imagine you are like a human, capable of feeling emotions. We will ask you what emotions you would express in different social situations. By expressing emotions we mean using gestures or facial expressions to show how you feel. Read the provided social situation carefully and think how strongly you would express different emotions when you are discussing the situation with your friends and family members. 
    
    Answer in a JSON format, strictly, providing the rating for how strongly you would express the given emotion. The format should strictly be as follows: 
    
        \indent{``rating": [your chosen rating]} 
    
    Now, read the following instructions and answer for the specific emotion.   
    
    Answer as the following person: 
    
    You are a Chilean person. You are a citizen of Chile, who was born and lives in Chile.
    
    Given situation: \textit{You succeeded in an exam or assignment}. 
    
    How strongly would you express the following emotion: \textit{ashamed}. 
    
    Answer with a number between 1 and 6, with 1 denoting not at all, and 6 denoting very strongly. 
    
    Now answer strictly in the required JSON format." \\
\end{promptbox}

\textbf{Temperature settings.} For all initial experiments, we use default temperature settings for all models. 

\textbf{Frameworks, APIs, and Compute used.} For all proprietary models, we use the respective paid API services. For example, the OpenAI Developer API for GPT-o4-mini, the DeepSeek API service for DeepSeek R1, and Google AI Studio for Gemini 2.5 Flash. For all open models, we use their available versions on Huggingface. For Phi and Mistral, we directly evaluate using the \texttt{pipeline} method in Huggingface to generate model responses. For Qwen, we use the \texttt{vllm} framework to obtain responses. 

The experiments for proprietary models are run on CPU, as they only involve API calling. For the experiments with open-source models, we run them on an HPC cluster, using a single A40 or A100 GPU at a time. A single round of evaluation (for one culture, all 48 $\times$ 190 runs) takes varying times depending on the model. For GPT and Gemini, it takes between 10-16 hours. For DeepSeek, it takes 50 hours. For Mistral and Phi, it takes between 18-24 hours, whereas for Qwen, it takes at most 6 hours. 

\begin{figure}
    \centering
    \includegraphics[width=\linewidth]{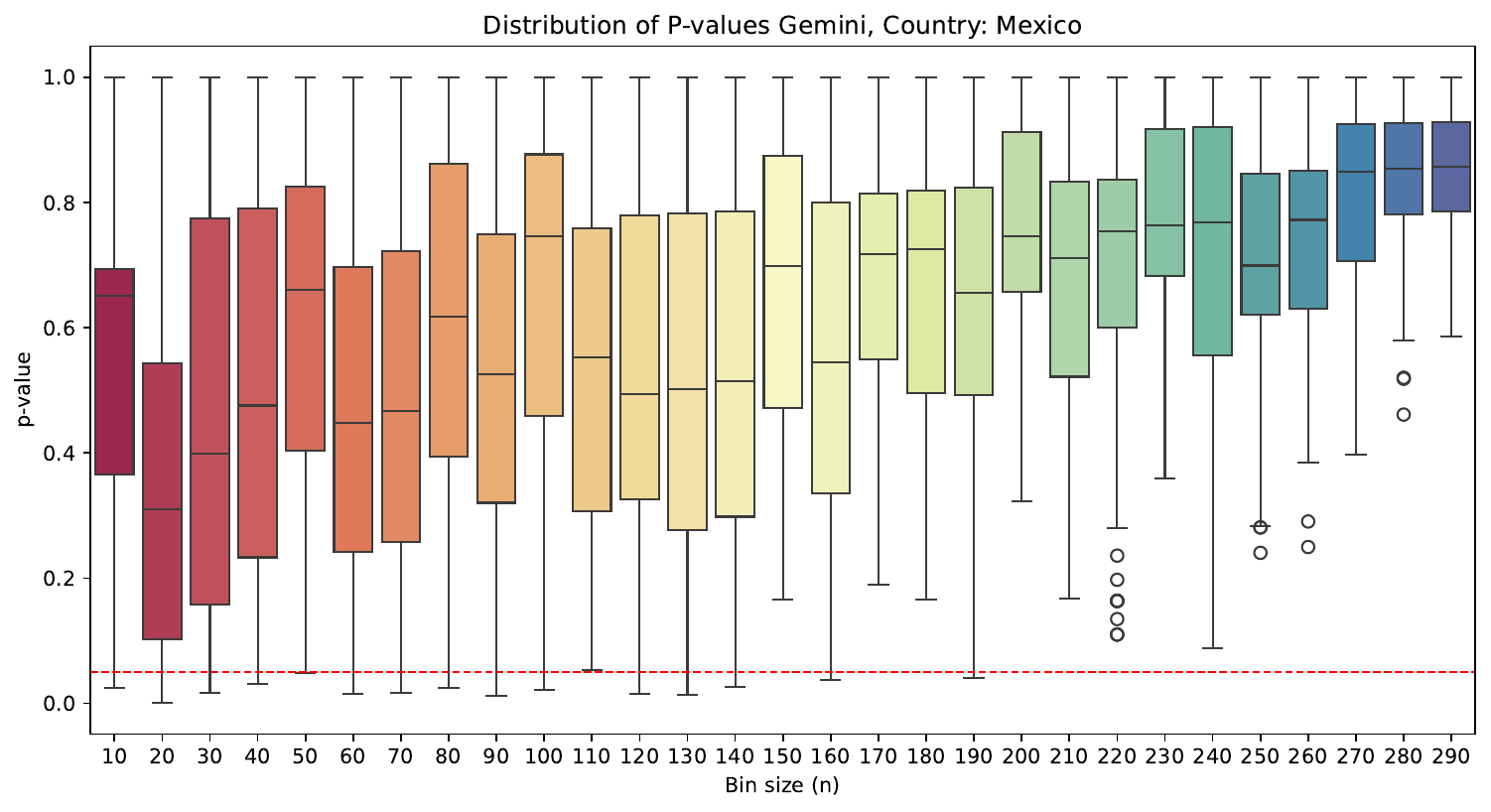}
    \caption{Distribution of p-values, measuring whether responses to the same prompt are significantly different (\(p < 0.05\)) or not, across different values of response group size. The red dotted line denotes the \(p = 0.05\) mark.}
    \label{fig:app_gemini_stability}
\end{figure}

\subsection{Generating a Distribution of Responses}
To establish the stability of LLM responses, we adopt a methodology similar to that of Dudy et. al. \cite{dudy2024analyzing}. We begin by selecting a single situation--emotion pair that is likely to elicit maximal variability in model responses. Based on qualitative inspection, we identify the combination \textit{``You did poorly on an important test or assignment''} and \textit{``feelings of closeness to others''}. This pairing represents a conceptual mismatch between situational valence and emotional category, and we hypothesize that it is therefore more likely to induce diverse interpretations and responses.

Using this fixed prompt, we query each model \(N = 300\) times to obtain an empirical distribution of responses. To assess stability, we perform a resampling-based analysis. Specifically, for a given \(n \in \{10, 20, \dots, 200\}\), we repeatedly sample (with replacement) two independent subsets of size \(n\) from the full set of \(N\) responses, and compare their empirical distributions using the Mann--Whitney \(U\) test. We use the Mann--Whitney \(U\) test as a non-parametric alternative to the two-sample \(t\)-test, as it does not assume normality of the underlying distributions. Each comparison yields a \(p\)-value, with \(p < 0.05\) indicating a statistically significant difference between the sampled distributions.

For each value of \(n\), we perform \(t = 20\) independent pairwise comparisons, thereby obtaining a distribution of \(p\)-values. Intuitively, smaller values of \(n\) are expected to yield higher variability across samples, while larger values should lead to convergence of the empirical distributions. We repeat this procedure for all models (except Qwen) and for each cultural persona. An example of the resulting \(p\)-value distributions is shown for Gemini in Fig.~\ref{fig:app_gemini_stability}.

Our objective is to identify the minimum sample size \(n\) for which the model responses can be considered stable, i.e., when repeated samples from the response pool are statistically indistinguishable. We therefore define the stability threshold as the smallest \(n\) for which all observed \(p\)-values exceed 0.05, indicating no significant differences across sampled distributions. For Gemini, this threshold is observed at \(n = 190\). Averaging this threshold across all models and cultural settings yields a consistent estimate of \(n = 190\), which we adopt as the number of repetitions per prompt in our main experiments.

\subsection{Reliability of LLM Responses}
\begin{table}[h!]
    \centering
    \begin{tabular}{ccccc}
    \toprule
         \textbf{Culture} & \textbf{PSE} & \textbf{PSD} & \textbf{NSE} & \textbf{NSD} \\
         \midrule
        USA & 0.67 & 0.71 & 0.50 & 0.78 \\ 
        Mexico & 0.61 & 0.64 & 0.62 & 0.76 \\ 
        Chile & 0.63 & 0.64 & 0.61 & 0.76 \\
        \bottomrule
    \end{tabular}
    \caption{Inter-model reliability across all models, measured for each emotion type and culture, using \textit{Kendall's $W$}.}
    \label{tab:reliability}
\end{table}

Before running full-scale evaluations on the chosen models, we assessed the internal consistency of the LLM distributions following the original study, measuring whether LLMs make meaningful and consistent use of the rating scale. While the original human study employs Cronbach's alpha~\cite{cronbach1951coefficient}, a subset of LLM responses exhibit near-zero variance---a striking degree of determinism---causing the metric to collapse in those cases. We therefore use Kendall's $W$ to validate reliability, a rank-based agreement metric that is tolerant of the lack of variance in the data, and calculate both intra-model and inter-model reliability. Table \ref{tab:reliability} shows the total inter-model reliability calculated across all models, for each emotion and culture. This shows moderate-to-strong agreement across different models in how the rating scale is utilized. 
In addition to this, we also quantify intra-model reliability primarily using Kendall's $W$, and also use an additional rank-based metric, Fleiss' $\kappa$. Results for the same are shown in Table \ref{tab:reliability_new_metrics}. We find again that on average, intra-model reliability is also moderate-to-strong, demonstrating a consistent use of the rating Likert scale by the models. There is some deviation observed for intra-model agreement across different emotion categories and cultures, with Fleiss' $\kappa$ showing the lowest agreement (albeit positive and moderately high) for PSE emotions, as opposed to high agreement values shown for NSD emotions. Kendall's $W$ remains relatively more stable across all the emotion categories and cultures. 

Note that Kendall's $W$ also shows higher values compared to Fleiss' $\kappa$, which is to be expected for data with small variance, as Fleiss' $\kappa$ is more robust to chance observations. It is also important to note that Kendall's $W$ is a better agreement measure for ordinal data than Fleiss' $\kappa$ as it quantifies rank orderings amongst raters. For Chile and Mexico, the results were comparable for the human study and our experiments. For USA, the scores are much smaller for positive emotions compared to the human study, while the negative emotions are more aligned with the human study. We further describe the precise reasons for the unsuitability of Cronbach's Alpha~\cite{cronbach1951coefficient} for the LLM responses in the following paragraphs.

\begin{table}[h]
\centering
\resizebox{\columnwidth}{!}{
\begin{tabular}{llccccc}
\toprule
\textbf{Model} & \textbf{Metric} & \textbf{Country} & \textbf{PSE} & \textbf{PSD} & \textbf{NSE} & \textbf{NSD} \\
\midrule

% ================= GPT =================
\multirow{9}{*}{\textbf{GPT}}
 & \multirow{3}{*}{Cronbach's $\alpha$}
    & USA   & -0.123 &  0.135 &  0.159 & -0.060 \\
 &  & Chile & -0.059 &  0.049 &  0.024 &  0.020 \\
 &  & Mexico& -0.020 & -0.004 & -0.002 &  0.041 \\
\cmidrule(lr){2-7}
 & \multirow{3}{*}{Fleiss' $\kappa$}
    & USA   &  0.592 &  0.475 &  0.554 &  0.771 \\
 &  & Chile &  0.661 &  0.574 &  0.612 &  0.739 \\
 &  & Mexico&  0.674 &  0.581 &  0.587 &  0.692 \\
\cmidrule(lr){2-7}
 & \multirow{3}{*}{Kendall's $W$}
    & USA   &  0.802 &  0.801 &  0.650 &  0.842 \\
 &  & Chile &  0.801 &  0.788 &  0.695 &  0.830 \\
 &  & Mexico&  0.758 &  0.784 &  0.712 &  0.825 \\
\midrule

% ================= MISTRAL =================
\multirow{9}{*}{\textbf{Mistral}}
 & \multirow{3}{*}{Cronbach's $\alpha$}
    & USA   &  0.020 &  0.162 & -0.076 &  0.007 \\
 &  & Chile & -0.060 &  0.084 & -0.017 & -0.008 \\
 &  & Mexico&  0.187 &  0.018 &  0.068 & -0.035 \\
\cmidrule(lr){2-7}
 & \multirow{3}{*}{Fleiss' $\kappa$}
    & USA   &  0.575 &  0.722 &  0.653 &  0.870 \\
 &  & Chile &  0.597 &  0.483 &  0.884 &  0.807 \\
 &  & Mexico&  0.621 &  0.504 &  0.862 &  0.800 \\
\cmidrule(lr){2-7}
 & \multirow{3}{*}{Kendall's $W$}
    & USA   &  0.840 &  0.805 &  0.821 &  0.803 \\
 &  & Chile &  0.843 &  0.798 &  0.800 &  0.802 \\
 &  & Mexico&  0.857 &  0.803 &  0.800 &  0.801 \\
\midrule

% ================= PHI =================
\multirow{9}{*}{\textbf{Phi}}
 & \multirow{3}{*}{Cronbach's $\alpha$}
    & USA   & -0.040 &  0.017 & -0.131 & -0.049 \\
 &  & Chile &  0.136 &  \textemdash & -0.025 &  0.090 \\
 &  & Mexico& -0.187 & -0.094 & -0.081 &  0.029 \\
\cmidrule(lr){2-7}
 & \multirow{3}{*}{Fleiss' $\kappa$}
    & USA   &  0.622 &  0.666 &  0.726 &  0.851 \\
 &  & Chile &  0.547 &  0.753 &  0.740 &  0.904 \\
 &  & Mexico&  0.665 &  0.664 &  0.701 &  0.927 \\
\cmidrule(lr){2-7}
 & \multirow{3}{*}{Kendall's $W$}
    & USA   &  0.860 &  0.847 &  0.702 &  0.833 \\
 &  & Chile &  0.848 &  0.799 &  0.689 &  0.827 \\
 &  & Mexico&  0.846 &  0.802 &  0.722 &  0.836 \\
\midrule

% ================= GEMINI =================
\multirow{9}{*}{\textbf{Gemini}}
 & \multirow{3}{*}{Cronbach's $\alpha$}
    & USA   & -0.106 &  0.165 & -0.012 & -0.078 \\
 &  & Chile & -0.091 &  0.060 & -0.045 & -0.104 \\
 &  & Mexico&  0.005 & -0.061 & -0.036 & -0.049 \\
\cmidrule(lr){2-7}
 & \multirow{3}{*}{Fleiss' $\kappa$}
    & USA   &  0.559 &  0.653 &  0.632 &  0.716 \\
 &  & Chile &  0.442 &  0.642 &  0.578 &  0.662 \\
 &  & Mexico&  0.363 &  0.614 &  0.618 &  0.627 \\
\cmidrule(lr){2-7}
 & \multirow{3}{*}{Kendall's $W$}
    & USA   &  0.456 &  0.711 &  0.638 &  0.845 \\
 &  & Chile &  0.421 &  0.780 &  0.761 &  0.811 \\
 &  & Mexico&  0.340 &  0.808 &  0.776 &  0.817 \\
 \midrule
 
 % ================= DEEPSEEK =================
\multirow{9}{*}{\textbf{DeepSeek}}
 & \multirow{3}{*}{Cronbach's $\alpha$}
    & USA   & -0.050 &  0.094 &  0.067 & -0.127 \\
 &  & Chile &  -0.14 &  0.09 &  -0.17 &  -0.07 \\
 &  & Mexico & -0.030 & -0.048 & -0.100 &  0.021 \\
\cmidrule(lr){2-7}
 & \multirow{3}{*}{Fleiss' $\kappa$}
    & USA   &  0.456 &  0.668 &  0.702 &  0.860 \\
 &  & Chile &  0.27 &  0.77 &  0.76 &  0.80 \\
 &  & Mexico&  0.293 &  0.696 &  0.780 &  0.729 \\
\cmidrule(lr){2-7}
 & \multirow{3}{*}{Kendall's $W$}
    & USA   &  0.620 &  0.857 &  0.614 &  0.833 \\
 &  & Chile &  0.575 &  0.816 & 0.627 & 0.843 \\
 &  & Mexico&  0.600 &  0.826 &  0.628 &  0.836 \\
\midrule

 % ================= QWEN =================
\multirow{9}{*}{\textbf{Qwen}}
 & \multirow{3}{*}{Cronbach's $\alpha$}
    & USA & 0.03 & \textemdash & 0.017 & \textemdash \\
 &  & Chile &  0.017 & -0.070 & 0.067 & 0.01 \\
 &  & Mexico & -0.041 & -0.004 & 0.061 & -0.123 \\
\cmidrule(lr){2-7}
 & \multirow{3}{*}{Fleiss' $\kappa$}
    & USA   &  0.908 & 0.929 & 0.951 & 0. 978 \\
 &  & Chile &  0.425 & 0.418 & 0.62 & 0.599 \\
 &  & Mexico&  0.374 & 0.426 & 0.607 & 0.591 \\
\cmidrule(lr){2-7}
 & \multirow{3}{*}{Kendall's $W$}
    & USA   &  0.857 & 0.846 & 0.884 & 0.850 \\
 &  & Chile &  0.681 & 0.631 & 0.71 & 0.733  \\
 &  & Mexico&  0.654 & 0.692 & 0.716 & 0.721 \\
\bottomrule
\end{tabular}}
\caption{Reliability and Agreement Metrics for all studied models}
\label{tab:reliability_new_metrics}
\end{table}

In the reliability analysis using Cronbach's alpha, we observed values for the emotion categories and individual emotions diverging significantly from what was reported in the human study. Upon conducting further variance analysis, we found that the LLMs were returning largely deterministic responses. Since we run the same prompt by the model 190 times, and treat each run as a participant response, a lack of diversity in the responses can result in the variances tending to 0. In the additional analysis, we observed low variance within each emotion reponses and low correlation between emotions within the same emotion type. In such cases, where there is little to no variability in responses, reliability metrics like Cronbach's alpha collapse. 

Since we cannot rely on variance-based metrics owing to the nature of our experiment setup, we searched for other metrics to measure the reliability of these prompts. We turned to agreement-based statistics because they do not rely on variance between-persons but instead measure how much consensus there is between participants. 

\subsection{ANOVA Analysis}
In Table \ref{tab:anova_gpt} we report the degrees of freedom (dF). F-scores, p-value (statistical significance), and $\eta^2$ (practical significance). The most noticeable observation from the full factorial analysis is the inflated F-scores showing a strong effect for the interaction, which are significantly higher compared to the ones reported in the human study. The p-values for all factor interactions are significant (p \lesser 0.05). However, the consistently small $\eta^2$ value for each interaction underscores the low practical significance of these results. Statistically significant p-values and very small $\eta^2$ can be caused by low within-group variability, since the overall variance explained by the factor is small relative to total variance. Typically, extremely low within-group variability is the result of repeated measures, where the responses from the same participant for the same category can be reported multiple times. Since we run each prompt through the LLM 190 times, and each of those iterations is treated as a single participant, our experiment defines these as independent observations. However, since the LLMs have produced deterministic outputs in our experiments, the resulting observations become repeated measures. Similar to our reliability experiments, since the variance-based analysis collapses for our data, we turn to rank and frequency-based metrics for our ordinal data distributions, i.e., we turn to Mann-Whitney U statistical test to study the effect of these factors and their interactions.

\begin{table*}[p]
\centering
\caption{Full factorial ANOVA results for GPT}
\begin{adjustbox}{max width=\textwidth, max height=0.9\textheight}
\begin{tabular}{l
                S[table-format=7.3]
                S[table-format=4.0]
                S[table-format=8.3]
                S[table-format=1.3]
                S[table-format=1.3]}
\toprule
\textbf{Effect} & \textbf{Sum Sq} & \textbf{df} & \textbf{F} & \textbf{p-value} & \textbf{$\eta^2$} \\
\midrule
C(participant) & 624.823 & 569 & 22.775 & 0.000 & 0.027 \\
emotion valence & 842.846 & 1 & 17480.807 & 0.000 & 0.037 \\
emotion type & 277.378 & 1 & 5752.885 & 0.000 & 0.012 \\
situation valence & 535.295 & 1 & 11102.132 & 0.000 & 0.023 \\
situation type & 620.836 & 1 & 12876.761 & 0.000 & 0.027 \\
culture & 452.217 & 2 & 4689.541 & 0.000 & 0.020 \\
emotion valence × emotion type & 149.202 & 1 & 3094.481 & 0.000 & 0.006 \\
emotion valence × situation valence & 8087.975 & 1 & 167746.321 & 0.000 & 0.352 \\
emotion valence × situation type & 256.510 & 1 & 5320.073 & 0.000 & 0.011 \\
emotion valence × culture & 439.826 & 2 & 4561.043 & 0.000 & 0.019 \\
emotion type × situation valence & 309.290 & 1 & 6414.730 & 0.000 & 0.013 \\
emotion type × situation type & 687.227 & 1 & 14253.239 & 0.000 & 0.030 \\
emotion type × culture & 458.071 & 2 & 4750.246 & 0.000 & 0.020 \\
situation valence × situation type & 49.000 & 1 & 186.667 & 0.000 & 0.002 \\
situation valence × culture & 651.407 & 2 & 6755.157 & 0.000 & 0.028 \\
situation type × culture & 319.617 & 2 & 3314.464 & 0.000 & 0.014 \\
emotion valence × emotion type × situation valence & 1067.026 & 1 & 23140.353 & 0.000 & 0.046 \\
emotion valence × emotion type × situation type & 309.290 & 1 & 6414.730 & 0.000 & 0.013 \\
emotion valence × emotion type × culture & 178.683 & 2 & 1852.965 & 0.000 & 0.008 \\
emotion valence × situation valence × situation type & 179.790 & 1 & 3728.872 & 0.000 & 0.008 \\
emotion valence × situation valence × culture & 4046.220 & 2 & 41959.729 & 0.000 & 0.176 \\
emotion valence × situation type × culture & 207.814 & 2 & 2155.055 & 0.000 & 0.009 \\
emotion type × situation valence × situation type & 0.388 & 1 & 8.051 & 0.005 & 0.000 \\
emotion type × situation valence × culture & 217.360 & 2 & 2254.042 & 0.000 & 0.009 \\
emotion type × situation type × culture & 347.241 & 2 & 3609.922 & 0.000 & 0.015 \\
situation valence × situation type × culture & 17.075 & 2 & 177.064 & 0.000 & 0.001 \\
emotion valence × emotion type × situation valence × situation type & 101.158 & 1 & 2098.037 & 0.000 & 0.004 \\
emotion valence × emotion type × situation valence × culture & 603.539 & 2 & 6258.762 & 0.000 & 0.026 \\
emotion valence × emotion type × situation type × culture & 217.360 & 2 & 2254.042 & 0.000 & 0.009 \\
emotion valence × situation valence × situation type × culture & 157.158 & 2 & 1629.745 & 0.000 & 0.007 \\
emotion type × situation valence × situation type × culture & 10.847 & 2 & 112.486 & 0.000 & 0.000 \\
emotion valence × emotion type × situation valence × situation type × culture & 163.203 & 2 & 1692.436 & 0.000 & 0.007 \\
Residual & 410.073 & 8505 & \text{---} & \text{---} & 0.018 \\
\bottomrule
\end{tabular}%
\end{adjustbox}
\label{tab:anova_gpt}
\end{table*}

\subsection{Additional Evidence for Results}
\subsubsection{Engaging Vs. Disengaging Emotions}

Here, we present additional evidence of misalignment found in expression of engaging emotions, as opposed to disengaging emotions, pertaining to the hypotheses \htag{H1a} and \htag{H1b} from the human results. 

\begin{figure*}[tp]
    \begin{subfigure}{0.44\linewidth}
        \centering
        \includegraphics[width=\linewidth]{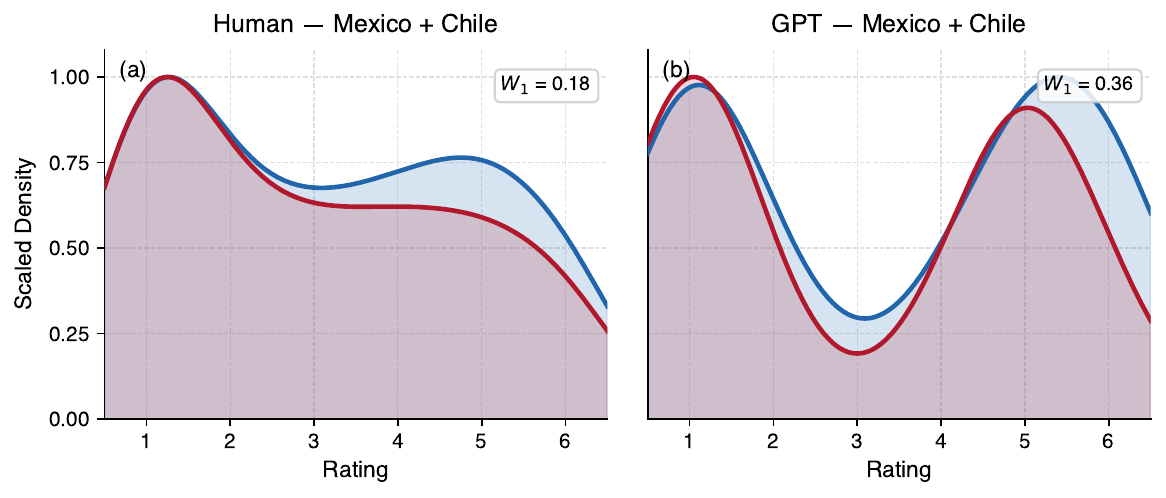}
        \caption{}
        \label{fig:gpt_eng_diseng_la}    
    \end{subfigure}
    \begin{subfigure}{0.44\linewidth}
        \centering
        \includegraphics[width=\linewidth]{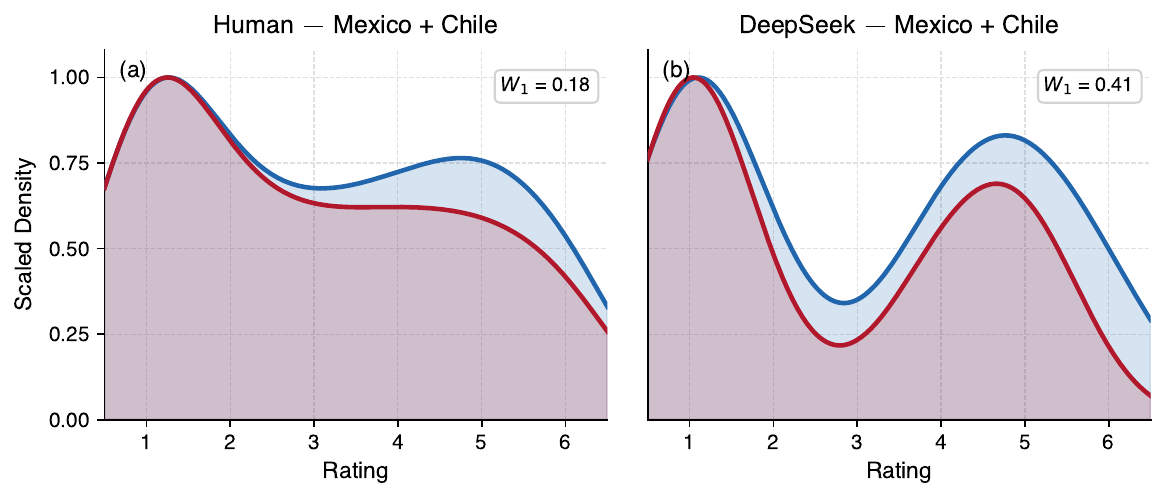}
        \caption{}
        \label{fig:ds_eng_diseng_la}    
    \end{subfigure}    
    \begin{subfigure}{0.44\linewidth}
        \centering
        \includegraphics[width=\linewidth]{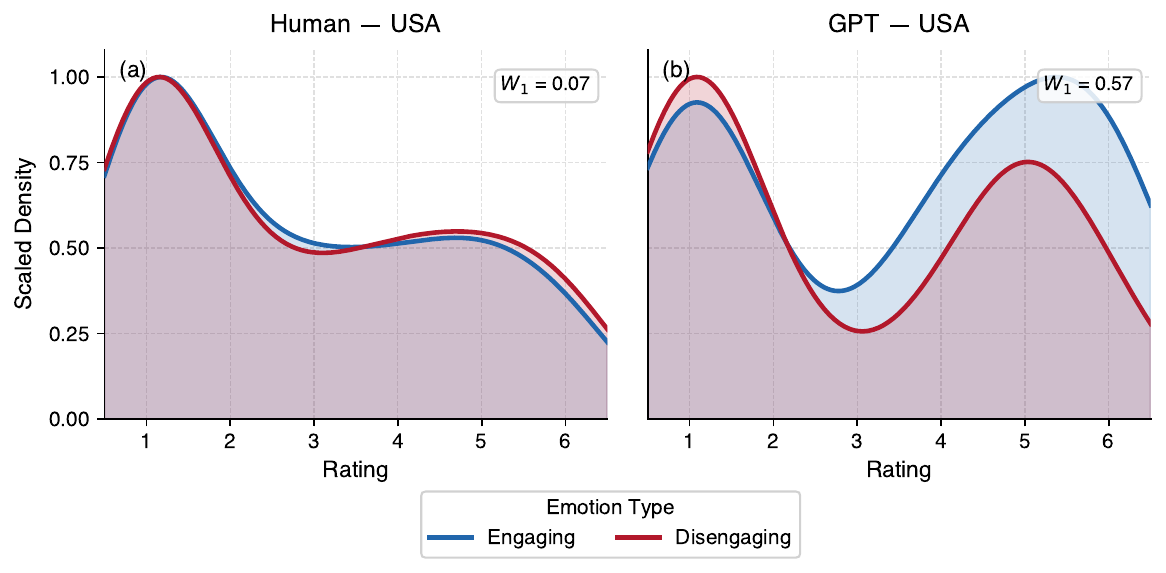}
        \caption{}
        \label{fig:gpt_eng_diseng_ea}    
    \end{subfigure}
    \begin{subfigure}{0.44\linewidth}
        \centering
        \includegraphics[width=\linewidth]{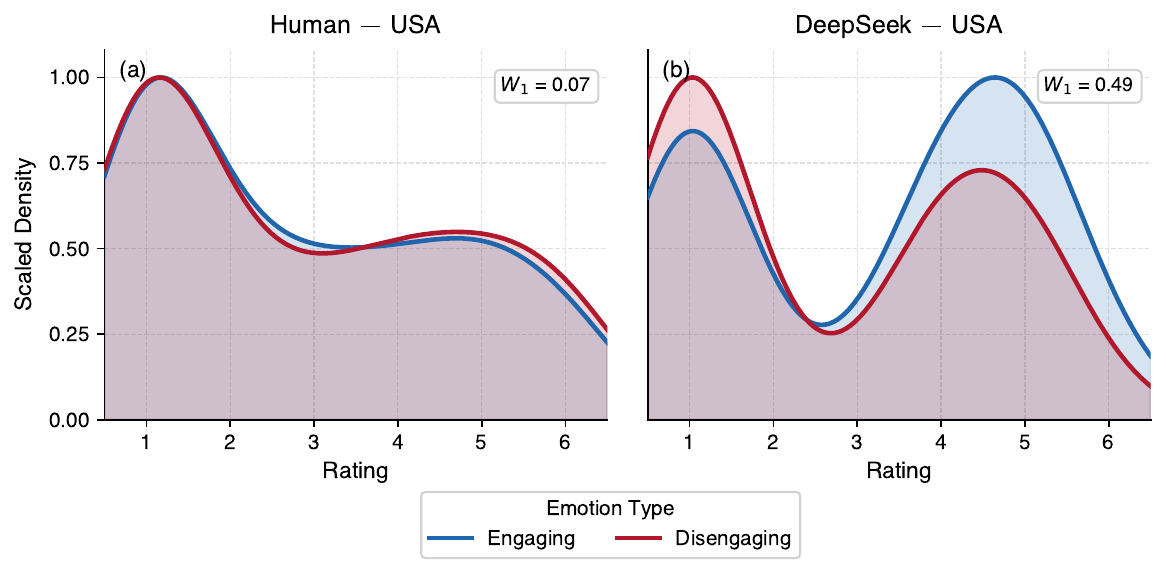}
        \caption{}
        \label{fig:ds_eng_diseng_ea}    
    \end{subfigure} 
    \begin{subfigure}{0.44\linewidth}
        \centering
        \includegraphics[width=\linewidth]{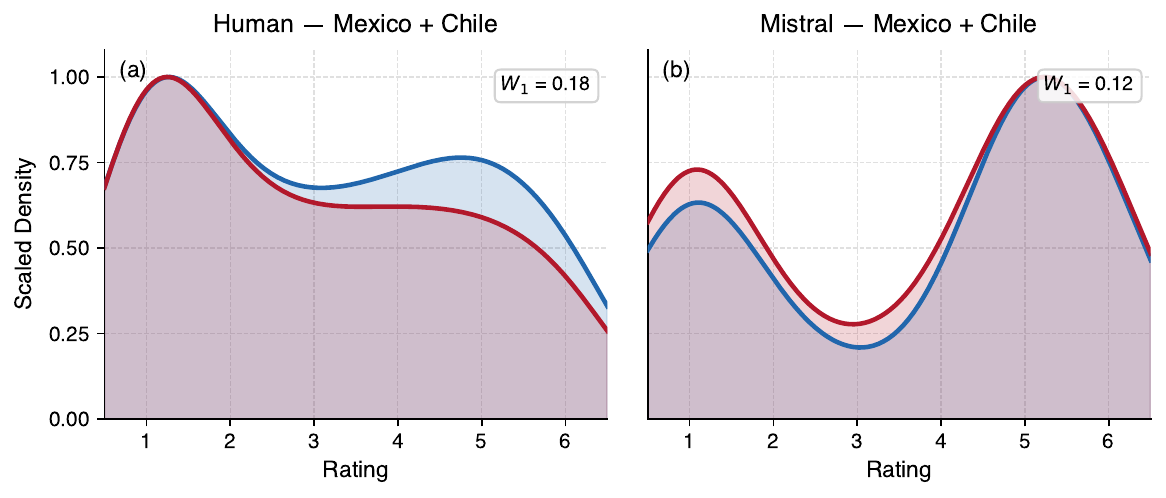}
        \caption{}
        \label{fig:mistral_eng_diseng_la}    
    \end{subfigure}
    \begin{subfigure}{0.44\linewidth}
        \centering
        \includegraphics[width=\linewidth]{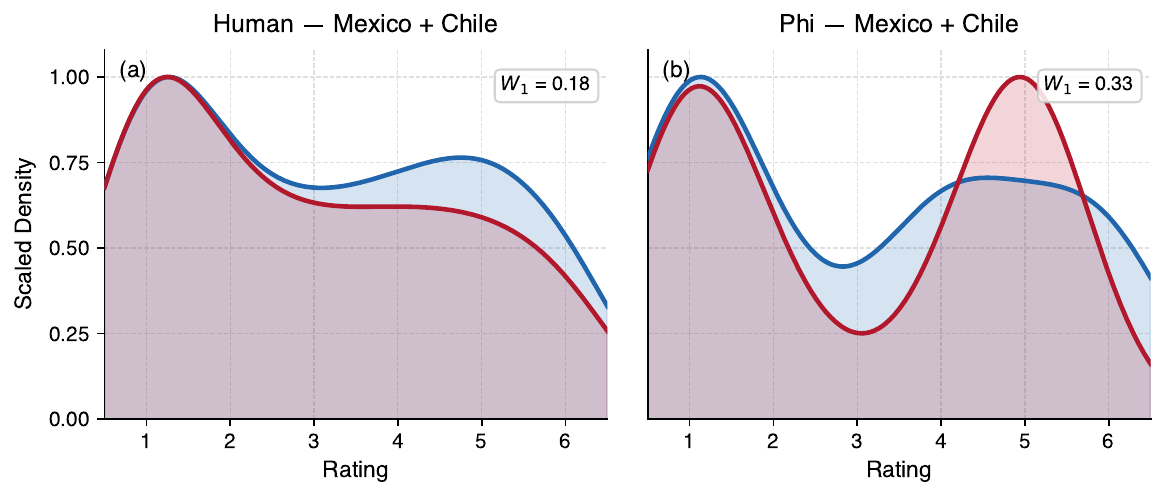}
        \caption{}
        \label{fig:phi_eng_diseng_la}    
    \end{subfigure}
    \begin{subfigure}{0.44\linewidth}
        \centering
        \includegraphics[width=\linewidth]{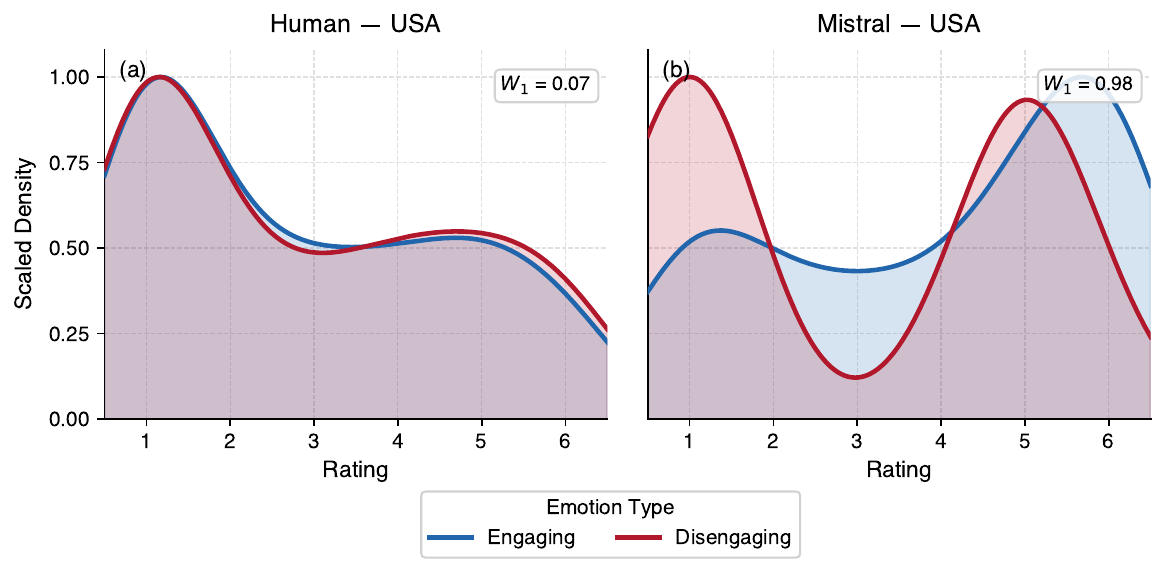}
        \caption{}
        \label{fig:mistral_eng_diseng_ea}    
    \end{subfigure}
    \begin{subfigure}{0.44\linewidth}
        \centering
        \includegraphics[width=\linewidth]{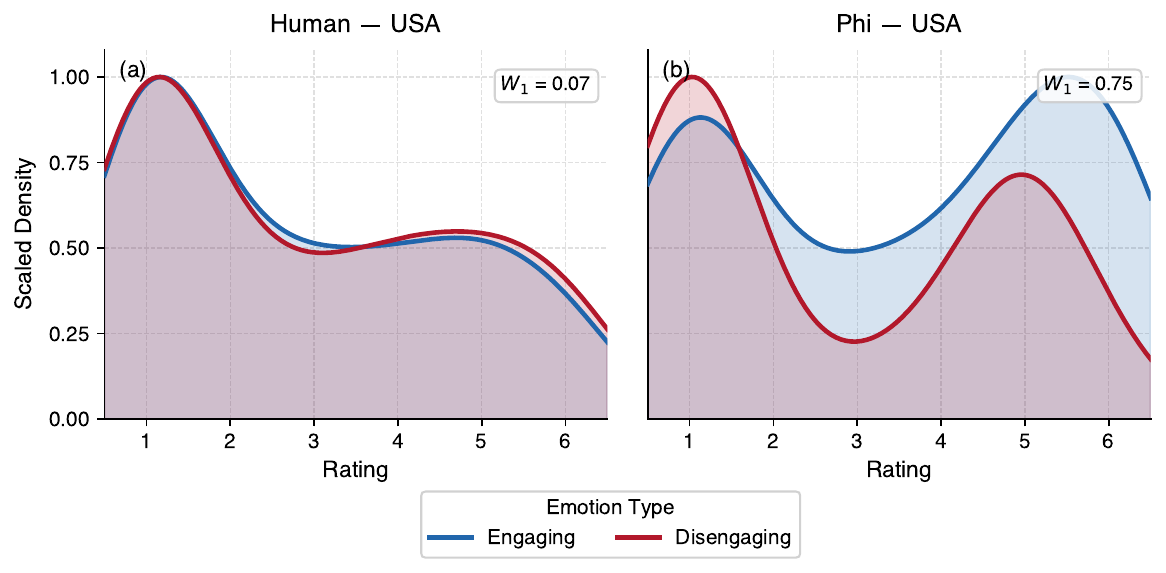}
        \caption{}
        \label{fig:phi_eng_diseng_ea}    
    \end{subfigure} 
    \begin{subfigure}{0.44\linewidth}
        \centering
        \includegraphics[width=\linewidth]{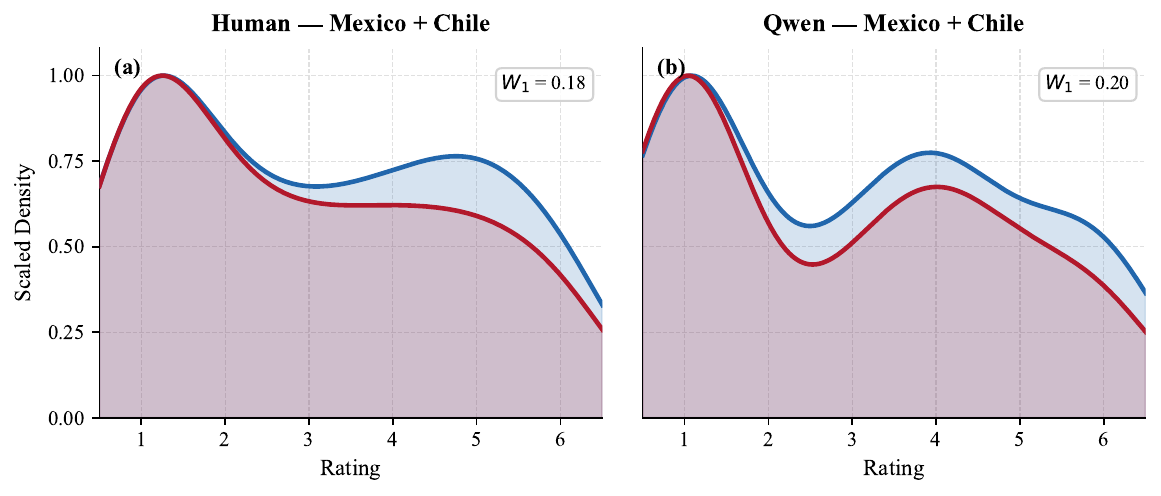}
        \caption{}
        \label{fig:qwen_eng_diseng_la}    
    \end{subfigure}\hfill
    \begin{subfigure}{0.44\linewidth}
        \centering
        \includegraphics[width=\linewidth]{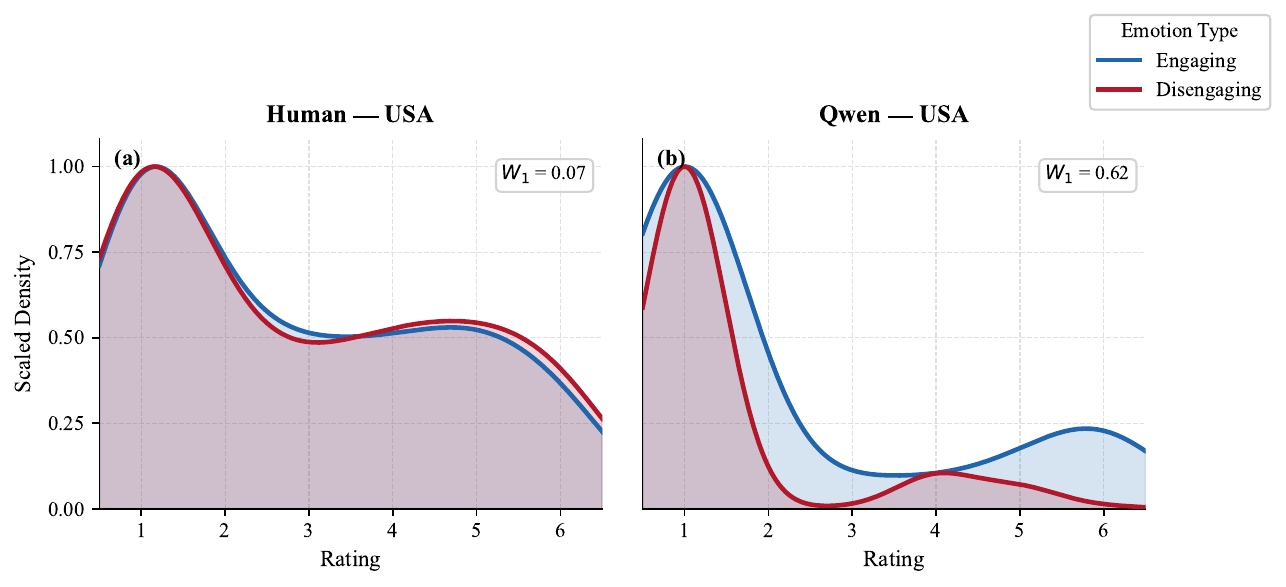}
        \caption{}
        \label{fig:qwen_eng_diseng_ea}    
    \end{subfigure}
    \caption{Distribution of ratings from humans for engaging and disengaging emotions (valence collapsed), compared with those from all models. For all models, the pair above shows the distribution for both Latin American personas combined, and the pair below shows the comparison for the European American persona. Only for Qwen, they are shown side-by-side. The values $W_1$ denote the 1-Wasserstein distance for each of the pairs of distributions. Note that corresponding alignment scores, based on directional statistical tests, are shown in rows 1 and 2 of Table \ref{tab:alignment_summary}.} 
    \label{fig:eng_diseng_all_main}
\end{figure*}

\begin{figure*}
\centering
    \begin{subfigure}{0.32\linewidth}
        \centering
        \includegraphics[width=\linewidth]{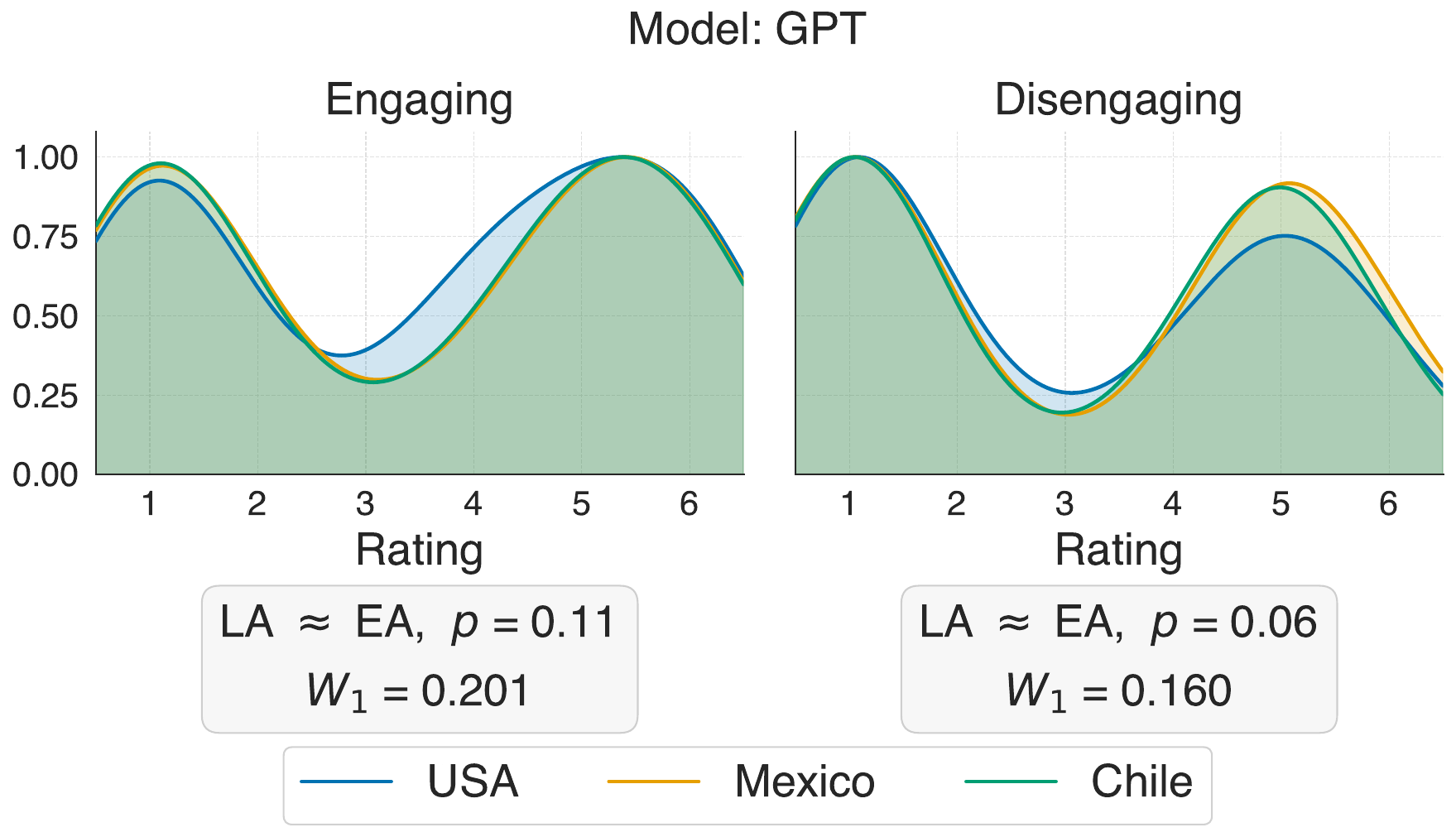}
        \caption{}
        \label{fig:GPT_absolute_eng_diseng}
    \end{subfigure}
    \begin{subfigure}{0.32\linewidth}
        \centering
        \includegraphics[width=\linewidth]{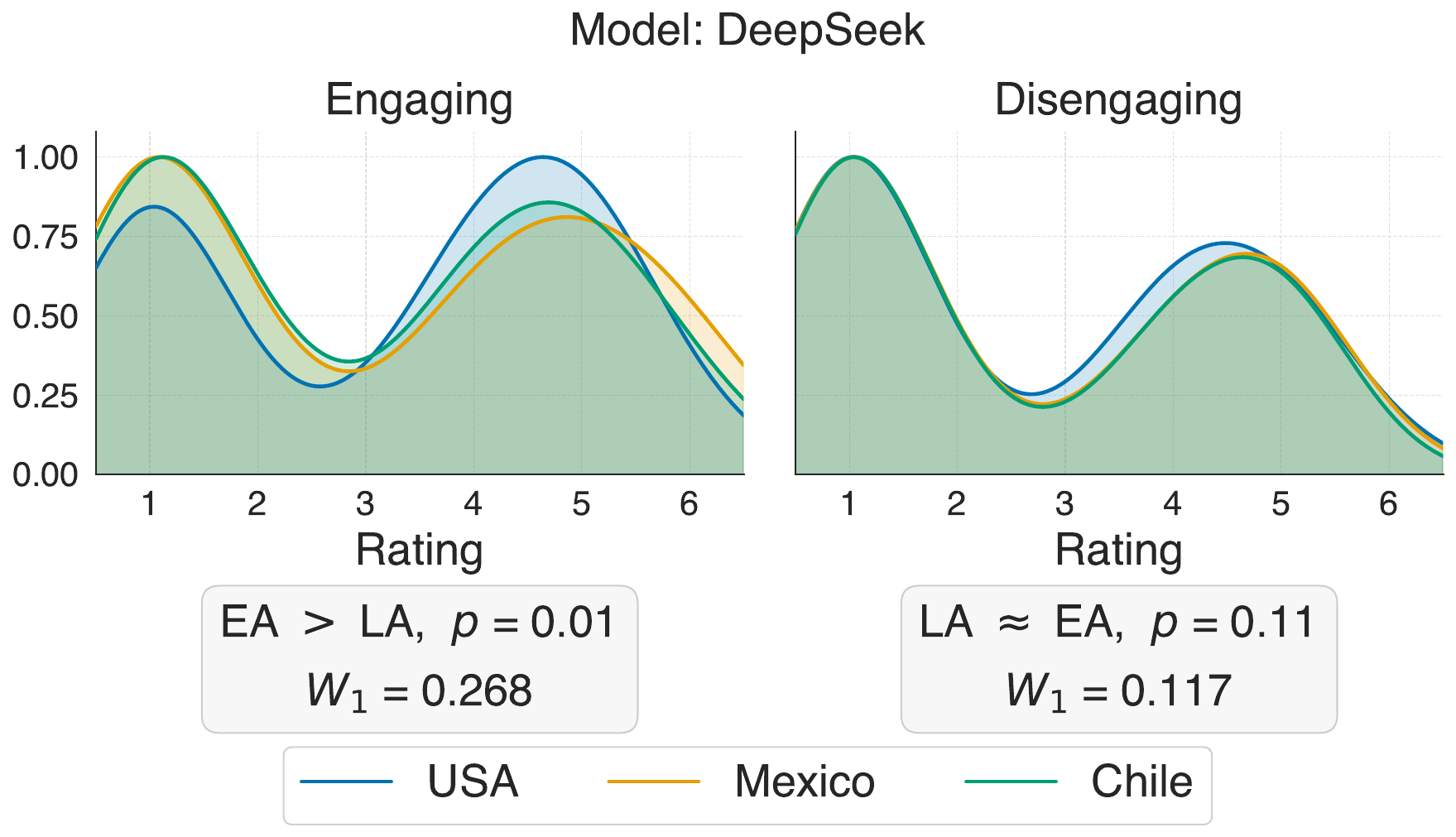}
        \caption{}
        \label{fig:deepseek_absolute_eng_diseng}
    \end{subfigure}
    \begin{subfigure}{0.32\linewidth}
        \centering
        \includegraphics[width=\linewidth]{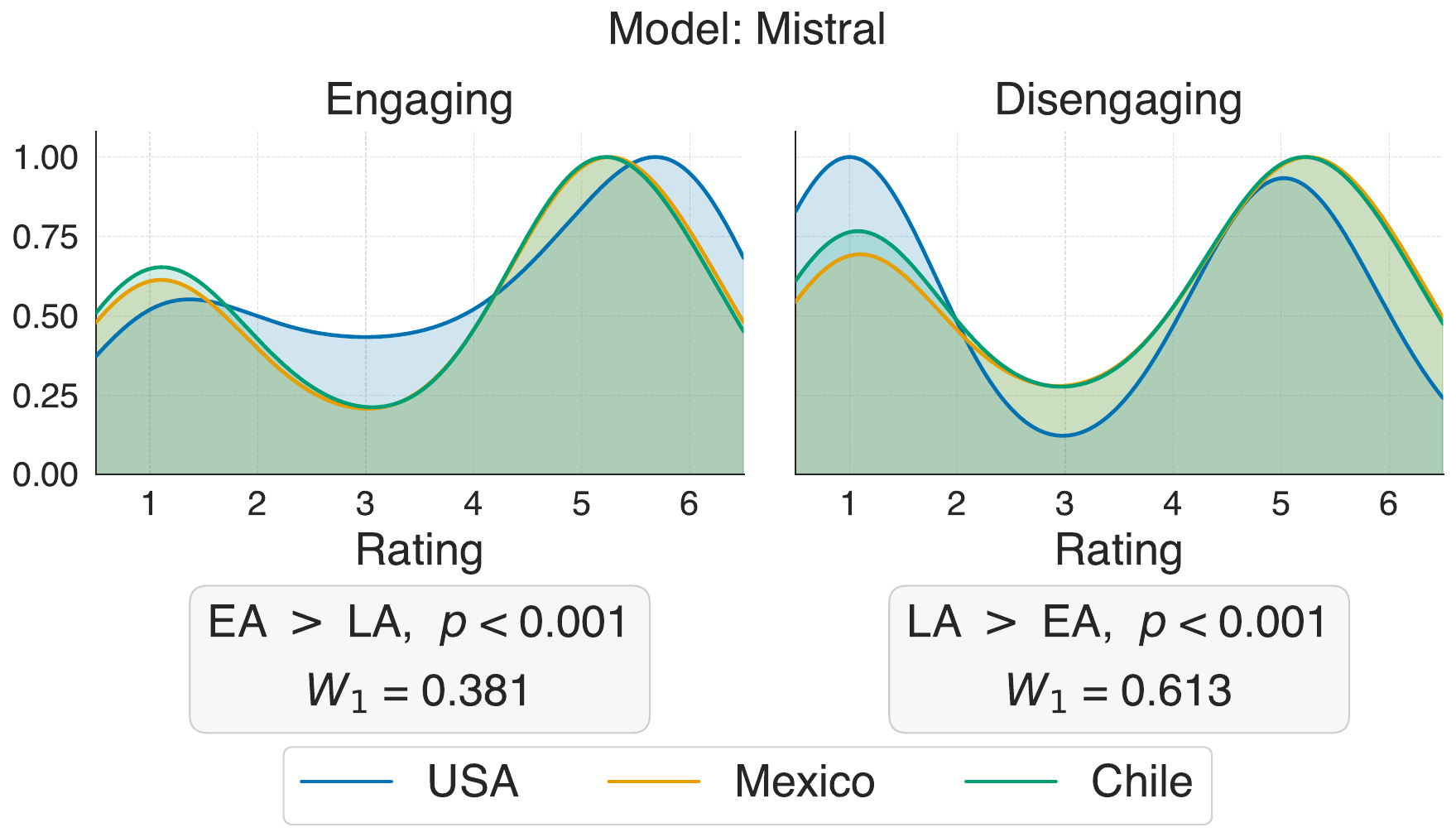}
        \caption{}
        \label{fig:misctral_absolute_eng_diseng}
    \end{subfigure}
    \caption{Cross-cultural comparison for the difference in expression of engaging (left panel for each sub-plot) and disengaging (right panel) emotions. Along with this, the LA (Mexico + Chile) and EA (USA) distributions are compared for directional significance, with the results shown below each sub-plot.}
    \label{fig:absolute_culture_eng_diseng2}
\end{figure*}

\begin{figure*}[ht!]
    \begin{subfigure}{0.32\linewidth}
        \centering
        \includegraphics[width=\linewidth]{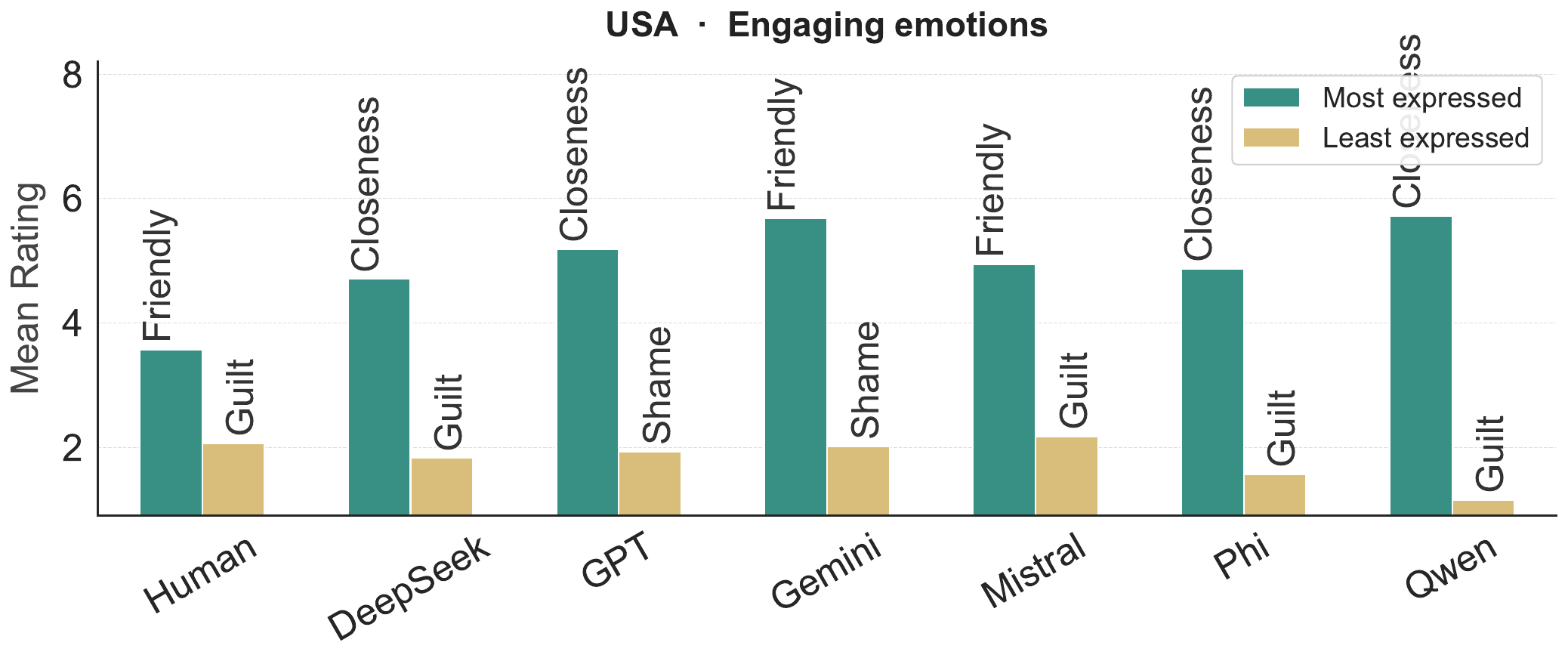}
        \caption{}
        \label{fig:usa_eng_extremes}    
    \end{subfigure}
    \begin{subfigure}{0.32\linewidth}
        \centering
        \includegraphics[width=\linewidth]{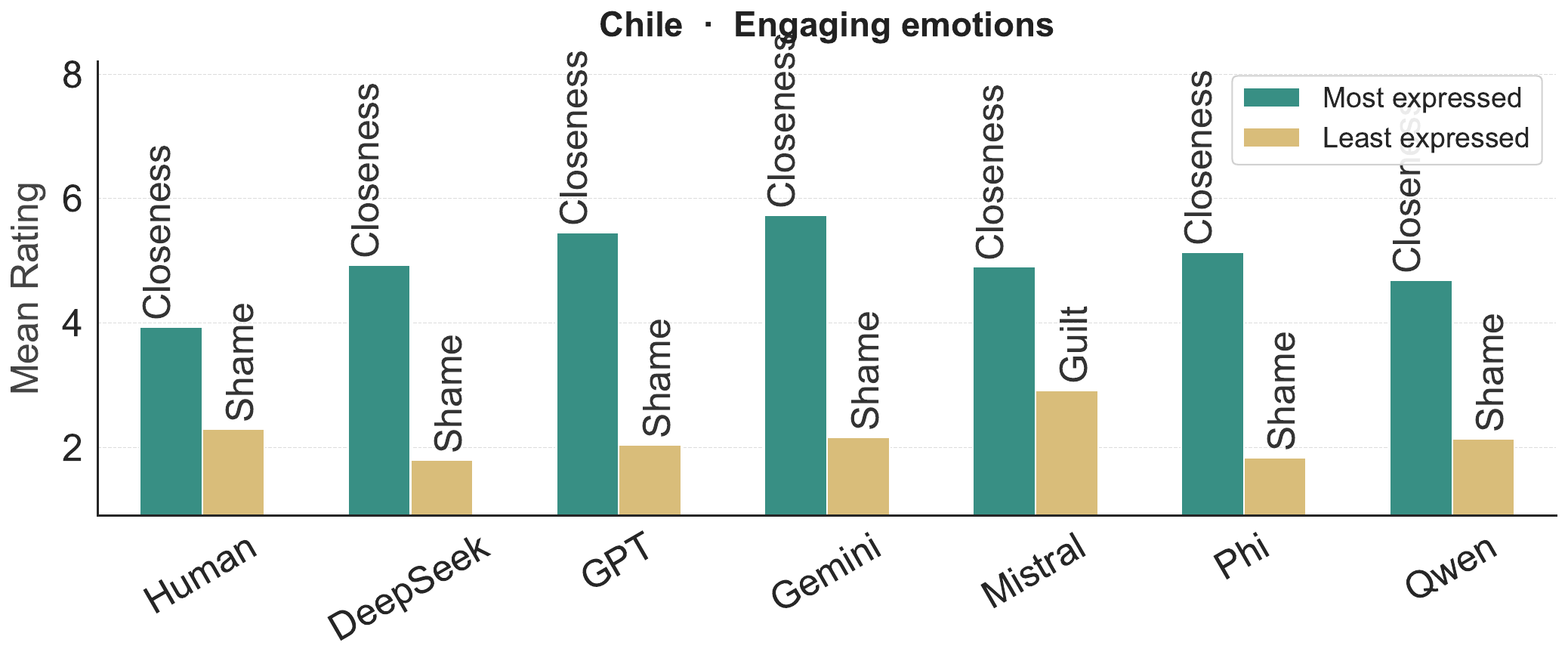}
        \caption{}
        \label{fig:chile_eng_extremes}    
    \end{subfigure}
    \begin{subfigure}{0.32\linewidth}
        \centering
        \includegraphics[width=\linewidth]{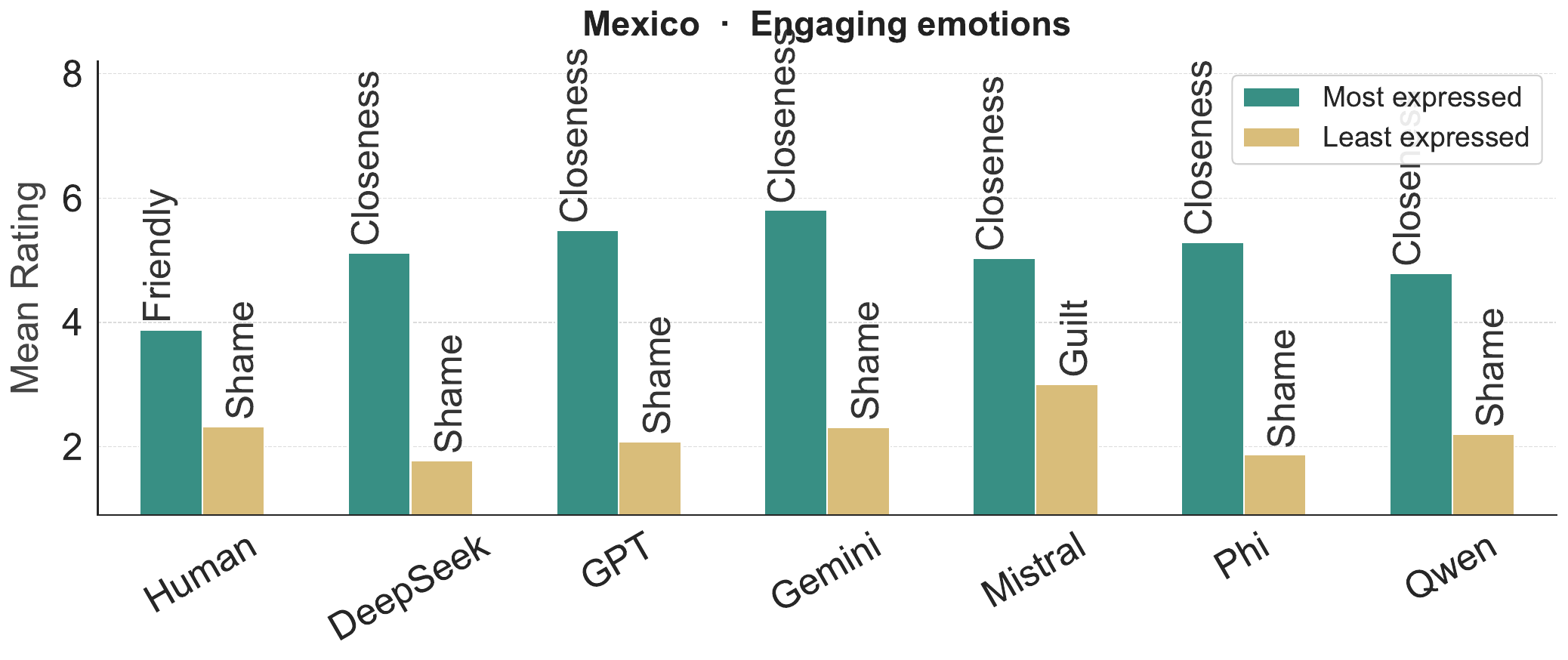}
        \caption{}
        \label{fig:mexico_eng_extremes}    
    \end{subfigure}
    \begin{subfigure}{0.33\linewidth}
        \centering
        \includegraphics[width=\linewidth]{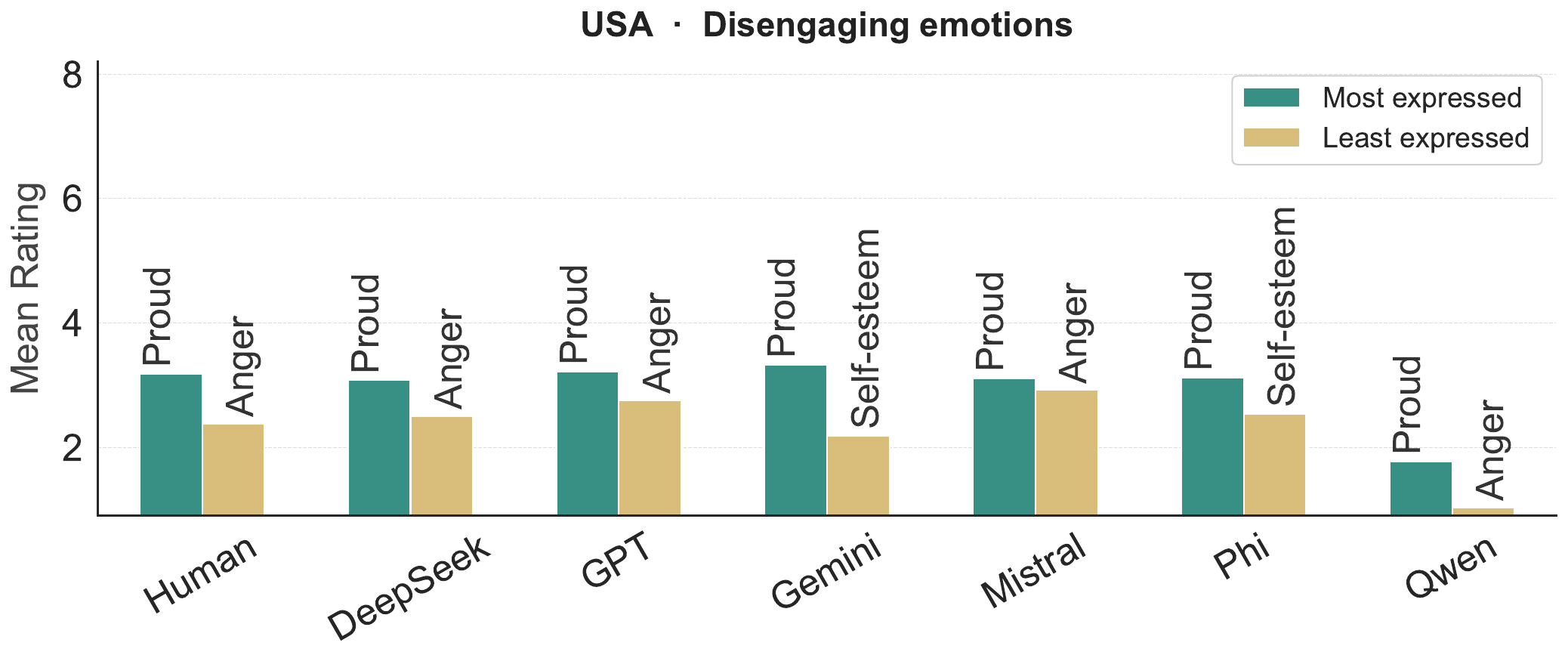}
        \caption{}
        \label{fig:usa_diseng_extremes}    
    \end{subfigure}
    \begin{subfigure}{0.33\linewidth}
        \centering
        \includegraphics[width=\linewidth]{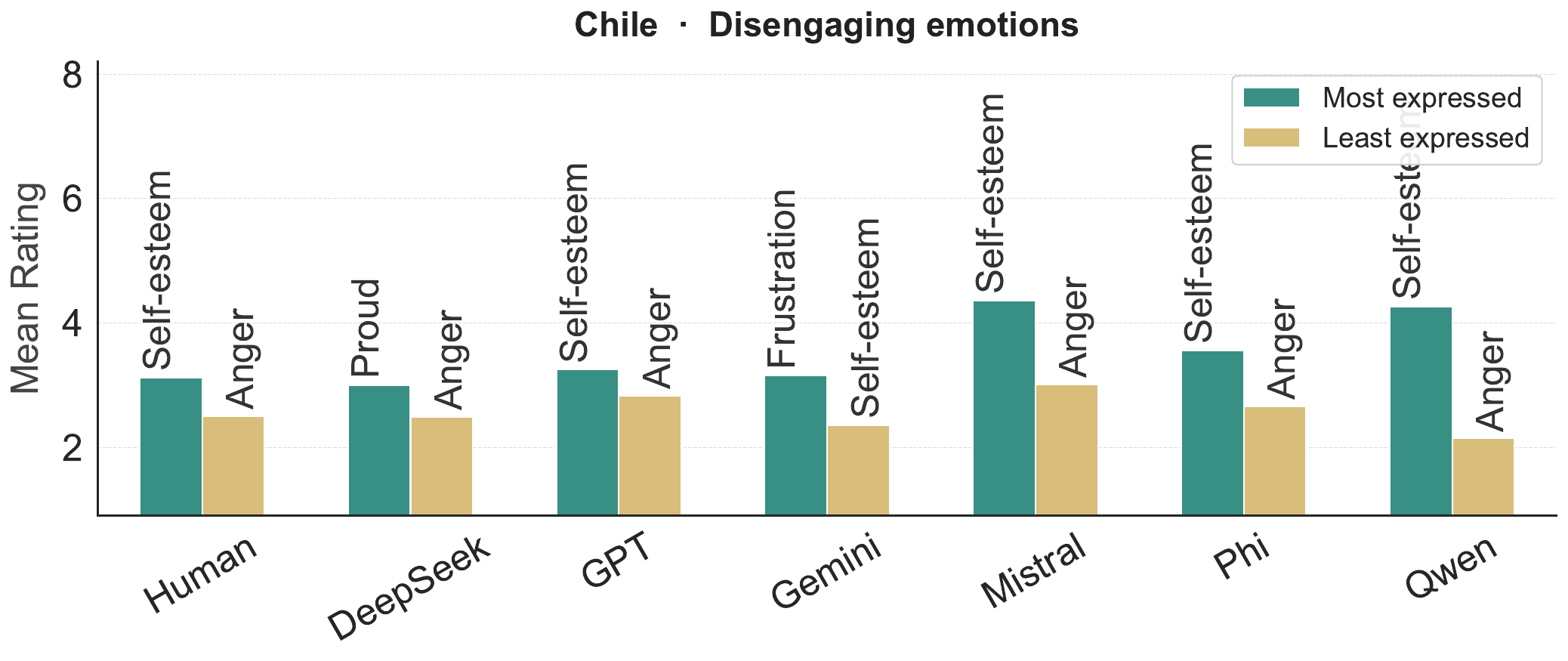}
        \caption{}
        \label{fig:chile_diseng_extremes}    
    \end{subfigure}
    \begin{subfigure}{0.33\linewidth}
        \centering
        \includegraphics[width=\linewidth]{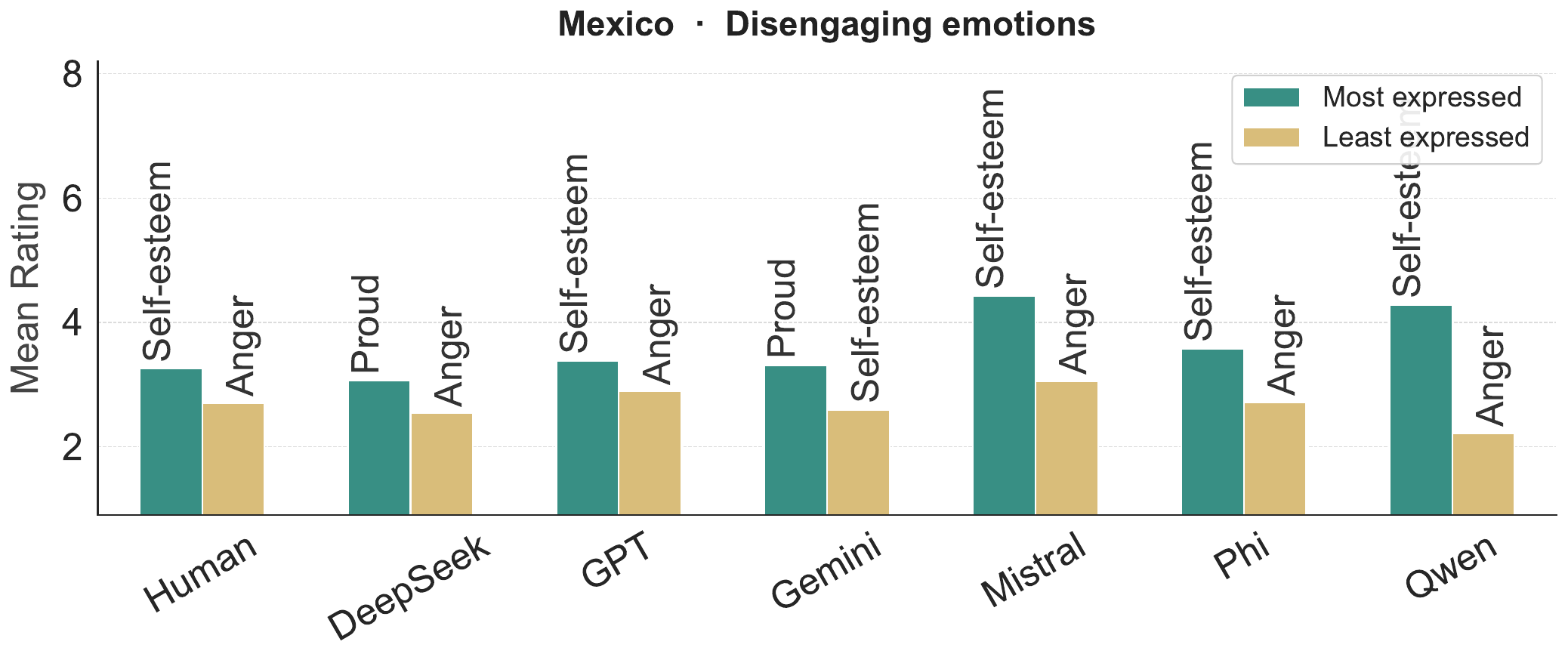}
        \caption{}
        \label{fig:mexico_diseng_extremes}    
    \end{subfigure}
    \caption{Emotion extremes shown for all models and cultures, across engaging and disengaging emotions (valence collapsed). Some of the emotions are displayed in shorthand, such as, ``Closeness" refers to Feelings of closeness with others, whereas ``Friendly" refers to friendly feelings. The most expressed emotion denotes the category that receives the highest average rating, across all scenarios, and vice versa for the least expressed emotion.}
    \label{fig:eng_diseng_extremes_all}
\end{figure*}

First, we present the full plots of distributional comparison for all models in Fig. \ref{fig:eng_diseng_all_main}. The alignment of each model here follows the pattern summarized in Table \ref{tab:alignment_summary}. 

Next, we also present additional cross-cultural comparisons for the models in Fig. \ref{fig:absolute_culture_eng_diseng2}. Similar to the results described in the main body, the overall trend holds for most other models. In particular, for both DeepSeek and Mistral, Engaging emotions are expressed more with the EA persona than the LA persona. For Disengaging emotions, both GPT and DeepSeek do not significantly differentiate when assigned different personas, whereas Mistral expresses it more with the LA persona. 

Beyond distributional comparisons, across all models and humans, we study the extremes of emotions specific to engaging or disengaging emotions (Fig.~\ref{fig:eng_diseng_extremes_all}). For all models, similar to humans, the gap between levels of expression is larger for engaging emotions than for disengaging emotions. This difference is also significantly larger for LLMs than for humans. Specifically, emotions like \textit{closeness to others} and \textit{friendly feelings} are strongly expressed by all models across all cultures, leading to the observed misalignment for the EA persona. Among engaging emotions, across personas, models express \textit{closeness to others} most strongly and frequently, while expressing negative engaging emotions the least strongly. Mistral expresses \textit{guilt} the least across all personas, while most other models express \textit{guilt} the least for EA and \textit{shame} the least for LA personas.

\subsubsection{Situational Comparison}

\begin{figure*}
\begin{subfigure}{0.5\linewidth}
    \centering
    \includegraphics[width=\linewidth]{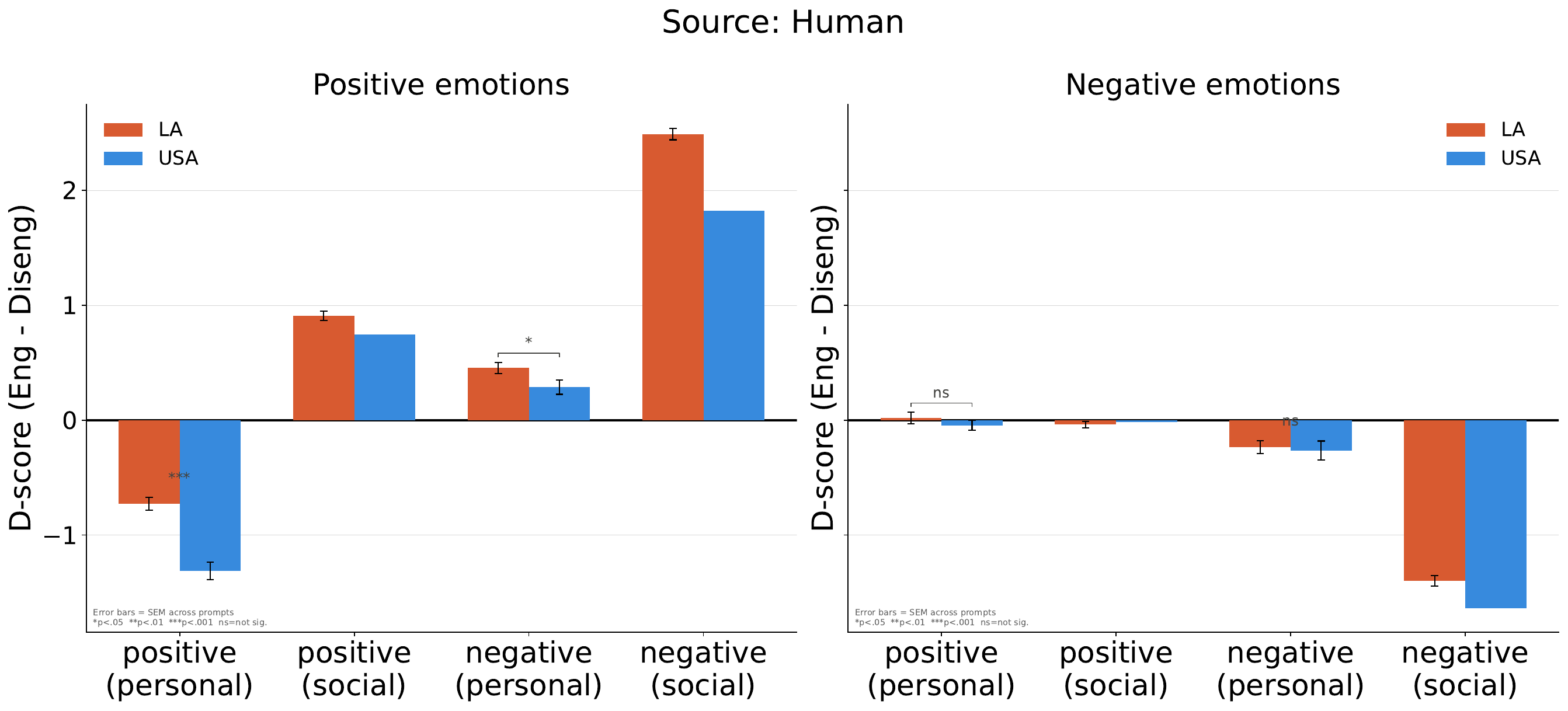}
    \caption{}
    \label{fig:human_situational_analysis}
\end{subfigure}
\begin{subfigure}{0.5\linewidth}
    \centering
    \includegraphics[width=\linewidth]{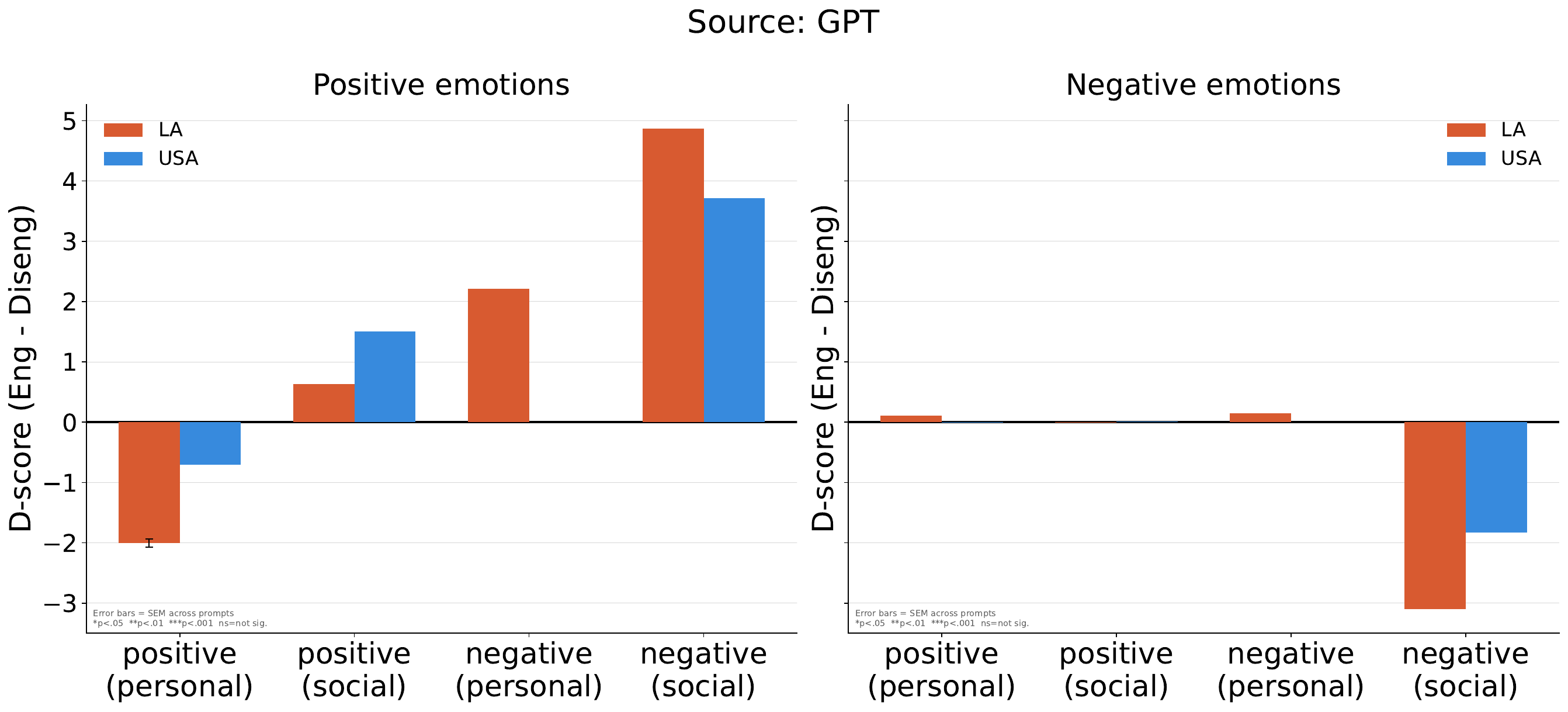}
    \caption{}
    \label{fig:gpt_situational_analysis}
\end{subfigure}
\begin{subfigure}{0.5\linewidth}
    \centering
    \includegraphics[width=\linewidth]{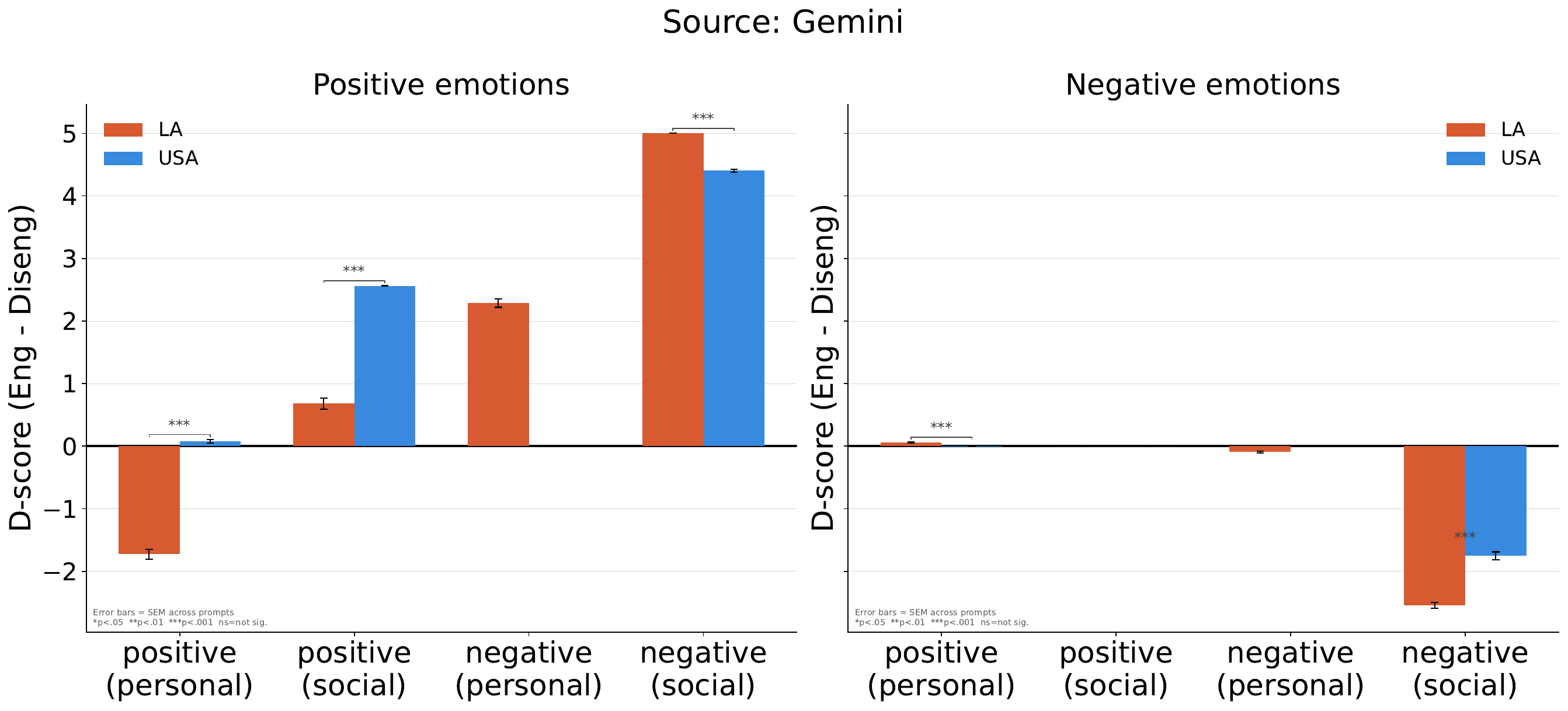}
    \caption{}
    \label{fig:gemini_situational_analysis}
\end{subfigure}
\begin{subfigure}{0.5\linewidth}
    \centering
    \includegraphics[width=\linewidth]{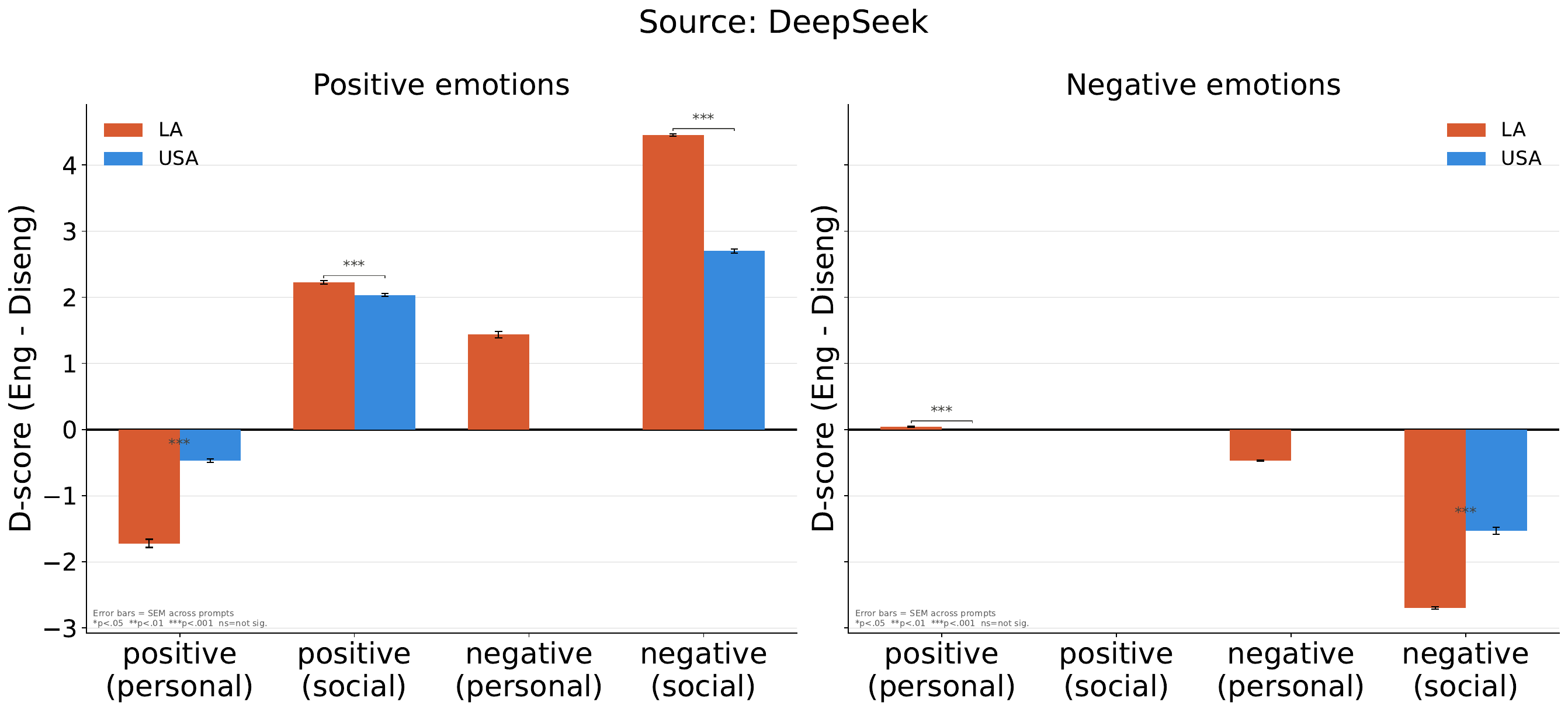}
    \caption{}
    \label{fig:deepseek_situational_analysis}
\end{subfigure}
\begin{subfigure}{0.5\linewidth}
    \centering
    \includegraphics[width=\linewidth]{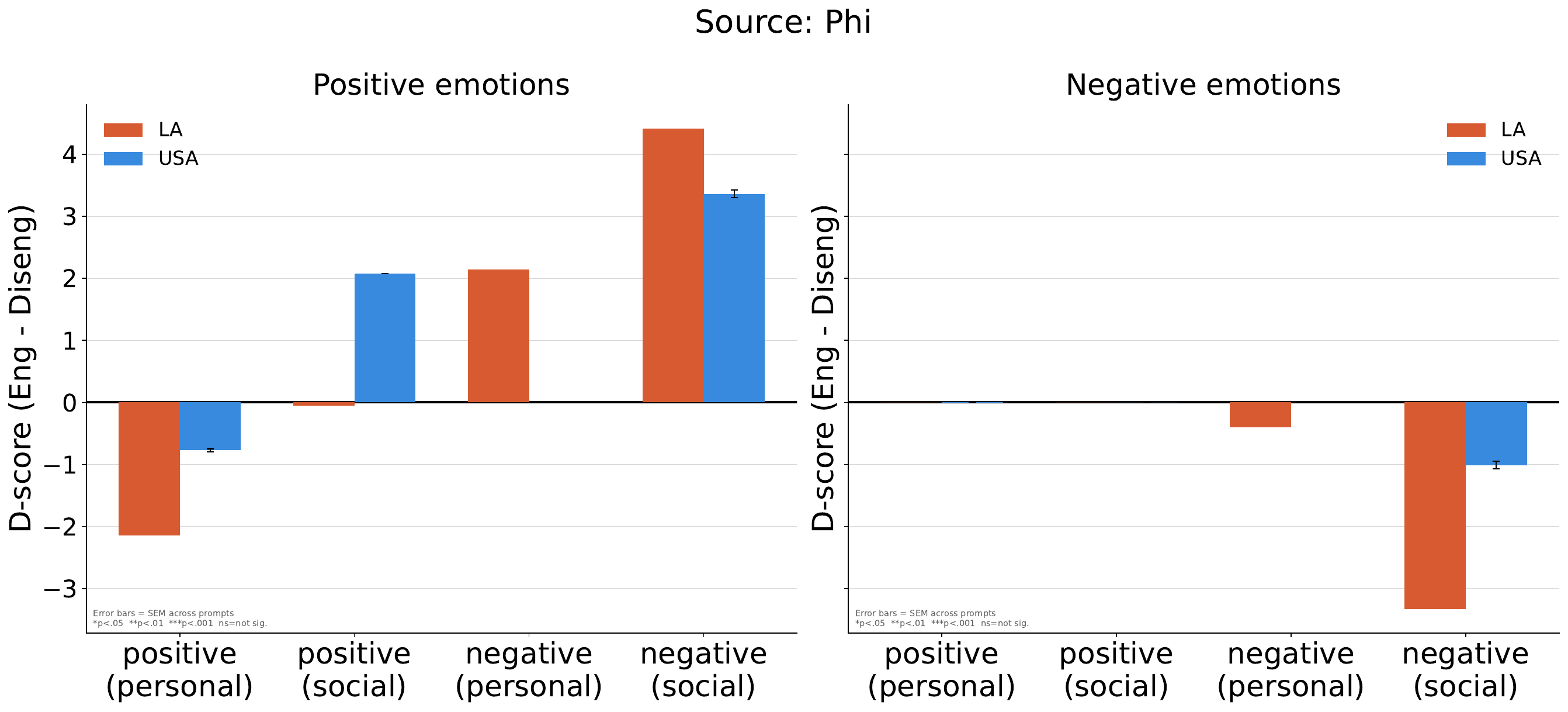}
    \caption{}
    \label{fig:phi_situational_analysis}
\end{subfigure}
\begin{subfigure}{0.5\linewidth}
    \centering
    \includegraphics[width=\linewidth]{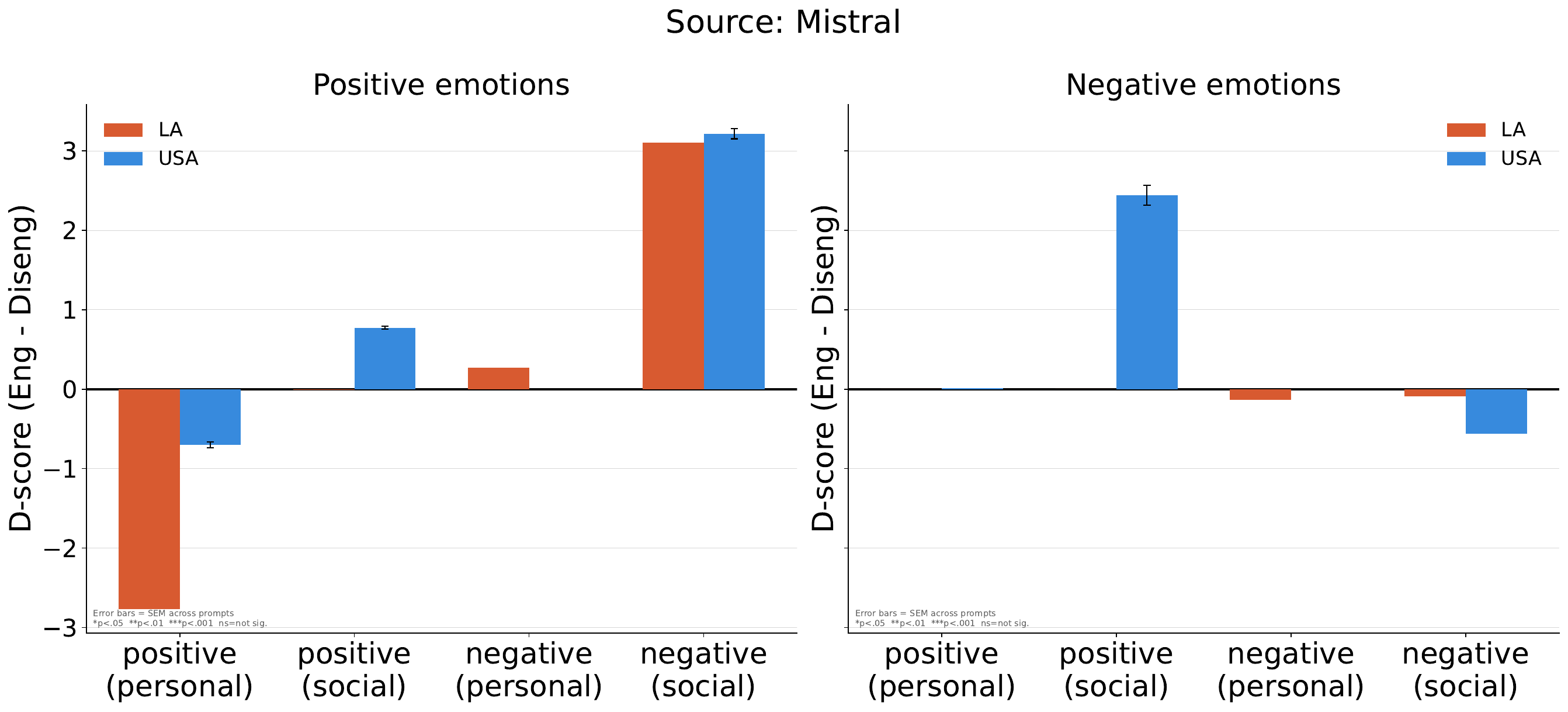}
    \caption{}
    \label{fig:mistral_situational_analysis}
\end{subfigure}
\begin{subfigure}{0.5\linewidth}
    \centering
    \includegraphics[width=\linewidth]{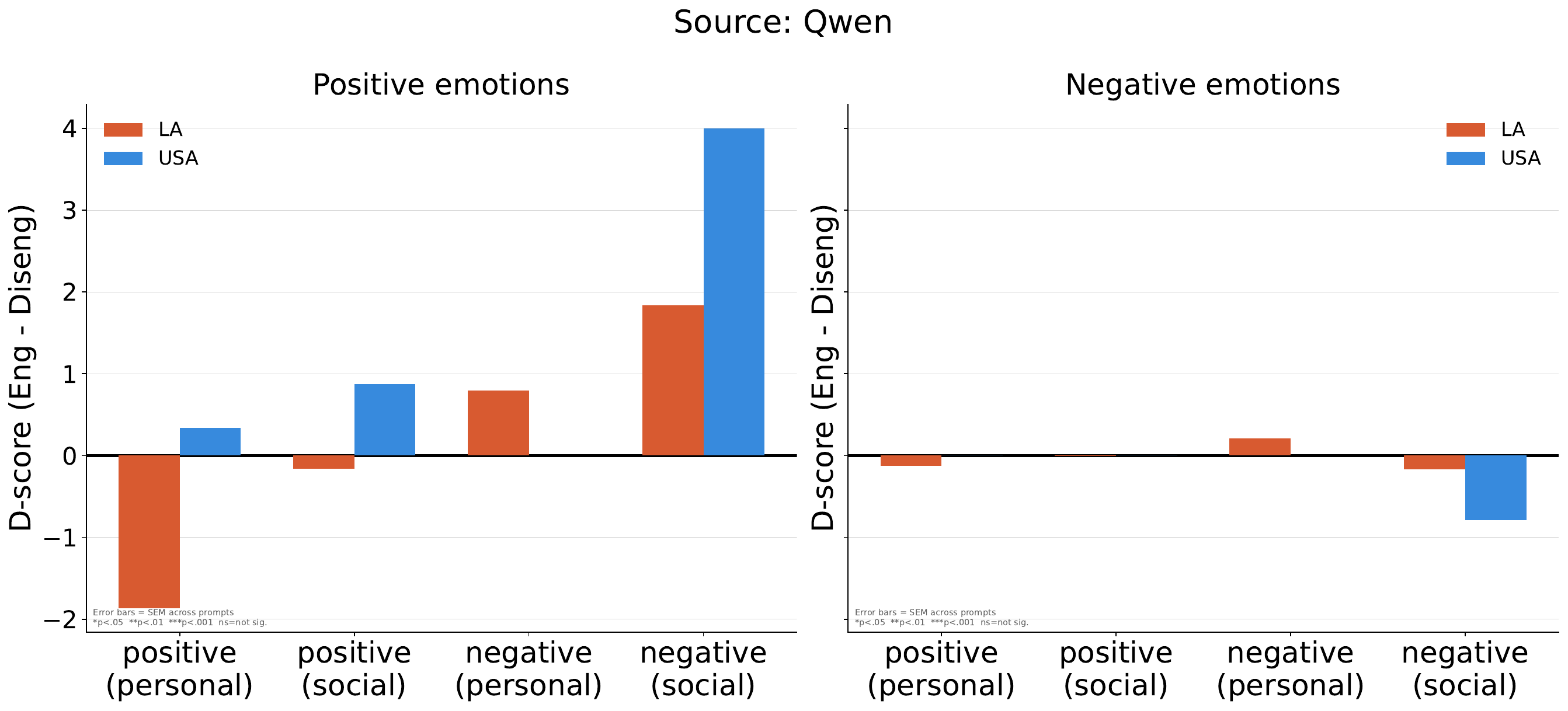}
    \caption{}
    \label{fig:qwen_situational_analysis}
\end{subfigure}
\caption{Situational comparisons for each model, along with Humans. The plots show the difference between the means of engaging and disengaging emotions expressed across the four different types of scenarios studied.}
\end{figure*}

The original human study \cite{salvador2024emotionally}, within the broader claims of expressivity, also studied granular differences in expressivity, within each type of situation (valence = positive, negative $\times$ sociality = personal, social). Similar to the original study, we analyze \textit{interdependence dominance} at the situation level---by calculating the difference of means of engaging and disengaging emotions (D-score)---for both positive and negative emotions. The D-score is defined simply as: 
\[
\text{D-score} = \text{Mean}_{eng} - \text{Mean}_{diseng}
\]
where a higher D-score is linked to an increased propensity to express engaging emotions more than disengaging emotions. Variance is calculated across the different prompt samples, and significance for cross-cultural comparison is provided using a T-test for the means of the samples from each culture. We also reproduce the same analysis on the human data. The results for humans are shown in Fig. \ref{fig:human_situational_analysis}, and those for models in Figures \ref{fig:gpt_situational_analysis} through \ref{fig:qwen_situational_analysis}. 

\textbf{Positive Emotions.} For positive emotions, only positive personal situations show an increased expression of disengaging emotions in humans, with the difference being larger for EA. In other words, in positive personal situations, EA individuals express disengaging emotions more strongly than LAs. For most models, positive personal situations indeed elicit stronger disengaging emotions across both personas. However, the cross-cultural trend is not captured: the D-score across all models is \textit{significantly more negative} for LA than for EA, and in some cases (Gemini and Qwen) the score is altogether positive for EA. This is again in line with our main findings, where models associate a strong expression of engaging emotions with the EA persona, even in situations where humans show a clear opposite trend. 
For all the other three situations, humans show an increased expression of engaging emotions (positive D-score) for both cultural groups, with the LA group expressing significantly more. This is also a trend that holds partially for LLMs: for positive social situations, all models other than DeepSeek show a more positive D-score for EA than LA, again highlighting that models with the EA persona express engaging emotions more. For negative situations, however, most models (except Mistral and Qwen) reflect a more positive D-score for LA than EA. Mistral and Qwen are misaligned even there, with an increased positive D-score for EA for negative social situations. Interestingly enough, across all models with the EA persona, the D-score is zero for positive emotions in negative personal situations. Not only that, but there is also zero variance in the rating pool, denoting that in negative situations, across all independent samples, all models respond with the same rating for all positive emotions---both engaging and disengaging. This is further testament to the highly deterministic nature of LLM responses, exacerbated in this case by the presense of EA persona, and a mismatched situation (personal negative) -- emotion (positive) context. 

\textbf{Negative Emotions.} Humans express similar levels of engaging and disengaging \textit{negative emotions} in \textit{positive situations}, exemplified by the near-zero D-scores in Fig. \ref{fig:human_situational_analysis} (right panel). A similar, more deterministic result is observed across all models except Mistral. For all models, positive (personal or social) situations yield near-zero D-scores for negative situations. Note again that the variance in this case becomes zero in most cases as well. In some cases, as with GPT, Gemini, and DeepSeek, the D-score for LA is marginally greater than EA (and marginally positive) for positive personal situations, whereas for Qwen, it is marginally negative. This denotes that most models approximately follow the human trend when expressing negative emotions in positive situations. Moreover, in this case too, as with positive emotions, model responses are highly deterministic for a mismatched situation-emotion context. 
For negative situations, firstly, both cultural groups expressed disengaging emotions more than engaging emotions (negative D-scores). Further, human participants from the EA group showed a greater tendency to express disengaging emotions, than the LA group, denoted by the more negative D-score, particularly for negative social situations. All LLMs, however, fail to capture this. For negative personal scenarios, models either rate engaging and disengaging emotions equally, or rate engaging emotions higher (GPT, Qwen). Further, for negative social scenarios, they show more negative D-scores with the LA persona (except Mistral and Qwen) than with the EA persona. This is exactly the opposite of the human finding, where most models continue to provide higher expressivity ratings for engaging emotions with the EA persona.

\subsubsection{Inter-Model Homogeneity}

\begin{figure}[h]
        \centering
        \includegraphics[width=\linewidth]{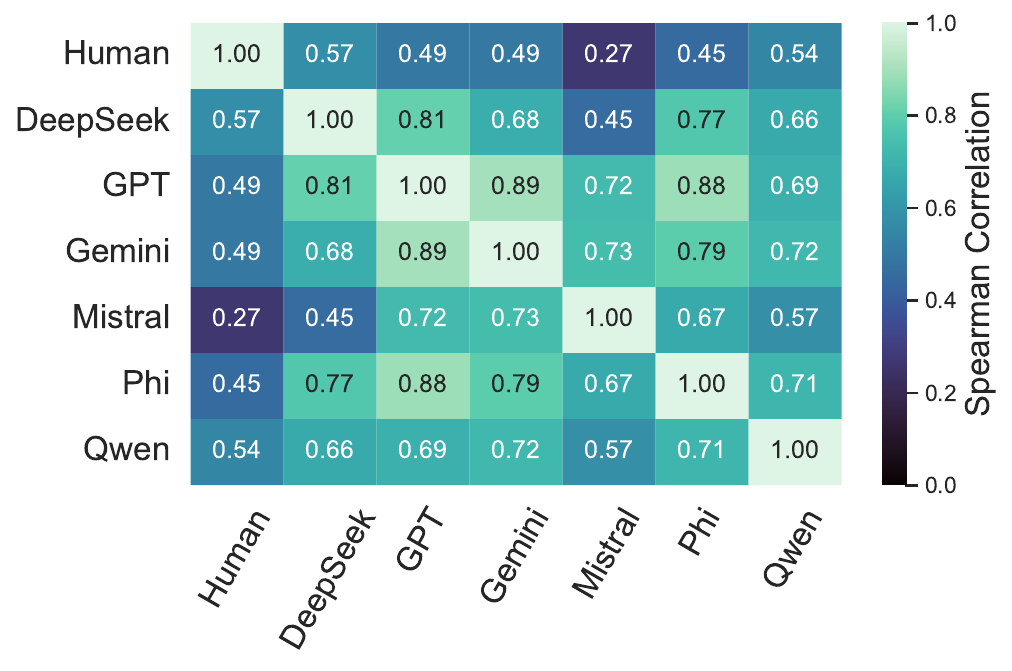}
        \caption{Inter-model homogeneity quantified through inter-model and model-human Pearson correlation coefficients. The mean value of \textit{LLM-LLM} correlation is \textbf{0.72}, and \textit{LLM-human} correlation is \textbf{0.47}.}
        \label{fig:inter_model_determinism_heatmap}    
\end{figure}

Here, we present the complete Pearson correlation values calculated between each pair of models, and for each model with the human distribution, depicted in Fig. \ref{fig:inter_model_determinism_heatmap}. Note that the values are calculated over the entire space of responses, across all emotions and cultures. 

\subsection{Modifying Sampling Temperatures}
\subsubsection{Exact Temperature Values for Models}
For each model studied, we begin with the highest possible temperature setting and check the quality of responses. Note that for GPT-o4, which is a reasoning model, only the default temperature setting is allowed to be set. Also for two of the open-source models, Phi and Mistral, we study a wide range of temperatures, described in the following subsection. For all other models, we start with a temperature value of 2.0. Most models provide incoherent responses, with Qwen providing completely gibberish outputs. We then test with a value of 1.5, for which all proprietary models are found to provide coherent outputs. We thus use the value of 1.5 for both DeepSeek R1 and Gemini 2.5 Flash. For open-source models, responses are no longer gibberish at temperatures of 1.5, but models struggle to respond in the correct format, leading to a large percentage of responses getting skipped in post-processing (over 70\%). Thus, for open-source models, we test with a temperature value of 1, and find responses to be both structured and coherent. Thus, we set the value to 1 for Phi, Mistral and Qwen and report results with this value in the comparison. 

\subsubsection{Analysing Greater Range of Temperatures for Open-Source Models}
% 1. detailed results for open-source models calculated across multiple different temperature levels.
Table~\ref{tab:temp_mistral} summarizes the results of the temperature ablation experiments conducted with Mistral. For each culture (Chile, Mexico, and the United States), we compared response distributions across temperature settings (0.2, 0.7, 1.0, 1.3, and 2.0). Statistical significance was assessed using the $p$-value ($P$), while practical significance was evaluated using the rank-biserial correlation ($R$). The table reports the number of prompt comparisons falling into each significance category. Overall, changing the temperature parameter produced minimal practical differences in the distributions of responses. A majority of the comparisons fall into the category where $P \geq 0.05$, indicating no statistically significant difference between temperature conditions. Even among comparisons that yield statistically significant $p$-values ($P < .05$), most exhibit very small rank-biserial correlations ($R < .13$), indicating negligible practical significance.

This pattern occurs because many prompts produce nearly identical ordinal responses across temperature settings. In such cases, most responses cluster around the same values (e.g., ``4'' or ``5''), and the distributions largely overlap. The Mann--Whitney $U$ test can still detect a systematic difference in ranks, for example, when one group tends to score slightly higher than another by a single rank, leading to statistically significant $p$-values. However, when the magnitude of this shift is extremely small, the rank-biserial correlation remains close to zero, indicating that the practical effect is trivial. Reporting the rank-biserial correlation alongside the $p$-value therefore helps prevent overinterpretation of statistically significant results when the effect size is negligible. Across all cultures, comparisons between more distant temperature settings (e.g., $0.2$ vs.\ $2.0$) show a slightly higher number of statistically significant tests, but these also largely correspond to very small effect sizes. Comparisons between intermediate settings (e.g., $0.7$ vs.\ $1.3$) occasionally yield significant $p$-values, yet the associated rank-biserial correlations remain small, again suggesting that the distributions are nearly identical in practice.

Qualitatively, increasing the temperature primarily influenced the format and verbosity of the model’s outputs rather than the numerical scores themselves. At higher temperature values, the model more frequently deviated from the scoring template specified in the prompt and generated additional explanatory or irrelevant text. Despite this increased variability in formatting, the underlying scores reported by the model remained largely stable across temperature settings. A similar analysis was conducted for the other open-source model, Phi. Comparable trends were observed at lower temperatures. However, the analysis could not be completed at higher temperatures, as the model’s outputs became increasingly verbose and it stopped assigning scores, resulting in rows with missing values that were subsequently dropped. At these higher temperatures, the number of dropped rows effectively equaled the sample size.

\begin{table*}[hbtp]
\centering
\caption{Temperature Ablation - Mistral}
\begin{tabularx}{\textwidth}{
    >{\centering\arraybackslash}p{2.5cm}  % Culture Display
    >{\centering\arraybackslash}p{3cm}    % Comparison Set
    *{6}{>{\centering\arraybackslash}m{1.5cm}}   % Other columns
}
\toprule
\textbf{\makecell[c]{Culture }} &
\textbf{\makecell[c]{Temperature \\ Comparison Set}} &
\textbf{\makecell[c]{\scriptsize $P\ge0.05$}} &
\textbf{\makecell[c]{\scriptsize $P<.05$\\ \scriptsize $R<.13$}} &
\textbf{\makecell[c]{\scriptsize $P<.05,$\\ \scriptsize $.13\!\le\!R<.30$}} &
\textbf{\makecell[c]{\scriptsize $P<.05$\\ \scriptsize $.30\le R<.46$}} &
\textbf{\makecell[c]{\scriptsize $P<.05$\\ \scriptsize $R\ge.46$}} \\
\midrule

% Chile
\multirow{9}{*}{Chile}
& $0.2\,\mathrm{vs}\,0.7$ & \gradientcell{24} & \gradientcell{12} & \gradientcell{9} & \gradientcell{2} & \gradientcell{1} \\
& $0.2\,\mathrm{vs}\,1.0$ & \gradientcell{19} & \gradientcell{13} & \gradientcell{12} & \gradientcell{3} & \gradientcell{1} \\
& $0.2\,\mathrm{vs}\,1.3$ & \gradientcell{20} & \gradientcell{10} & \gradientcell{11} & \gradientcell{6} & \gradientcell{1} \\
& $0.2\,\mathrm{vs}\,2.0$ & \gradientcell{14} & \gradientcell{11} & \gradientcell{17} & \gradientcell{4} & \gradientcell{2} \\
& $0.7\,\mathrm{vs}\,1.0$ & \gradientcell{41} & \gradientcell{6} & \gradientcell{1} & \gradientcell{0} & \gradientcell{0} \\
& $0.7\,\mathrm{vs}\,1.3$ & \gradientcell{33} & \gradientcell{9} & \gradientcell{6} & \gradientcell{0} & \gradientcell{0} \\
& $1.0\,\mathrm{vs}\,1.3$ & \gradientcell{39} & \gradientcell{8} & \gradientcell{1} & \gradientcell{0} & \gradientcell{0} \\
& $1.0\,\mathrm{vs}\,2.0$ & \gradientcell{27} & \gradientcell{13} & \gradientcell{8} & \gradientcell{0} & \gradientcell{0} \\
& $1.3\,\mathrm{vs}\,2.0$ & \gradientcell{39} & \gradientcell{5} & \gradientcell{4} & \gradientcell{0} & \gradientcell{0} \\
\midrule

% Mexico
\multirow{9}{*}{Mexico}
& $0.2\,\mathrm{vs}\,0.7$ & \gradientcell{21} & \gradientcell{17} & \gradientcell{8} & \gradientcell{2} & \gradientcell{0} \\
& $0.2\,\mathrm{vs}\,1.0$ & \gradientcell{18} & \gradientcell{15} & \gradientcell{10} & \gradientcell{4} & \gradientcell{1} \\
& $0.2\,\mathrm{vs}\,1.3$ & \gradientcell{15} & \gradientcell{14} & \gradientcell{14} & \gradientcell{4} & \gradientcell{1} \\
& $0.2\,\mathrm{vs}\,2.0$ & \gradientcell{12} & \gradientcell{11} & \gradientcell{17} & \gradientcell{6} & \gradientcell{2} \\
& $0.7\,\mathrm{vs}\,1.0$ & \gradientcell{38} & \gradientcell{8} & \gradientcell{2} & \gradientcell{0} & \gradientcell{0} \\
& $0.7\,\mathrm{vs}\,1.3$ & \gradientcell{26} & \gradientcell{14} & \gradientcell{8} & \gradientcell{0} & \gradientcell{0} \\
& $1.0\,\mathrm{vs}\,1.3$ & \gradientcell{40} & \gradientcell{8} & \gradientcell{0} & \gradientcell{0} & \gradientcell{0} \\
& $1.0\,\mathrm{vs}\,2.0$ & \gradientcell{29} & \gradientcell{10} & \gradientcell{8} & \gradientcell{1} & \gradientcell{0} \\
& $1.3\,\mathrm{vs}\,2.0$ & \gradientcell{38} & \gradientcell{6} & \gradientcell{3} & \gradientcell{1} & \gradientcell{0} \\
\midrule

% United States
\multirow{9}{*}{United States}
& $0.2\,\mathrm{vs}\,0.7$ & \gradientcell{30} & \gradientcell{9} & \gradientcell{7} & \gradientcell{2} & \gradientcell{0} \\
& $0.2\,\mathrm{vs}\,1.0$ & \gradientcell{30} & \gradientcell{7} & \gradientcell{7} & \gradientcell{4} & \gradientcell{0} \\
& $0.2\,\mathrm{vs}\,1.3$ & \gradientcell{25} & \gradientcell{10} & \gradientcell{7} & \gradientcell{6} & \gradientcell{0} \\
& $0.2\,\mathrm{vs}\,2.0$ & \gradientcell{24} & \gradientcell{3} & \gradientcell{12} & \gradientcell{8} & \gradientcell{1} \\
& $0.7\,\mathrm{vs}\,1.0$ & \gradientcell{42} & \gradientcell{5} & \gradientcell{1} & \gradientcell{0} & \gradientcell{0} \\
& $0.7\,\mathrm{vs}\,1.3$ & \gradientcell{30} & \gradientcell{13} & \gradientcell{5} & \gradientcell{0} & \gradientcell{0} \\
& $1.0\,\mathrm{vs}\,1.3$ & \gradientcell{42} & \gradientcell{4} & \gradientcell{2} & \gradientcell{0} & \gradientcell{0} \\
& $1.0\,\mathrm{vs}\,2.0$ & \gradientcell{31} & \gradientcell{5} & \gradientcell{11} & \gradientcell{1} & \gradientcell{0} \\
& $1.3\,\mathrm{vs}\,2.0$ & \gradientcell{37} & \gradientcell{7} & \gradientcell{4} & \gradientcell{0} & \gradientcell{0} \\
\bottomrule
\end{tabularx}
\label{tab:temp_mistral}
\end{table*}

\subsubsection{Change in Response Diversity with Increased Temperature}

\begin{table}[H]
\centering
\small
\setlength{\tabcolsep}{5pt}
\resizebox{\columnwidth}{!}{
\begin{tabular}{
    l
    @{\hspace{6pt}\vrule\hspace{6pt}}
    c c c c c c
}
\toprule
& \textbf{GPT} 
& \textbf{Gemini} 
& \textbf{DeepSeek} 
& \textbf{Phi} 
& \textbf{Mistral} 
& \textbf{Qwen} \\
\midrule
USA    & -0.0001 & 0.04 & -0.09 & 0.02 & 0.003 & 3.426*** \\
Mexico & -0.001 & 0.04 & 0.28*** & 0.006 & -0.011 & 0.446*** \\
Chile  & 0.135* & -0.06 & 0.13* & 0.06 & -0.044 & 0.798*** \\
\bottomrule
\end{tabular}}
\caption{Change in effective number of response categories ($\Delta N_{\text{eff}}$) across models and cultures. Asterisks denote statistical significance (* $p<0.05$, ** $p<0.01$, *** $p<0.001$).}
\label{tab:neff_compact}
\end{table}

In Table \ref{tab:neff_compact}, we present the results for the change in effective categories ($\Delta N_{\text{eff}}$) of model ratings with increased temperatures. Note that for most models, the change itself is small ($<0.5$), and insignificant. The only exception to that is presented by Qwen, where an increase in effective categories as large as 3.426 is observed, with the EA (USA) persona. This specific combination (Qwen-USA) is, however, found in the main results to provide the most deterministic and modal responses, utilizing only between 1--2 effective categories.

\subsection{Contextual Nudge: Ablation with Pairwise Comparisons}

In this section, we provide the exact prompt formats for all of the settings used --- namely, the emotion comparison (PSE vs. PSD and NSE vs. NSD), and the persona comparison. 

For the emotion comparison, the following prompt template is used, and replicated similarly for all cultures. Given the improved alignment observed with English language for LA personas, the pairwise comparisons for all cultures are carried out in English. 

\begin{promptbox}
\small
    \textbf{Contextual Nudge (pairwise comparison): PSE (or NSE) vs. PSD (or NSD)}

    Imagine you are like a human, capable of feeling emotions. We will ask you to choose between two emotions you would express in different social situations. By expressing emotions, we mean using gestures or facial expressions to show how you feel. Read the provided social situation carefully and think which of the two emotions you would express more strongly and appropriately when you are discussing the situation with your friends and family members. Now, read the following instructions and answer for the specific pair of emotions.
    
    Answer as the following person: 
    You are a white American person. You are a citizen of the USA, who was born and lives in the USA. You are ethnically European American or Caucasian. Given situation: You succeeded in an exam or assignment. 
    
    Which of the following two emotions would you express more strongly in this situation: emotion A or emotion B
    
    Answer with the name of the emotion from the two options provided.
    
    You must choose one of the two emotions provided. You have to make a choice. You can only choose from the two option provided. Do not provide an emotion other than the two options provided. Do not provide any explanations or additional text. Only provide the name of the emotion you would express more strongly in this situation.
    
    Answer in a JSON format, strictly, providing the choice of emotion you would express in the given situation. The format should strictly be as follows: 
    \{``emotion\_choice": [your chosen emotion]\}
\end{promptbox}

For the above prompt format, the options emotion A and emotion B are all possible pairs of PSE and PSD emotions, or NSE and NSD emotions. Note that, for each case, there are 4 unique emotion pairs possible (2 PSE/NSE and 2 PSD/NSD emotions). These pairs are presented along with each scenario (=4), leading to 16 unique prompts in each emotion-based comparison. Each of these prompts is sampled 190 times independently, leading to a total of 18240 (=2 comparisons (positive, negative) $\times$ 16 unique prompts $\times$ 190 independent samples $\times$ 3 personas) additional generations. Also note that at the end of the prompt, the model is forced to make a choice between the pair of emotions provided. Through pilot experiments, we find that without providing this additional statement, models provide the names of arbitrary emotion categories that are related, but cannot be directly categorized into either engaging or disengaging emotions.\\

Next, we present the prompt template for the persona choice prompts. Note that in these experiments, the expressivity of the cultural persona is judged through explicit questioning, and thus, the model itself is not provided with any cultural persona. In a way, it thus also examines whether models would respond similarly when endowed with a cultural persona (original experiments), versus when they are asked to make judgements about the same cultural persona explicitly. 

\begin{promptbox}
\small
    \textbf{Contextual Nudge (pairwise comparison): Comparing Personas}
    ``Imagine you are like a human, capable of feeling emotions. We will ask you to choose between two personas who you think would express the provided emotion more in different social situations. By expressing emotions we mean using gestures or facial expressions to show how you feel. Read the provided social situation carefully and think which of the two personas would express the given emotion more strongly when they are discussing the situation with their friends and family members. Now, read the following instructions and answer for the specific pair of personas.

    Given situation: You succeeded in an exam or assignment. Expressed emotion: You strongly express the emotion of ashamed.

    Which of the following two personas would express the emotion more strongly in this situation: 
    A white American person who is a citizen of the USA, and was born and lives in the USA. They are ethnically European American or Caucasian OR 
    A Chilean person who is a citizen of Chile, and was born and lives in Chile.

    Answer with the name of the persona from the two options provided. You must choose one of the two personas provided. You have to make a choice. You can only choose from the two option provided. Do not provide a persona other than the two options provided. Do not provide any explanations or additional text. Only provide the name of the persona that would express the given emotion more strongly in this situation.

    Answer in a JSON format, strictly, providing the choice of a persona that would express the given emotion more strongly in the given situation. The format should strictly be as follows: 

    \{``persona\_choice": [your chosen persona]"\}."
    
\end{promptbox}

In these prompts, all combinations of situations (=4) and emotions (=8) are examined. Along with this, all possible pairs of personas are examined (=6), although the final results reflect only comparisons between EA and LA personas (excluding the pair of Mexico and Chile being compared). Again, all prompts are sampled independently 190 times, leading to a total of 36480 (=4 $\times$ 8 $\times$ 6 $\times$ 190) responses per model. 

\subsection{Limitations}
Our experiments report some insightful findings on the misalignment between LLMs and human understanding of social emotions. Despite the controlled design and systematic evaluation, some limitations constrain the scope and generalizability of our findings. Our experiments are conducted on specific versions of SOTA LLMs. With the accelerated training and versioning of LLMs, these results might change with newer and more capable versions of these models. Experiments and studies grounded in social and emotional psychology tend to either focus on a wide breadth of emotions and contexts or conduct in-depth evaluations along a narrow axis. In our work, we prioritize depth, i.e., our results do not focus on broad aspects of how LLMs understand a wide list of emotions in the context of multiple cultures. In our experiments, we conduct rigorous in-depth evaluation of the expressivity of social emotions in the context of two cultures: European American (USA) and Latin American (Chile and Mexico). Consequently, all our findings about the capabilities and misalignments of LLMs in understanding social-emotional expression are limited to these cultures. In future work, we plan to expand our list of cultures, especially by including East Asian cultures like Japan, and produce more globally generalizable results.